\documentclass[letterpaper,journal]{IEEEtran}  
 
\IEEEoverridecommandlockouts 

\usepackage{array}
\usepackage{amsmath,amsfonts} 
\usepackage{graphicx}
\usepackage{caption}
\usepackage{subcaption}
\usepackage{CJKutf8}
\usepackage{multirow}
\usepackage{cite}
\usepackage{xcolor}
\usepackage{algorithm2e}
\usepackage{booktabs}
\usepackage{leftidx}
\usepackage{makecell}
\usepackage[breaklinks,colorlinks]{hyperref}
\usepackage{mathtools}

\graphicspath{{assets}{./assets}}

\newcommand{\revise}[1]{\noindent{\color{black}\text{}{#1}}}

\begin{document}

\title{\LARGE \bf
SG-Reg: Generalizable and Efficient Scene Graph Registration}

\author{Chuhao Liu, Zhijian Qiao, Jieqi Shi$^{\ast}$, Ke Wang, Peize Liu and Shaojie Shen
\thanks{Manuscript submitted on 8 Oct, 2024. Accepted on 19 Apr, 2025.}
\thanks{${\ast}$ Corresponding author: Jieqi Shi.}
\thanks{Chuhao Liu, Zhijian Qiao, Peize Liu and Shaojie Shen are with the Department of Electronic and Computer Engineering, the Hong Kong University of Science and Technology, Hong Kong, China. (email: \{cliuci,zqiaoac,pliuan\}@connect.ust.hk; eeshaojie@ust.hk)}
\thanks{Jieqi Shi is with the State Key Laboratory for Novel Software Technology, Nanjing University, Nanjing, China. (email: isjieqi@nju.edu.cn)}
\thanks{Ke Wang is with the School of Information Engineering, Chang’an
University. (email: kwangdd@chd.edu.cn)}

}

\markboth{IEEE Transaction on Robotics}{}

\maketitle
\begin{abstract}
\revise{This paper addresses the challenges of registering two rigid semantic scene graphs, an essential capability when an autonomous agent needs to register its map against a remote agent, or against a prior map.} The hand-crafted descriptors in classical semantic-aided registration, \revise{or} the ground-truth annotation reliance in learning-based scene graph registration, impede their application in \revise{practical} real-world environments. 
To address the challenges, we design a scene graph network to encode multiple modalities of semantic nodes: open-set semantic feature, local topology with spatial awareness, and shape feature. \revise{These modalities are fused to create compact semantic node features. The matching layers then search for correspondences in a coarse-to-fine manner.} In the back-end, we employ a robust pose estimator to decide transformation according to the correspondences. 
We manage to maintain a \revise{sparse and hierarchical scene representation. }\revise{Our approach demands fewer GPU resources and fewer communication bandwidth in multi-agent tasks.} \revise{Moreover, we design a new data generation approach using vision foundation models and a semantic mapping module to reconstruct semantic scene graphs. It differs significantly from previous works, which rely on ground-truth semantic annotations to generate data.} 
We validate our method in a two-agent SLAM benchmark. 
It significantly outperforms the hand-crafted baseline in terms of registration success rate. \revise{Compared to visual loop closure networks, our method achieves a slightly higher registration recall while requiring only 52 KB of communication bandwidth for each query frame.}
Code available at: \href{http://github.com/HKUST-Aerial-Robotics/SG-Reg}{http://github.com/HKUST-Aerial-Robotics/SG-Reg}.
\end{abstract}

\section{INTRODUCTION}\label{sec-intro}
Global registration in dense indoor scenes is crucial for visual simultaneous localization and mapping (SLAM) systems. It plays a key role in the performance of multi-agent SLAM\cite{xu2022dslam}, multi-session SLAM\cite{campos2021orb3,yin2023automerge}, and long-term SLAM\cite{toft2020long} systems. Relying on image matching\cite{arandjelovic2016netvlad,sarlin20superglue,lindenberger2023lightglue} for loop closure detection and global registration is challenging because image matching is sensitive to viewpoint differences. Furthermore, in multi-agent SLAM, broadcasting image data demands significant communication bandwidth.
To address the limitations of visual-based loop closure detection, semantic-aided registration has been proposed\cite{cubeslam2019yang,fusion++2018McCormac,Lin2021topobj}, offering stronger viewpoint invariance and a more data-efficient representation, thereby reducing communication bandwidth.

\begin{figure}[t]
    \centering    
    \includegraphics[width=1.01\columnwidth]{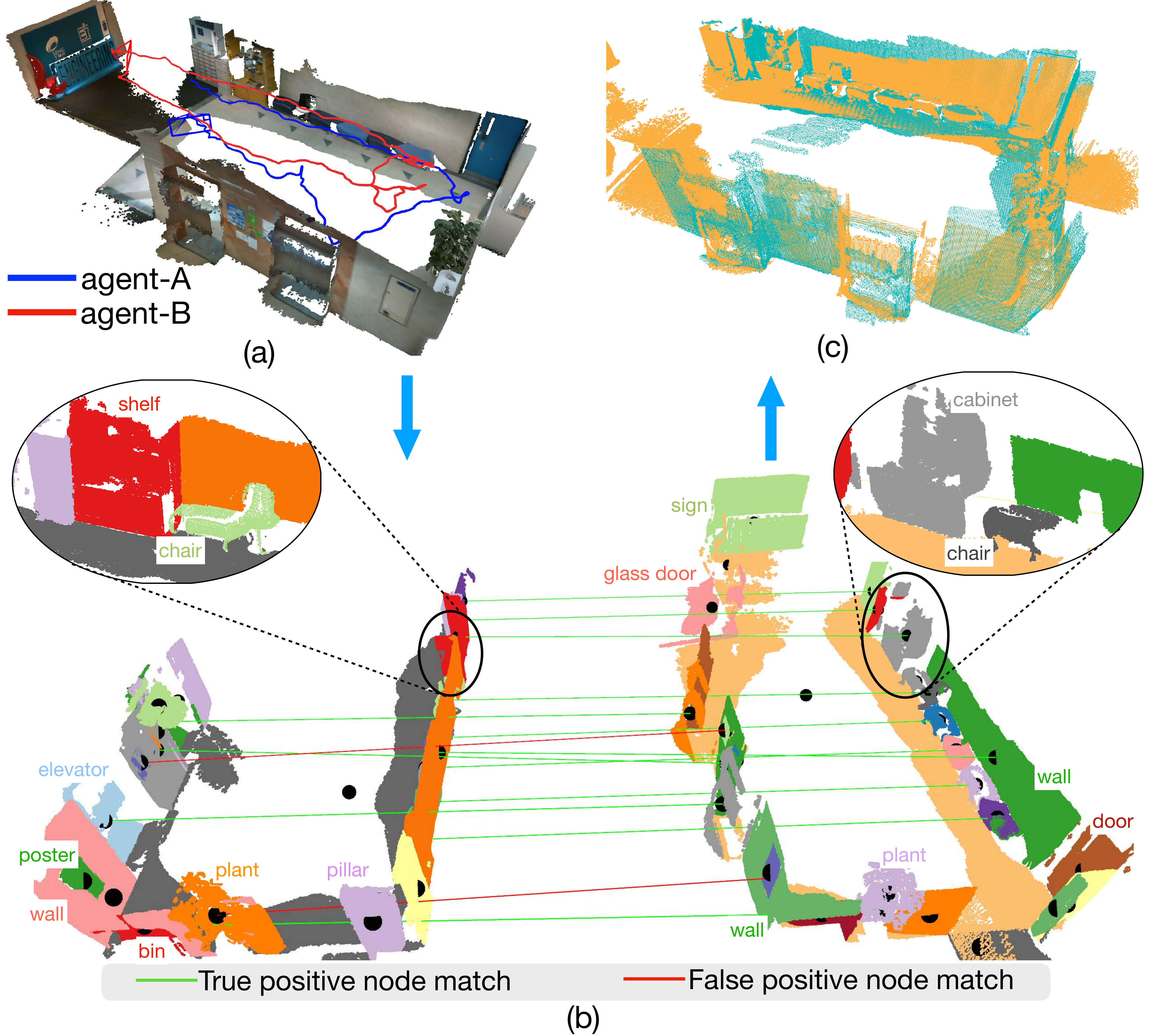}\vspace{-0.2cm}
    \caption{Register the semantic scene graphs in the two-agent SLAM system. (a) Captured RGB-D sequences from the two agents in a real-world indoor scene. The two agents move in an opposite direction, creating a large viewpoint difference between their cameras. (b) Visualization of the matched nodes between the semantic scene graphs, which are from the two agents. The scene graphs are constructed using FM-Fusion\cite{liu2024fmfusion}. The zoomed subvolume showcases examples of inconsistent semantic nodes. For better visualization, only a subset of the semantic labels is displayed. (c) Registration result.}
    \label{fig1}\vspace{-0.3cm}
\end{figure}

Early works in \revise{semantic-aided registration explicitly maintain the semantic objects. To match re-observed semantic objects, they design hand-crafted descriptors and similarity metrics based on 2D image feature\cite{fusion++2018McCormac}, projected bounding box \cite{Bowman2017prob,cubeslam2019yang}, quadratic representation\cite{quadricslam2018}, or graph topology\cite{Abel2018xview,Lin2021topobj,hughes2022hydra}. Using the calculated similarity metric, they identify matched nodes by either setting a threshold or designing a set of verification rules. 
Although these works introduced pioneering SLAM frameworks that incorporate semantic landmarks, they often face long-tailed cases in real-world experiments due to the noise commonly present in the reconstructed semantic objects.}

In recent years, learning-based semantic scene graph registration methods \cite{sarkar2023sgaligner,xie2024sgpgm} have emerged. These methods encode the \revise{semantic scene graph and determine multiple levels of correspondences by solving partial assignment problems}. They eliminate the need for hand-crafted similarity metrics between semantic nodes. However, the current scene graph learning methods are trained on datasets with ground-truth annotations, resulting in a significant domain gap between the semantic scene graphs from the annotated dataset and those in the real-world scenarios, especially when noise is present. \revise{As a result, these learning-based methods face challenges in generalization and demonstrate limited effectiveness in real-world experiments.}

Such \revise{real-world semantic noise and sensor noise motivate us to propose a new data generation method.} We reconstruct the semantic scene graphs using three vision foundation models \cite{zhang2023ram,liu2023grounding,kirillov2023sam} and a semantic mapping module: FM-Fusion \cite{liu2024fmfusion}.
In real-world indoor environments, the reconstructed semantic scene graphs exhibit higher quality compared to those generated using Mask R-CNN \cite{he2017mask} combined with earlier semantic mapping methods \cite{mccormac2017semanticfusion,Kimera2020Rosinol}. \revise{Although our method effectively limits the semantic noise to appropriate levels, it cannot completely eliminate the noise. For example, as shown in Fig. \ref{fig1}, the zoomed subvolumes from the two scene graphs reveal semantic inconsistencies due to noise in real-world reconstructions. In contrast to previous scene graph learning methods \cite{sarkar2023sgaligner,xie2024sgpgm}, which rely on ground-truth semantic annotations to generate data, our approach introduces a new and practical challenge.}

To address the challenges of noisy scene graphs and \revise{achieve generalizable registration}, we propose encoding multiple types of information in the scene graph: semantic label, local topology, and geometric shape. The semantic label is encoded by the \revise{pre-trained BERT\cite{devlin2018bert} model, which is known for its strong generalization capabilities and naturally accepts open-set labels}. In learning the local topology {of a} semantic node, we introduce a triplet descriptor and integrate it into the graph neural network (GNN)\cite{brody2021gatv2} to boost the spatial awareness of the learned feature. It enhances the descriptive power of the local topology while ensuring invariance to a global transformation in 4 degree-of-freedom (DoF). \revise{For shape information, the shape network reads the point cloud of each semantic node and aggregates a geometric shape feature. We fuse all the information to generate a comprehensive semantic node feature. The scene graph represents the original dense 3D point cloud into sparse object-level features, thus enabling the network layers to process the whole scene while demanding fewer GPU resources.}
\revise{The last module of our network consists of two levels of graph-matching layers: one for matching semantic nodes and another for matching their point clouds. Correspondences at each level are efficiently determined using an optimal transport algorithm\cite{Cuturi2013sinkhorn}. }

The \revise{back-end reads the point correspondences. It prunes outliers via maximum clique\cite{mangelson2018pcm} and estimates pose through the robust estimator G3Reg \cite{qiao2024g3reg}. The back-end is designed to tolerate a certain degree of point outliers, thereby further enhancing the overall generalization ability in cross-domain experiments, where the outlier ratio tends to be relative high.}

This work focuses on the semantic scene graph registration task in indoor visual-inertial SLAM conditions. We highlight the key contributions of our work as follows:
\begin{itemize}
    \item We introduce a learning-based semantic scene graph registration approach. Our approach employs a \revise{triplet-boosted GNN layer}, which is designed to capture the semantic nodes' topological relationships along with spatial awareness. Compared to a vanilla GNN network, it offers a more descriptive representation of semantic nodes while maintaining invariance under global transformations in 4-DoF.
    \item \revise{The shape network generates a shape feature for each semantic node, which is then fused into the node's feature representation to improve scene graph matching performance. Thanks to the sparse nature of semantic nodes, the computational complexity and GPU memory consumption are significantly reduced.}
    \item We propose an automatic data generation method for scene graph learning. We utilize the vision foundation models \cite{zhang2023ram,liu2023grounding,kirillov2023sam} and a semantic mapping\cite{liu2024fmfusion} module to construct semantic scene graphs. \revise{This innovative data generation approach enables training the network using posed RGB-D sequences, eliminating the reliance on ground-truth annotations.} Furthermore, it significantly reduces the domain gap between the training data and real-world evaluation data.
    \item \revise{We thoroughly evaluate our method against the baseline model \cite{xie2024sgpgm} using real-world reconstructed scene graphs from 3RScan \cite{wald2019rio}. Our approach achieves significantly higher registration recall than SG-PGM \cite{xie2024sgpgm}, while requiring substantially less GPU resources.} 
    \item We deploy our work in a two-agent SLAM system and \revise{register the two scene graphs in a coarse-to-fine paradigm}, as shown in Fig. \ref{fig1}. With far lower communication bandwidth, our \revise{registration substantially outperforms hand-crafted semantic descriptor}\cite{hughes2022hydra} and achieves a success rate \revise{$0.7\%$ higher than} the combination of NetVLAD and LightGlue\cite{arandjelovic2016netvlad,lindenberger2023lightglue,sarlin2019coarse}. 
\end{itemize}

\section{Related works}
\subsection{Image registration}
The current state-of-the-art (SOTA) visual SLAM systems\cite{qin2018vins, campos2021orb3, Kimera2020Rosinol} rely on DBoW\cite{bow2012galvez} to detect loop closure and register camera poses. \revise{In VINS-Mono\cite{qin2018vins}, for example, if DBoW detects a loop closure candidate, it extracts BRIEF descriptors and performs brute-force matching between the looped images. Thereafter, VINS-Mono registers images using RANSAC-based Perspective-n-Point \cite{brachmann2017dsac}.} To reject false loop closure or inaccurate registration, the visual SLAM systems employ comprehensive geometric verification in pose graphs\cite{qin2018vins,Kimera2020Rosinol} or co-visibility graphs\cite{campos2021orb3}. 

In recent years, learning-based image matching has been proposed and primarily improves registration performance in visual SLAM. SuperGlue\cite{sarlin20superglue} and LightGlue \cite{lindenberger2023lightglue} encode image features using attention layers and search feature correspondences. \revise{NetVLAD\cite{arandjelovic2016netvlad} is combined with them} to detect loop closure in a coarse-to-fine paradigm\cite{sarlin2019coarse,sarlin2022lamar}. These learning-based methods significantly outperform the classical loop closure detection in visual SLAM. \revise{Nevertheless, in multi-agent SLAM, the learning-based image registration methods require considerable communication bandwidth. Additionally, they often result in false registrations under significant viewpoint disparities.}

\subsection{Point cloud registration}
Given that \revise{visual SLAM is capable of generating a dense point cloud map\cite{dai2017bundlefusion}, it can be integrated with point cloud registration to compute a global transformation.} Traditional methods for point cloud registration\cite{zhou2016fastreg} involve extracting FPFH features\cite{rusu2009fpfh} from the input clouds and solving the pose estimation problem using the Gaussian-Newton method. Subsequently, robust pose estimators\cite{yang2020teaser,qiao2024g3reg} were introduced to estimate poses even with a significant outlier presence. 

Meanwhile, numerous learning-based PCR\cite{wang2019dcp,choy2020deep,qin2022geometric} are proposed. They use 3D convolution network\cite{thomas2019kpconv} or PointNet\cite{qi2017pointnet} as a point cloud backbone, leveraging the extracted features to establish correspondences. They subsequently solve the relative pose in a closed-form using singular value decomposition (SVD) or weighted SVD\cite{besl1992method}. Among the learning-based PCR, GeoTransformer \cite{qin2022geometric} demonstrates promising registration performance in the 3DMatch dataset\cite{zeng20173dmatch}, which includes partial scans of point clouds in real-world indoor environments. Despite its superior capabilities, GeoTransformer encodes a superpoint feature and incorporates attention layers\cite{vaswani2017attention} on these superpoints. \revise{Due to the density of the superpoints, the attention operation demands substantial GPU resources, limiting its scalability in large-scale scenes.}

\subsection{Scene Graph Construction}
In order to \revise{integrate a scene graph into visual SLAM, the initial step is to construct a scene graph.} 
Kimera\cite{Kimera2020Rosinol} and its series of works are pioneer works in this area. Using Mask R-CNN\cite{he2017mask}, Kimera incorporates semantic segmentation obtained from images into a metric-semantic map. This map is built upon the TSDF voxel grid map\cite{voxblox2017Oleynikova} and is integrated following SemanticFusion\cite{mccormac2017semanticfusion}. Then, Kimera clusters the dense metric-semantic map into hierarchical levels of representations: objects, places, floors, and buildings. 
\revise{Later, S-Graph\cite{bavle2022sgraph} and S-Graph+\cite{bavle2022sgraphs+} segment 3D planes to build their scene graph, which is further integrated in an optimizable factor graph.}
The semantic scene graph closes the gap between robot perception and human perception. It is regarded as the spatial perception engine\cite{hughes2024foundations} for spatial intelligence in the future.

Since the release of vision foundation models, recent advancements in semantic mapping \cite{werby23hovsg,gu2023conceptgraphs} have increasingly incorporated these models to reconstruct 3D semantic maps. Among these, FM-Fusion \cite{liu2024fmfusion} focuses on RGB-D SLAM in indoor environments. It integrates object detections from RAM-Ground-SAM \cite{zhang2023ram,liu2023grounding,kirillov2023sam} to reconstruct an instance-aware semantic map. Like Fusion++ \cite{fusion++2018McCormac}, FM-Fusion represents each semantic instance in a separate TSDF submap. It fuses semantic labels using a Bayesian filter while independently modeling the measurement likelihood from RAM \cite{zhang2023ram} and GroundingDINO \cite{liu2023grounding}. By leveraging vision foundation models, FM-Fusion has achieved more accurate semantic instance segmentation compared to Kimera on the ScanNet benchmark. \revise{Based on the semantic instances, we construct a higher-quality semantic scene graph without relying on ground-truth annotations.}
However, RAM-Ground-SAM still predicts noisy instance segmentation and incorrect semantic labels. \revise{When compared to scene graphs generated using annotations in ScanNet or 3RScan, the reconstructed scene graphs using FM-Fusion exhibit significant noise.}

\begin{table*}[ht]
    \centering
    \begin{tabular}{c|c c c c c c c}
        \toprule
        & \underline{Year} & \multicolumn{3}{c}{\underline{Descriptor in matching nodes/objects}} & \underline{Dense Matching} &\multicolumn{2}{c}{\underline{Real-world Visual SLAM Experiment}} \\
         & & Type & Topology & Shape & & Semantic Annotation & LCD Baselines \\
         \hline
        Bowman \textit{et al.}\cite{Bowman2017prob} & 2017 & Explicit & $\times$ &B-box & $\times$ & DPM & DBoW \\ 
        Fusion++\cite{fusion++2018McCormac} & 2018 & Explicit & $\times$ & $\times$ &2D Points & Mask R-CNN & -\\
        X-View\cite{Abel2018xview} & 2018 & Explicit & \checkmark & $\times$ &$\times$ & SegNet & DBoW,NetVLAD\\ 
        Lin \textit{et al.}\cite{Lin2021topobj} & 2021 & Explicit & \checkmark & B-box & 2D Points & SOLOv2 & DBoW+ORB \\ 
        Kimera\cite{Kimera2020Rosinol} & 2021 & Explicit& $\times$ & B-box & $\times$ & Mask R-CNN & - \\
        Hydra\cite{hughes2022hydra} & 2022 & Explicit & \checkmark & B-box & 2D Points & Mask R-CNN & - \\
        SlideSLAM\cite{liu2024slideslam} & 2024 & Explicit & $\times$ & B-box &$\times$ & YOLO & - \\
        Found. SPR \cite{hughes2024foundations} & 2023 & Exp. \& Imp. & \checkmark & B-box & 2D Points & Mask R-CNN & - \\
        SGAligner\cite{sarkar2023sgaligner} & 2023 & Implicit & \checkmark & $\times$ &3D Points & Ground-truth & N/A \\
        SG-PGM\cite{xie2024sgpgm} & 2024 & Implicit & \checkmark & Point Cloud & 3D Points & Ground-truth & N/A \\
        Living Scenes\cite{zhu2023living} & 2024 & Implicit & \checkmark & Point Cloud &3D Points & Ground-truth & N/A\\
        \textbf{SG-Reg} (Ours) & 2024 & Implicit & \checkmark & Point Cloud &3D Points & RAM-Ground-SAM & NetVLAD+LightGlue \\
        \bottomrule
    \end{tabular}
    \vspace{+0.1cm}
    \caption{A summary of semantic data association methods. It covers the loop closure detection module in semantic SLAM and graph match modules in scene graph registration. The methods based on implicit feature descriptor have incorporated neural networks to predict the data association. The B-box refers to a bounding box and the LCD refers to loop closure detection.}
    \label{tab:related_works}\vspace{-0.3cm}
\end{table*}

\subsection{Semantic Data Association}
Data \revise{association between semantic representations is the most important step to guarantee a successful registration. In semantic SLAM works, The step relies on explicit representations.} 
Bowman et al. \cite{Bowman2017prob} represent each object using a bounding box. If two object nodes belong to the same semantic category and their intersection-over-union (IoU) exceeds a predefined threshold, they are considered a match. Building on Kimera, Hydra \cite{hughes2022hydra} constructs hand-crafted semantic descriptors by aggregating semantic histograms from nearby objects and computes node similarity using these descriptors. More recently, SlideSLAM \cite{liu2024slideslam} determines object similarity based on identical semantic categories and similar bounding box shapes. 
To incorporate topological information into semantic data association, X-View \cite{Abel2018xview} introduces a random walk descriptor (RWD) that records semantic labels along walking routes. \revise{The descriptor represents the topology information explicitly. When calculating the similarity metric, X-View counts the number of rows with identical RWDs to determine a similarity score.}
To enhance the spatial descriptiveness of RWD, Lin et al. \cite{Lin2021topobj} integrate distance information into the arrangement of RWDs. Similarly, Liu et al. \cite{liu2023towards} utilize spatial priors to construct RWDs.
To reduce false matches between nodes, some approaches \cite{Qian2022covloop} explicitly construct edge descriptors to verify geometric consistency among matched nodes. Along the same lines, Julia et al. \cite{julia2023retrieval} employ node triplets to validate the correctness of matched nodes.

The \revise{aforementioned semantic-related descriptors are explicitly represented through labels, bounding boxes, and topology, rather than being implicitly encoded from these attributes.}
They solved their targeting scenarios but faced corner cases in general real-world evaluation. For example, in the inconsistent scene graphs, as shown in Fig. \ref{fig1}, they may need to set a topology or edge threshold for each semantic category. The parameter tuning workload is huge. 

Inspired by the learning-based graph matching models \cite{wang2021neuralgm}, we believe training a neural network to learn semantic data association is a promising direction. The latest version of Hydra \cite{hughes2024foundations} proposes encoding object nodes using a Graph Neural Network (GNN), though it omits shape features. SGAligner \cite{sarkar2023sgaligner} is the first work, to our knowledge, to focus on learning scene graph matching. It encodes multiple modalities from semantic nodes, including semantic labels, center positions, and relationship labels. Subsequently, SG-PGM enhances this approach by incorporating shape features into scene graph learning. It samples points from each semantic node and aggregates them using GATv2 \cite{brody2021gatv2}. \revise{SGAligner and SG-PGM adopted the geometric-related layers from GeoTransformer, including its superpoint matching, point matching, and local-to-global registration layer.} Both of them are trained and evaluated on the 3RScan dataset \cite{wald2019rio}, requiring ground-truth annotations to construct their scene graph.

We provide a comprehensive summary of the semantic data association methods in Table \ref{tab:related_works}. Our approach is fundamentally different from previous semantic SLAM works, which reconstruct explicit semantic representations and create hand-crafted descriptors. In contrast, we learn to encode scene graphs. The work most similar to ours is SG-PGM, but we have three core differences.
First, our triplet-boosted GNN encodes semantic nodes with enhanced spatial awareness, outperforming the vanilla GAT used in SG-PGM. Second, we avoid aggregating point cloud features through attention layers, significantly reducing GPU memory usage and speeding up inference. Third, we evaluate our method in real-world SLAM experiments, whereas SG-PGM and prior learning-based works only assess performance using scene graphs with ground-truth annotations. We integrate the data generation process with semantic mapping \cite{liu2024fmfusion}, allowing us to train and evaluate our network on scene graphs constructed from semantic mapping.\vspace{-0.3cm}


\section{Preliminary}

\vspace{-0.5cm}
\begin{figure}[ht]
    \centering
    \includegraphics[width=\columnwidth]{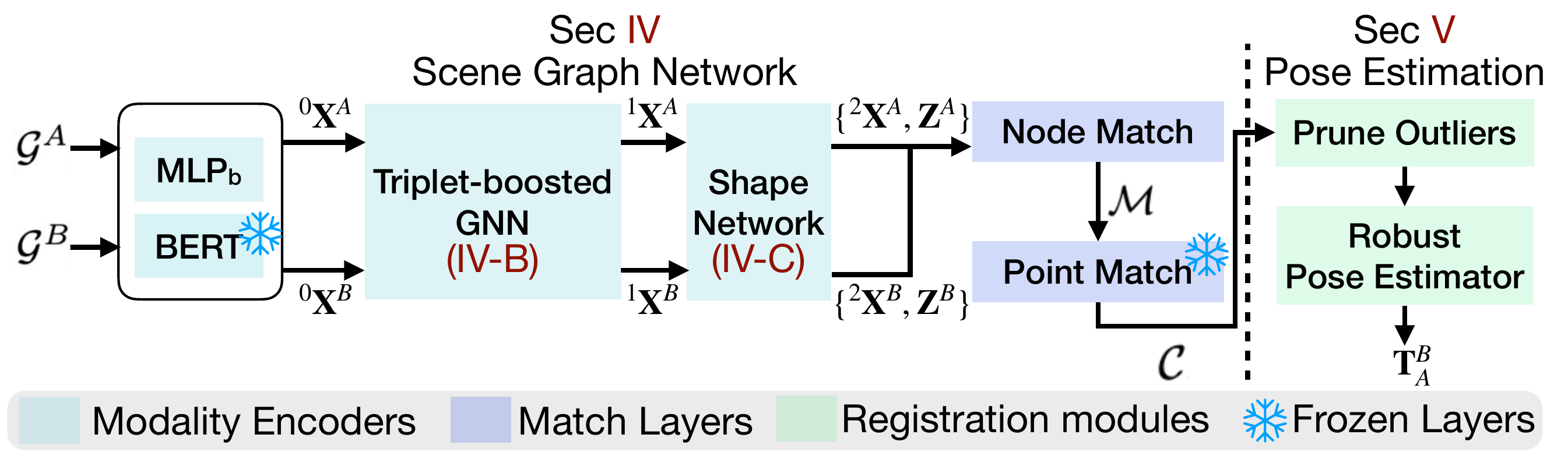}\vspace{-0.1cm}
    \caption{\revise{Our system overview. We denote the encoded node features as ${}^l\mathbf{X}^{A/B}$, where its layer index $l \in \{0,1,2\}$.}}\label{fig:pipeline}
    \vspace{-0.3cm}
\end{figure}

As shown in Fig. \ref{fig:pipeline}, SG-Reg is composed of a scene graph network and a robust pose estimator. The scene graph network reads a pair of semantic scene graphs, $\{\mathcal{G}^A,\mathcal{G}^B\}$, as input and produces node matches $\mathcal{M}$ along with point correspondences $\mathcal{C}$. Using these hierarchical matches, the pose estimator calculates a relative transformation $\mathbf{T}^B_A$ between the two scene graphs. Below, we summarize the scene representations used in our approach.

\newcommand*{\vargraph}{\ensuremath{\mathcal{G}}}
\newcommand*{\varbr}{\ensuremath{\mathbb{R}}}
\newcommand*{\varcp}{\ensuremath{\mathcal{P}}}
\newcommand*{\varcq}{\ensuremath{\mathcal{Q}}}
\newcommand*{\varcc}{\ensuremath{\mathcal{C}}}
\newcommand*{\varffeat}{\ensuremath{\mathbf{F}}}
\newcommand*{\varpxx}{\ensuremath{\tilde{\mathbf{X}}}}
\newcommand*{\varpmM}{\ensuremath{\mathcal{M}}}
\newcommand*{\varzZ}{\ensuremath{\mathbf{Z}}}
\newcommand*{\varvV}{\ensuremath{\mathcal{V}}}

\newcommand*{\varhn}{\ensuremath{\mathbf{h}}}
\newcommand*{\vartt}{\ensuremath{\mathbf{t}}}
\newcommand*{\varmsg}{\ensuremath{\mathbf{m}}}
\newcommand*{\varfeat}{\ensuremath{\mathbf{f}}}
\newcommand*{\varxx}{\ensuremath{\mathbf{x}}}
\newcommand*{\varxxX}{\ensuremath{\mathbf{X}}}
\newcommand*{\varyy}{\ensuremath{\mathcal{Y}}}
\newcommand*{\varpcd}{\ensuremath{\mathbf{P}}}
\newcommand*{\varbe}{\ensuremath{\mathbf{e}}}
\newcommand*{\varbo}{\ensuremath{\mathbf{o}}}
\newcommand*{\varbu}{\ensuremath{\mathbf{u}}}
\newcommand*{\varbv}{\ensuremath{\mathbf{v}}}
\newcommand*{\varz}{\ensuremath{\mathbf{z}}}
\newcommand*{\vardim}{\ensuremath{d}}
\newcommand*{\varaA}{\ensuremath{\mathcal{A}}}
\newcommand*{\varbB}{\ensuremath{\mathcal{B}}}
\newcommand*{\vartT}{\ensuremath{\mathbf{T}}}

\subsubsection{Explicit representation}
We denote a semantic scene graph as $\mathcal{G}=\{\mathcal{V, E}\}$, where $\mathcal{V}$ is the node set and $\mathcal{E}$ is the edge set.
A semantic node $\mathbf{v}_i$ has the following attributes.
\begin{itemize}
    \item ${s}$ is its open-set semantic label \revise{in text format}.
    \item $\mathbf{b} \in \varbr^3$ represents the bounding box length, width and height .
    \item $\varbo\in \varbr^3$ is its geometric center.
    \item $\varpcd$ is its point cloud.
\end{itemize}

\subsubsection{Implicit representation}
It incorporates multiple modalities. 
\begin{itemize}
    \item Shape feature ${\varfeat}_i \in \varbr^{\vardim_s}$, \revise{where $d_s$ is the dimension of the shape feature}.
    \item The node feature $^l\varxx_i$, where $l$ is the layer index. Before the shape network, the node feature is in the dimension $^{0,1}\varxx_i \in \varbr^d$. After the fusion with shape features, the final node feature is in the dimension $^2\varxx_i \in \varbr^{d+d_s}$. 
    \item Point features ${\varz}_i \in \varbr^{K_p \times {d}_z}$, where $K_p$ is the number of sampled points and \revise{$d_z$ is the dimension of the point feature}.
    \item We use small symbols to denote features from one node and capital symbols to denote features from a scene graph. For example, $^2\varxx_i$ is a node feature, and $^2\varxxX^A$ is the stacked node feature from $\vargraph^A$.
\end{itemize}

\subsubsection{Node index}
For a pair of scene graphs $\{\vargraph^A, \vargraph^B\}$, we denote their node indices $\mathcal{A}=[1,...,|\mathcal{V}^A|]$ and $\mathcal{B}=[1,...,|\mathcal{V}^B|]$.

\subsubsection{Data association}
The scene graph network generates node matches $\mathcal{M}$ and point matches $\mathcal{C}$. 
\begin{itemize}
    \item A node match $m=\{i,j,\omega_n\}$, where $m \in \mathcal{M}$, consists of the associated node indices $i\in \varaA, j\in \varbB$ and the node confidence $\omega_n \in [0,1]$. 
    \item A global set of point correspondences $\varcc=\{(\mathbf{p}_k,\mathbf{q}_k)\}$, where $\mathbf{p}_{k},\mathbf{q}_k \in \varbr^3$.
\end{itemize}

\subsubsection{Transformation}. The relative transformation from $\mathcal{G}^A$ to $\mathcal{G}^B$ is denoted as $\mathbf{T}^B_A \in \text{SE}(3)$.

\section{Scene Graph Network}\label{sec:sgnet}

\subsection{Semantic Scene Graph Construction}
\begin{figure}[ht]
    \centering
    \vspace{-0.3cm}\includegraphics[width=0.95\columnwidth]{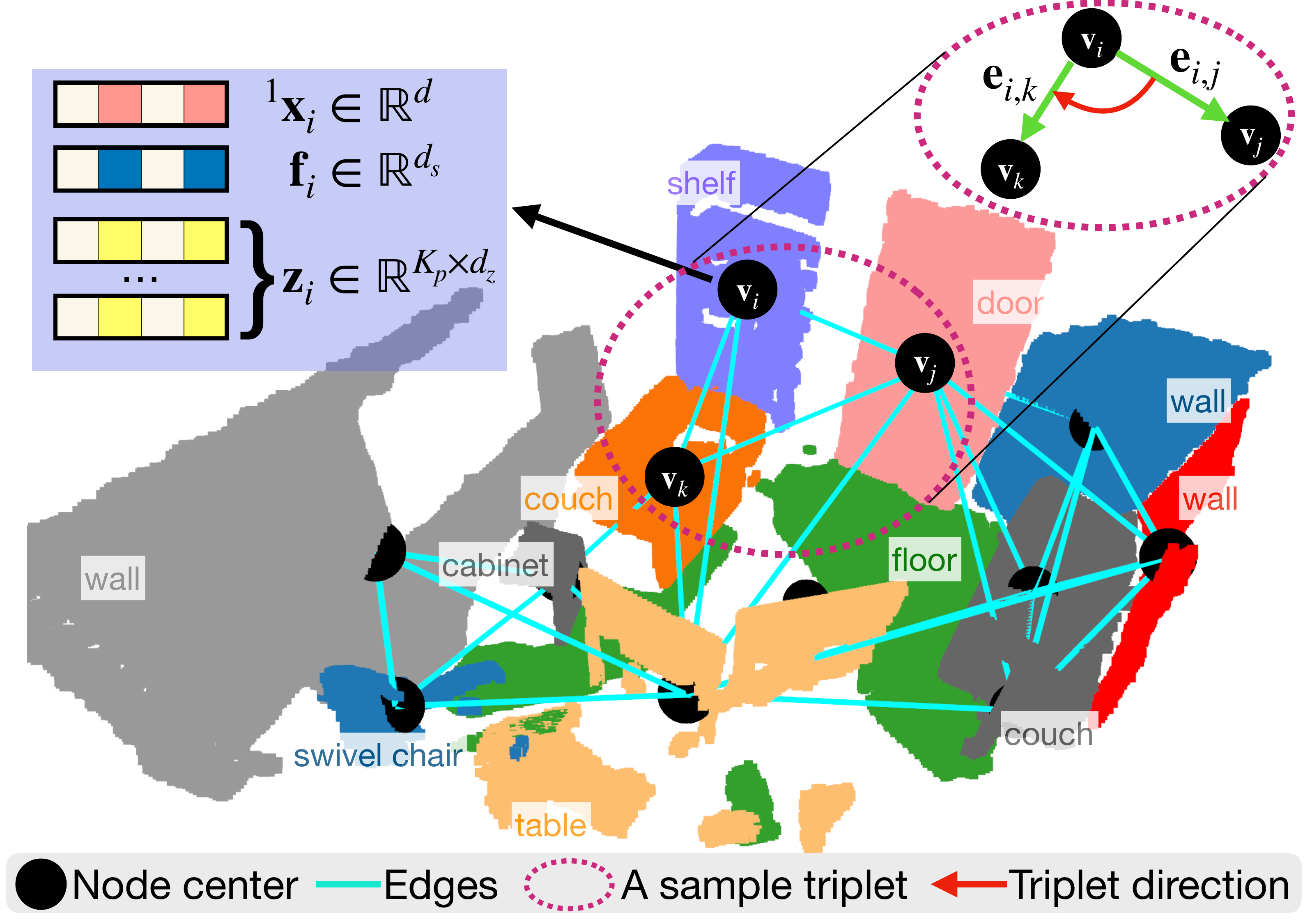} 
    \vspace{-0.2cm}
    \caption{Visualization of a semantic scene graph from ScanNet \textit{scene0025\_00}. \revise{Each node's point cloud is distinctly colored. For node $\mathbf{v}_i$, we illustrate one of its triplet. Additionally, the implicit features derived from $\mathbf{v_i}$ are displayed.}}
    \label{fig:explicit_sg}\vspace{-0.2cm}
\end{figure}

The semantic scene graph is constructed using FM-Fusion\cite{liu2024fmfusion}. 
We extract its semantic instances and construct semantic nodes $\mathcal{V}$ in the scene graph. Those instances that are too small or have insufficient observations are excluded. 
For a node $\varbv$, its semantic label ${s}$ comes from the prediction from FM-Fusion. \revise{The node can be observed multiple times by GroundingDINO, and FM-Fusion fuses multiple label measurements to predict a final label $s$.} The point cloud $\varpcd$ is extracted from the instance-wise submap via 3D interpolation. The interpolation process is similar to marching cube\cite{lorensen1987marching}, but it only interpolates points that have three neighboring voxels. We extract the node center $\varbo$ from $\varpcd$. Then, we calculate the minimum bounding box from $\varpcd$ using O'Rourke's algorithm\cite{orouke1985minbox}. The bounding box shape $\mathbf{b} \in \varbr^3$ is recorded, representing its length, width, and height. 
We ignore the orientation from each semantic node to minimum additional noise. Lastly, a semantic node is constructed $\mathbf{v}=\{s,\mathbf{b,o},\varpcd\}$. 
We construct the representations for other nodes and generate $\mathcal{V}$.

Once the nodes set $\mathcal{V}$ is constructed, we compute the distance between each pair of nodes. The distance can be calculated from the node's center. If their distance is less than a threshold $\tau_d$, we connect an edge between them. The distance threshold $\tau_d$ is decided by node's shape size, meaning that larger nodes connect to more distant nodes and a greater number of nodes. We record the generated edge set $\mathcal{E}$. As shown in Fig. \ref{fig:explicit_sg}, we visualize an example of scene graph $\vargraph=\{\mathcal{V,E}\}$.

The input semantic scene graphs $\{\vargraph^A,\vargraph^B\}$ have a relative transformation in 4-DoF, including 3D position and relative yaw rotation. Relative rotation in roll and pitch angle is already estimated by visual-inertial SLAM\cite{qin2018vins} accurately.

\subsection{Triplet-boosted Graph Neural Network (GNN)}
Firstly, we initialize the input features for the GNN module. We run BERT\cite{devlin2018bert} to encode semantic labels and run a single-layer MLP to encode bounding box size. A node $\mathbf{v}_i \in \mathcal{V}^A$ has its node feature initialized as below.
\begin{equation}
    ^0\varxx_i = [\text{BERT}({s}_i) \Vert \text{MLP}_b(\mathbf{b}_i)]
\end{equation}
where $^0\varxx_i \in \varbr^d$ and $[\cdot \Vert \cdot]$ denotes a concatenation.

\newcommand*{\funcsin}{\ensuremath{\psi}} 

The current attentional GNN incorporates a relative position encoding (RPE) in its self-attention aggregation \cite{lindenberger2023lightglue}.
\begin{equation}
    \text{RPE}(\varbo_i,\varbo_j) = \Phi^R (\varbo_i - \varbo_j)
\end{equation}
where $\varbo_{i/j}$ are the nodes centers and $\Phi^R$ is a rotary encoding. RPE has been shown to enhance learning in vision tasks \cite{lindenberger2023lightglue}\cite{zhang2022dino} and language tasks \cite{devlin2018bert}. \revise{However, RPE exhibits a critical limitation in 3D geometric tasks: it is variant to yaw-angle rotations. Specifically, when the input graphs are transformed by a yaw-angle rotation, RPE fails to produce rotation-invariant features, which are essential for robust matching in such scenarios.}

To address the challenge, we design a triplet descriptor to enhance GNN learning. The triplet descriptor is designed to be invariant to a global transformation in 4-DoF. As shown in Fig. \ref{fig:explicit_sg}, we set a semantic node $\varbv_i \in \mathcal{V}^A$ as the anchor of a triplet and randomly sample two of its neighbor nodes $\{\varbv_j,\varbv_k\}$ to be the corners of the triplet. The triplet feature is $\vartt^i_{j,k}$. In computing $\mathbf{t}^i_{j,k}$, we keep the corner nodes $\varbv_j,\varbv_k$ in an anti-clockwise order along the z-axis in a visual SLAM coordinate\cite{qin2018vins}. 

Mathematically, the triplet feature is defined as follows:
\begin{equation}
    \vartt^i_{j,k}=
    \begin{cases}
        \vartt^i_{[j,k]} & \text{if } (\varbe_{i,j} \times \varbe_{i,k})_z \geq 0\\
        \vartt^i_{[k,j]} & \text{if }(\varbe_{i,j} \times \varbe_{i,k})_z < 0
    \end{cases}
    \label{eq:triplet}
\end{equation}
where the relative position vector ${\varbe}_{i,j} = \varbo_j-\varbo_i$, and $(\cdot \times \cdot)_z$ is the z-axis value after a cross product.
Then, an ordered triplet feature is initialized.
\begin{equation}
    \vartt^i_{[j,k]}=[^0\varxx_j \Vert ^0\varxx_k \Vert \mathbf{g}^i_{j,k}]
\end{equation}
The $\mathbf{g}^i_{j,k}$ is a geometric embedding.
\begin{equation}\label{eq:geoemb}
    \mathbf{g}^i_{j,k} = [\funcsin^L(\vert \varbe_{i,j}\vert) \Vert \funcsin^L(|\varbe_{i,k}|)\Vert \funcsin^A(\hat{\varbe}_{i,j} \cdot \hat{\varbe}_{i,k})]
\end{equation}
where the length of triplet edges $\vert\varbe_{i,j}\vert,\vert \varbe_{i,k} \vert \in \varbr$ and $\hat{\varbe}_{i,j}, \hat{\varbe}_{i,k}$ are the normalized vectors. The term $(\hat{\varbe}_{i,j} \cdot \hat{\varbe}_{i,k})$ represents cosine value of the triplet angle.
Sinusoidal functions $\funcsin^L$ and $\funcsin^A$ are applied to embed the edge length and \revise{the triplet angle}. \revise{The sinusoidal functions have been used in GeoTransformer\cite{qin2022geometric} to encode their geometric scalars.}

Now that we have illustrated the creation of triplet features, we explain how to integrate them into a GNN. \revise{The triplet-boosted GNN reads the initialized node features ${^0}\varxxX^A$ and outputs them as ${}^1\varxxX^A$. The residual message passing update for all $i$ in $A$ is:}
\begin{equation}\label{eq:gnn}
    ^1\varxx_i = {}^0\varxx_i+\text{MLP}[^0\varxx_i \Vert \varmsg_i].
\end{equation}
To \revise{compute the message feature $\varmsg_i$, we first sample a set of triplets from its associated edge set $\mathcal{E}(i,\cdot)$ The feature is then computed using an attention mechanism \cite{vaswani2017attention}, which aggregates the embeddings of all sampled triplets.}
\begin{equation}
    \varmsg_i = \sum_{(j,k) \in \mathcal{E}(i,\cdot),j \neq k}\alpha^i_{j,k} \Big(\mathbf{W}^V(\vartt^i_{j,k})\Big)\\
\end{equation}
\revise{The attention score $\alpha^i_{j,k}$ is computed by a softmax over all of the query-key similarities:}
\begin{equation}
    \alpha^i_{j,k} =\text{softmax}_i \Bigl(\mathbf{W}^Q(\vartt^i_{j,k})\big(\mathbf{W}^K(\vartt^i_{j,k})\big)^T \Bigl),
\end{equation}
where $\mathbf{W}^K,\mathbf{W}^Q,\mathbf{W}^V$ are projection encoders before the attention layer. 
\revise{This step leverages the attention mechanism to aggregate relevant triplets associated with each node.}

We follow equation (\ref{eq:gnn}) to encode all the nodes in $\vargraph^A$ and $\vargraph^B$. \revise{The outputs after the triplet-boosted GNN are:}
\begin{equation}
\{^1\varxxX^A,{}^1\varxxX^B \big\vert {}^1\varxxX^A \in \varbr^{ |\varaA| \times d}, {}^1\varxxX^B \in \varbr^{|\varbB| \times d} \}.
\end{equation}

Compared to the vanilla graph attention network\cite{brody2021gatv2}, the triplet boosts the spatial awareness of the GNN. Compared to the GNN incorporated with RPE\cite{lindenberger2023lightglue}, our approach ensures that the features remain invariant to yaw rotation.
Additionally, our process aligns more closely with human intuition. Humans recognize a room layout by describing its object topology with spatial awareness. For example, a human may refer to a couch in his office facing a television and having a table on its left-hand side. Our triplet descriptor captures local topology with spatial awareness, mimicking human intuition.
In the SLAM domain, similar triplet descriptors \cite{jiang2019triangle, yuan2023std,zou2024lta,yuan2024btc} have been proposed, although they represent the triplet explicitly. We believe triplet is a relatively stable local structure in the cross-domain scenario we target. Even under noisy scene graphs, if a node feature aggregates one or two correct triplets, it is still highly likely to find the correct match. 

\subsection{Shape Network}\label{sec-shape}
\newcommand*{\varpPcd}{\ensuremath{\mathcal{P}}}
\newcommand*{\varqPcd}{\ensuremath{\mathcal{Q}}}
\newcommand*{\varsS}{\ensuremath{\mathcal{S}}}
\newcommand*{\varhH}{\ensuremath{\mathbf{H}}}
\newcommand*{\pcdA}{\ensuremath{\mathcal{X}}}
\newcommand*{\pcdB}{\ensuremath{\mathcal{Y}}}

The shape network generates geometric features at two levels. It includes a point backbone that learns point features and a shape backbone that learns the node-wise shape features.

Given a pair of scene graphs $\{\vargraph^A,\vargraph^B\}$, we extract their global point cloud $\{\pcdA,\pcdB\}$, where each point is annotated with its parent node index. We downsample them into four resolutions $\{{}^l\pcdA,{}^l\pcdB\}_l$, where the layer index $l=\{0,1,2,3\}$. 
The parent node index is maintained during down-sampling. 

\begin{figure}[ht]
    \centering
    \vspace{-0.2cm}\includegraphics[width=\columnwidth]{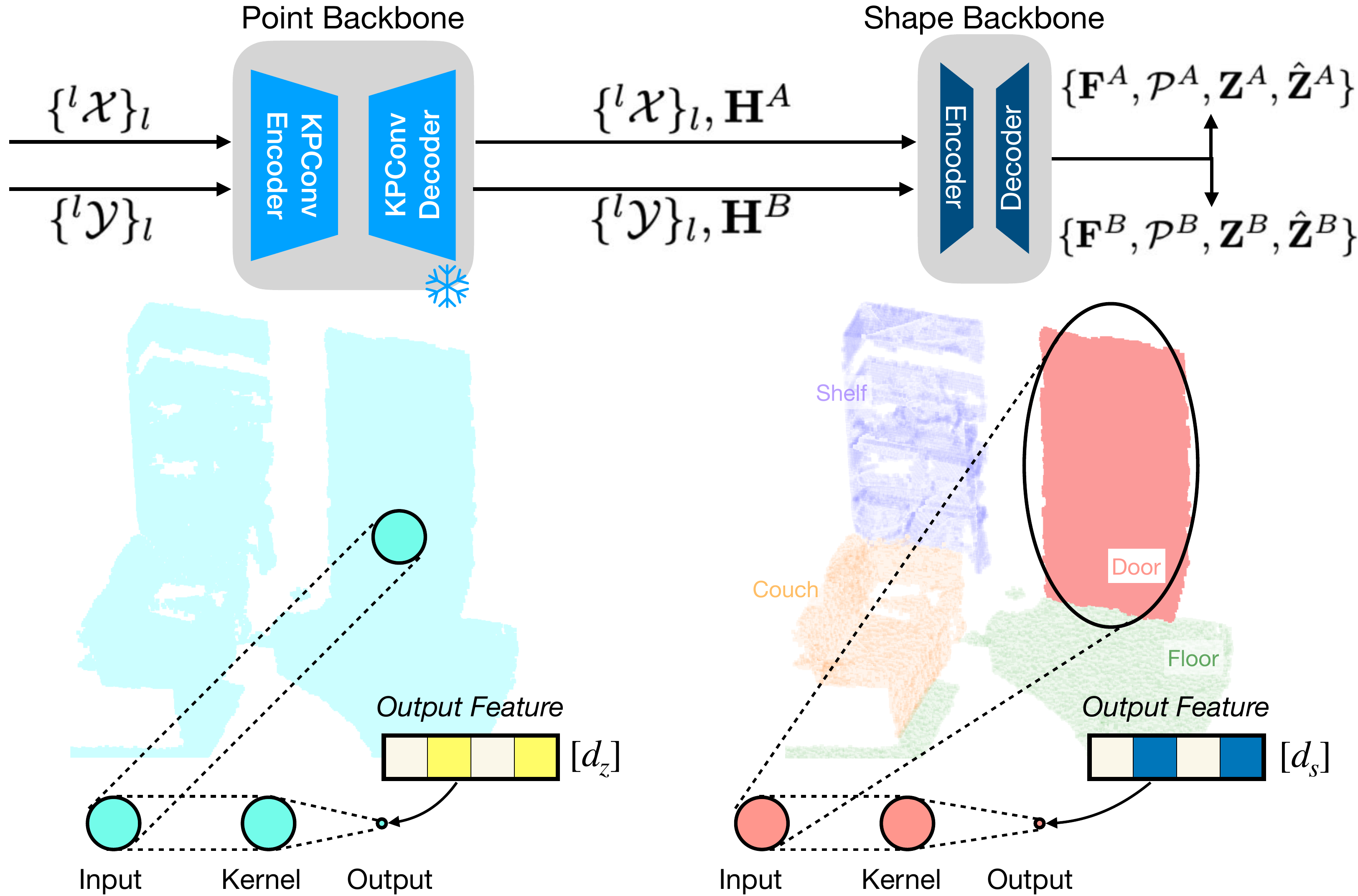}
    \caption{Visualization of the shape network structure and its point aggregation kernels. Point backbone uses grid sub-sampling to decide aggregation kernels, which are small and dense. Shape backbone following instance segmentation to create aggregation kernels, which are large and sparse. 
    }
    \label{fig:shape-encoder}\vspace{-0.3cm}
\end{figure}


In the following steps, we explain the encoding for $\{{}^l\pcdA\}_l$ and it goes the same way for $\{{}^l\pcdB\}_l$.
As shown in Fig. \ref{fig:shape-encoder}, the point backbone generates point features in multiple layers. We only take the second finest level of point features as a hidden state feature $\varhH^A$. We did not take the finest level of point cloud features, because they are redundant, as noted in previous work\cite{qin2022geometric}. Utilizing the node indices maintained in ${}^{1\!}\pcdA$, we can directly sample the node-wise points from $\varhH^A$. 
During the sampling, we only keep $K_p$ points for each node. If a node finds too few or too many points, we adjust the number to $K_p$ through random sampling or zero padding. 
The sampled node-wise points and features are $\varpPcd^A$ and $\varzZ^{A}$. 

The shape backbone has one layer of encoder and decoder. It aggregates point features following the annotated node indices in ${}^{1\!}\pcdA$ and generates node-wise shape features $\varffeat^A$. Its decoder generates decoded point features $\hat{\varzZ}^{A}$, allowing dense supervision of the shape backbone during training.

We summarize the output representations of $\vargraph^A$ as below. 
\begin{itemize}
    \item Node shape feature $\varffeat^A \in \varbr^{|\varaA|  \times d_s}$. 
    \item Node points $\varpPcd^A \in \varbr^{|\varaA| \times K_p \times 3}$. 
    \item Node points feature ${\varzZ}^A\in \varbr^{|\varaA| \times K_p\times{d}_z}$. 
    \item Decoded points feature $\hat{\varzZ}^A \in \varbr^{|\varaA| \times K_p \times {d}_z}$. 
\end{itemize}

At the last step of the shape network, we fuse the shape features into the node features by concatenation.
\begin{align*}
    ^2\varxxX^A = [{}^1\varxxX^A \Vert \varffeat^A ], {}^2\varxxX^A \in \varbr^{|\varaA| \times(d+d_s)}
\end{align*}
Notice that the shape fusion can run before the triplet-GNN or after the triplet-GNN. We investigate the influence of early fusion and late fusion in our experiment in Sec \ref{sec:eval_shape}.

Some previous PCR works\cite{qin2022geometric}\cite{wang2019dcp} also employ a point cloud backbone to extract point cloud features. Our method's uniqueness stems from the addtion of a shape backbone after the point backbone. This approach offers two key advantages. Firstly, it reduces the density of the output features from the point backbone. In the previous point cloud learning methods\cite{wang2019dcp,qin2022geometric}, they run multiple attention layers on the hidden state features after the point backbone. Due to the densely distributed hidden state features, their attention layers consume a huge GPU memory and can only run on small-scale scenes. 
Secondly, \revise{the shape backbone generates a shape feature to represent the geometric attribute of each semantic node, enabling the direct fusion of shape features with other modality features to create a compact semantic node feature.}

\subsection{Hierarchical graph match}\label{sec-gmatch}
With the learned node features and their point features, we can run hierarchical graph matching layers to search for correspondences. As shown in Table. \ref{tab:assign}, we summarize the node assignment matrix $\mathbf{A}^X$ and point assignment matrix $\mathbf{A}^Z$. The decoded point assignment matrix $\hat{\mathbf{A}}^Z$ is used to supervise the shape network training and is skipped in the inference.

\begin{table}[ht]
    \centering
    \begin{tabular}{c c c c}
    \toprule 
    &  Input & Dimensions & Inf.\\
     \hline
    Node Assignment $\mathbf{A}^X$ & $\{{}^{l}\varxxX^A,{}^{l}\varxxX^B\}$ & $|\varaA| \times |\varbB|$ & \checkmark\\
    Point Assignment $\mathbf{A}^Z$ & $\{\varpmM,\varzZ^A,\varzZ^B\}$ & $|\varpmM|\times K_p \times K_p$ & \checkmark\\
    Point Assignment $\hat{\mathbf{A}}^Z$ & $\{\bar{\varpmM},\hat{\varzZ}^A,\hat{\varzZ}^B\}$ & $|\bar{\varpmM}|\times K_p \times K_p$& $\times$\\
    \bottomrule
    \end{tabular}
    \caption{Assignment matrices dimensions. The Inf. refers to the assignment computed at the inference stage. $\bar{\mathcal{M}}$ refers to the ground-truth node matching during the training.}\label{tab:assign}\vspace{-0.3cm}
\end{table}

\newcommand*{\varsim}{\ensuremath{\mathbf{S}}}
\newcommand*{\varassign}{\ensuremath{\mathbf{A}}}
\newcommand*{\varbp}{\ensuremath{\mathbf{p}}}
\newcommand*{\varbq}{\ensuremath{\mathbf{q}}}

Next, we illustrate the construction of the assignment matrix at each level. In previous work, SuperGlue\cite{sarlin20superglue} formulates the graph matching problem as a differential partial assignment problem. It determines the optimal match by the Sinkhorn algorithm\cite{Cuturi2013sinkhorn}. We use similar techniques in graph matching.

At the node match layer, we calculate the node similarity matrix and \textbf{node assignment matrix} as below.
\begin{equation}
    \varsim_{i,j}^X = \text{Linear}(\varxx_i)^T \text{Linear}(\varxx_j), \forall(i,j)\in \varaA \times \varbB
\end{equation}
\begin{equation}\label{eq:dualnorm}
    \varassign_{i,j}^X =\mathop{\text{Softmax}}\limits_{(k\in \varaA)}\big(\mathbf{S}^X_{k,j}\big)_k \mathop{\text{Softmax}}\limits_{k\in \varbB}\big(\mathbf{S}^X_{i,k}\big)_k
\end{equation}
We take dual normalization in computing $\mathbf{A}^X$ to suppress the negative match pairs.
If an assignment score $\varassign_{i,j}^X$ is higher than a threshold and it is a mutual top-$k$ score in $\varassign^X$, we extract it as a matched node pair $m=\{i,j,\omega\}$, where match confidence $\omega = \varassign_{i,j}^X$. Thus, we predict a set of node matches $\mathcal{M}$.

Guided by the node matches $\mathcal{M}$, point matching layer searches for the point correspondences $\mathcal{C}$. Regarding a matched node pair $(i,j)$, its point similarity matrix is calculated as follows,
\begin{equation}\label{eq:dotprod}
    {\mathbf{S}}^Z_{i,j} = ({\varz}^A_i)^T {\varz}^B_j, {\mathbf{S}}^Z_{i,j} \in \varbr^{K_p \times K_p}
\end{equation}
Then, we apply optimal transport algorithm \cite{Cuturi2013sinkhorn} on $\mathbf{S}^Z_{i,j}$ to compute the \textbf{point assignment matrix} ${\varassign}^Z_{i,j} \in \varbr^{K_p \times K_p}$. \revise{The optimal transport algorithm can be seen as a differential version of the Hungarian algorithm\cite{munkres1957hungarian}. It has been applied to solve the bipartite matching problem in SuperGlue\cite{sarlin20superglue} and GeoTransformer\cite{qin2022geometric}.} 

We select the mutual top-$k$ pairs in $\mathbf{A}^Z_{i,j}$ as the point correspondences $\varcc_{i,j}$.
\begin{equation}
    \varcc_{i,j} = \{\varbp_k,\varbq_k|\varbp_k \in \varpPcd_i^A,\varbq_k \in \varpPcd_j^B, k\in \text{mtop-}k(\mathbf{A}^Z_{i,j}) \}
\end{equation}
where $\{\varpPcd_i^A,\varpPcd_j^B\}$ are the node-wise points as illustrated in Sec. \ref{sec-shape}.
Finally, we search point correspondences in all the matched nodes $\varpmM$ and construct a global set of point correspondences $\varcc=\{(\varbp_k,\varbq_k)\}$.


\subsection{Training}
\subsubsection{Loss function}
\newcommand*{\varpos}{\ensuremath{\bar{\mathcal{M}}}}
\newcommand*{\varneg}{\ensuremath{\mathcal{N}}}
We train the network as below.
\begin{equation}
    \mathcal{L} = \mathcal{L}_{gnn} + \mathcal{L}_{shape} 
\end{equation}

$\mathcal{L}_{gnn}$ is proposed to supervise the triplet-GNN and the node match layers, while $\mathcal{L}_{shape}$ is proposed to supervise the shape backbone. Specifically, 
\begin{equation}\label{eq:gnnloss}
    \mathcal{L}_{gnn}=\frac{1}{2}\sum_{l=1,2}\sum_{(i,j)\in \Bar{\mathcal{M}}}{\log \big( \prescript{l\!\!}{}{\varassign}^X_{i,j}} \big)\\
\end{equation}
where $\Bar{\mathcal{M}}$ is ground-truth node matches and $l$ is the layer index of the node features. Since the dual normalization in Equation (\ref{eq:dualnorm}) already suppresses the negative pairs, $\mathcal{L}_{gnn}$ has implicitly involved a negative loss term. Thus, we omit the loss penalty regarding the negative match pairs. 

The shape loss is constructed by designing a contrastive loss \cite{xie2020pointcontrast} on shape features and an optimal transport loss \cite{qin2022geometric} on the decoded point features $\{\hat{\varfeat}^A_i,\hat{\varfeat}^B_j\}$.
\begin{align}\label{eq:shape}
    \begin{split}
    \mathcal{L}_{shape} = \frac{1}{2\vert \bar{\mathcal{M}} \vert} \sum_{(i,j) \in \varpos} \big( &\mathcal{L}_{ot}(\hat{\varz}^A_i,\hat{\varz}^B_j) \\+\mathcal{L}_{cont}(\varfeat^A_j,\varfeat^B_i) +&\mathcal{L}_{cont}(\varfeat^A_i,\varfeat^B_j)\big)
    \end{split}
\end{align}
where $\varpos$ is the ground-truth node pairs. We explain the contrast loss and optimal transport loss terms below.
\begin{align*}    
    \mathcal{L}_{cont}(\varfeat^A_i,\varfeat^B_j) &= \frac{\exp(\varfeat^A_i \cdot \varfeat^B_j)}{\sum_{k \in \varneg_i}\exp(\varfeat^A_i \cdot \varfeat^B_k)}\\
    \mathcal{L}_{ot}(\hat{\varz}^A_i,\hat{\varz}^B_j) &= -\sum_{(u,v)\in \bar{\mathcal{C}}_{i,j}} \log \hat{\mathbf{A}}^Z_{i,j}(u,v) \\ 
    &- \sum_{u \in {\varneg}_i^Z} \log \hat{\mathbf{A}}^Z_{i,j}(u,K_p)-\sum_{v\in {\varneg}_j^Z}\log \hat{\mathbf{A}}^Z_{i,j}(K_p,v) 
\end{align*}
where $\varneg_i$ is the set of negative nodes compared with $\mathbf{v}_i$, $\bar{\mathcal{C}}_{i,j}$ is a set of the ground-truth point matches between $\{\mathbf{v}_i,\mathbf{v}_j\}$, and ${\varneg}_{i}^Z$ is the set of unmatched points in $\varpPcd_i$. 
\revise{The loss term $\mathcal{L}_{ot}$ is designed to supervise the point matching between two associated semantic nodes. It pulls the features of the matched points closer, while pushing the features of the unmatched points further apart.}

\subsubsection{Generate ground-truth match}
The ground-truth node match set $\bar{\mathcal{M}}$ and each set of negative nodes $\mathcal{N}_i$ are selected by calculating the intersection between node pairs. We compute the intersection-over-unit (IoU) between the nodes' point clouds to decide an intersection. Ground-truth point matches $\mathcal{N}_i^Z$ are those points with their relative distance less than a threshold (\textit{i.e.} $0.05m$).

\section{Robust Pose Estimator}\label{sec-pestimator}

Given a set of correspondences $\mathcal{C}=\{(\varbp_k, \varbq_k)\}$, we follow the G3Reg framework \cite{qiao2024g3reg}, which employs a distrust-and-verify approach, for robust transformation estimation. G3Reg generates multiple transformation hypotheses and uses a geometric verification function to select the optimal one. 

In the hypothesis proposal step, a pyramid compatibility graph is constructed. Specifically, for two pairs of correspondences $(\varbp_i,\varbq_i)$
 and $(\varbp_j,\varbq_j)$, we test their compatibility:
\begin{equation}
\label{eq:comp_test}
    \left|\left\|\varbp_i-\varbp_j\right\|_2-\left\|\varbq_i-\varbq_j\right\|_2\right|<\delta_{ij}
\end{equation}

Here, $\delta_{ij}$ is a threshold that helps reject potential outliers in the correspondence set $\mathcal{C}$.

By gradually increasing $\delta_{ij}$, we obtain a pyramid graph that becomes denser at each level. For every level, a maximum clique (MAC) is identified using the graduated maximum clique solver from G3Reg \cite{qiao2024g3reg}. The correspondences in this clique form a potential inlier set $\mathcal{C}^*$. A candidate transformation $(\mathbf{R}^*, \mathbf{t}^*)$ is then determined as follows:
\begin{subequations}
\label{eq:reg}
\begin{align}
\mathbf{R}^*, \mathbf{t}^* &= \underset{\mathbf{R} \in \operatorname{SO}(3), \mathbf{t} \in \mathbb{R}^3}{\arg \min } \sum_{\left(\varbp_k, \varbq_k\right) \in \mathcal{C}^*} \min \left(r\left(\varbq_k, \varbp_k\right), \bar{c}^2\right) \label{eq:1a} \\
d_{k} &= \varbq_k - \left(\mathbf{R} \varbp_k + \mathbf{t}\right) \label{eq:1b} \\
r\left(\varbq_k, \varbp_k\right) &= 
\begin{cases} 
d_{k}^{\mathrm{T}}\left(\Sigma_{y}^k + \mathbf{R} \Sigma_{x}^k {\mathbf{R}}^{\mathrm{T}} \right)^{-1} d_{k}, & \text{if instance centers} \\
d_{k}^{\mathrm{T}} d_{k}, & \text{otherwise}
\end{cases} \label{eq:1c}
\end{align}
\end{subequations}
where $\bar{c}$ is the inlier cost threshold, and $\Sigma$ is the statistical covariance matrix of the instance point cloud. This optimization problem can be solved using GNC \cite{yang2020gnc}.

We then introduce an evaluation function $g$ to determine the most suitable transformation based on the geometric information of the original point clouds $\mathcal{X}$ and $\mathcal{Y}$:
\begin{equation}
\label{eq:eval}
\tilde{\mathbf{R}}, \tilde{\mathbf{t}} = \underset{\tilde{\mathbf{R}} \in \left\{\mathbf{R}_{m}^{*}\right\}, \tilde{\mathbf{t}} \in \left\{\mathbf{t}_{m}^{*}\right\}}{\arg \min } g\left(\mathbf{R}_{m}^{*}, \mathbf{t}_{m}^{*} \mid \mathcal{X}, \mathcal{Y} \right)
\end{equation}
\revise{where $m$ is the index of a candidate transformation. }

The function $g$ is designed based on the geometric distribution of the voxelized point cloud, utilizing point-to-plane and point-to-point distances. More details are available in \cite{qiao2024g3reg}. The predicted transformation $\vartT^B_A$ is created from the optimal candidate $(\mathbf{\tilde{R},\tilde{t}})$.

{While the above framework demonstrates impressive performance for registration with very low inlier ratios, its efficiency and robustness may be compromised when dealing with higher inlier ratios and repetitive correspondence patterns. To address these limitations, we introduce two practical strategies aiming at enhancing both the efficiency and the robustness of the system.}

Firstly, \revise{while maximum clique pruning\cite{bustos2019practical} demonstrates superior robustness at low inlier ratios, its computational cost grows linearly with inlier ratio in practice. In contrast, GNC provides comparable accuracy but faster execution at high inlier ratios. Our hybrid approach therefore activates MAC only when the GNC-estimated inlier ratio falls below a certain value, such as 0.3. Otherwise, we take the prediction from GNC as the candidate transformation. The strategy achieves optimal speed-accuracy tradeoff.}
{Secondly, repetitive structure tends to create dense and intra-consistent point correspondences, which may construct a large outlier clique and disable the maximum clique inlier selection strategy\cite{yang2020teaser}. To mitigate this, we apply Non-Maximum Suppression (NMS) to the correspondences. For correspondences \((\varbp_i, \varbq_i)\) and \((\varbp_j, \varbq_j)\), if \(\|\varbp_i - \varbp_j\|_2^2\) is less than a predefined threshold, we retain only the correspondence with the higher score. }

The outlier pruning and the robust pose estimator do not suffer from the generalization issues. They guarantee the registration performance in cross-domain evaluation, which may hold a high outlier ratio.

\vspace{-0.3cm}

\section{Two-agent SLAM}\label{sec-multiagent}
\begin{figure}[ht]
    \centering
    \vspace{-0.3cm}\includegraphics[width=\columnwidth]{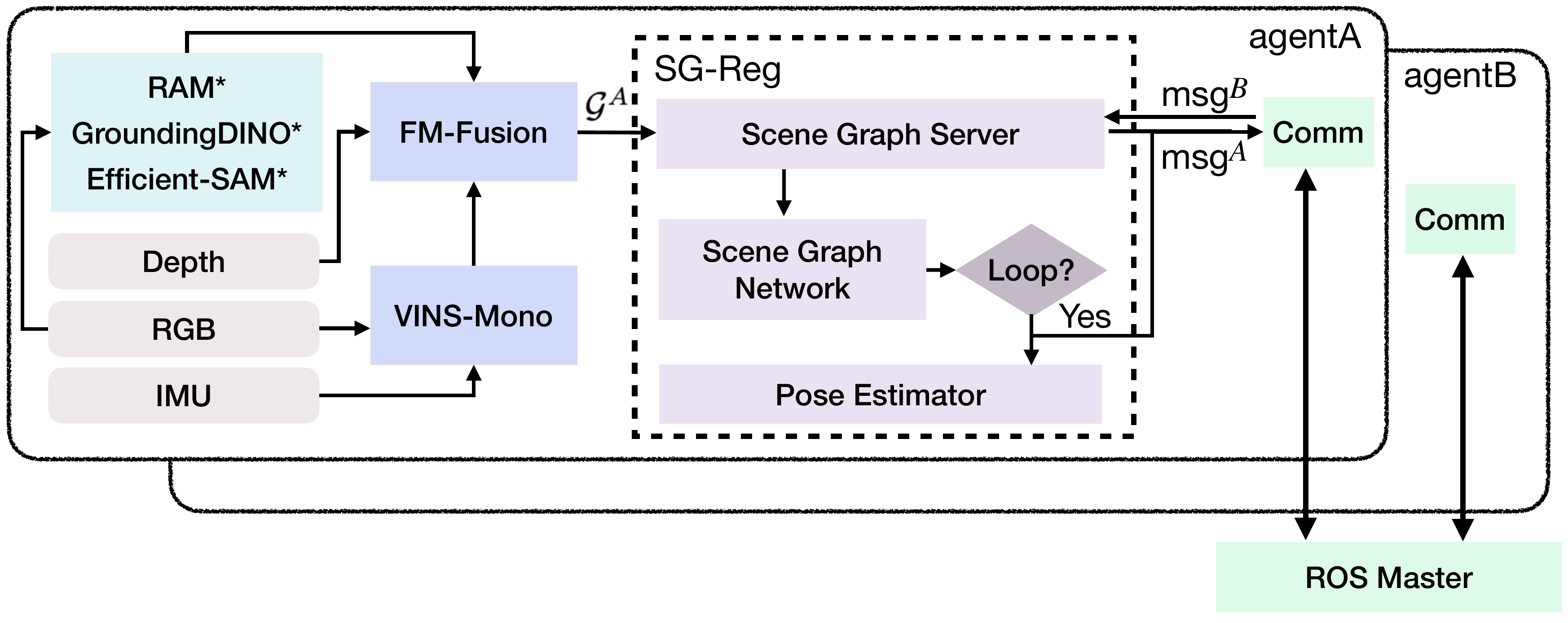}
    \caption{Two agent SLAM system structure. Module marked with * runs in offline.}\label{fig:multi-agent}\vspace{-0.5cm}
\end{figure}
To demonstrate how scene graph registration helps SLAM tasks, we integrate SG-Reg into a two-agent SLAM system to detect loop closures and register the scene graphs. The structure of the implemented system is illustrated in Fig. \ref{fig:multi-agent}.
\subsubsection{Coarse-to-fine communication}
\revise{We introduce a coarse-to-fine communication strategy to achieve optimal performance that balances accuracy and communication bandwidth. Initially, each agent broadcasts its \textbf{coarse message} at a steady rate (i.e. 1 Hz), allowing SG-Reg to match semantic nodes. At a query frame, if a minimum number of nodes are matched, the agent sends a request message to its remote agent. In response, the remote agent publishes a \textbf{dense message}, enabling the agent to execute the complete registration pipeline shown in Fig. \ref{fig:pipeline}.
To prevent the dense messages from being published too frequently, we establish a minimum interval between the query frames that allow for sending request messages. Below, we highlight the format of the broadcast messages from agent-A:}
\begin{itemize}
    \item Coarse message: $\{\varxxX^A,\mathbf{O}^A\}$.
    \item Dense message: $\{\varxxX^A,\mathbf{O}^A,\pcdA\}$.
\end{itemize}
where node features $\varxxX^A \in |\varaA| \times d$, node center points $\mathbf{O}^A \in |\varaA| \times 3$, and the stacked point cloud $\pcdA$ is as illustrated in Sec-\ref{sec-shape}.


To enhance the performance of receiving a coarse messages, we maintain the latest point correspondences $\tilde{\varcc}$ in the program. Upon receiving a coarse message, we merge the matched nodes' center with $\tilde{\varcc}$ and construct the final correspondence set $\varcc$, if $\tilde{\varcc}$ is available. Or we construct $\varcc$ only using the matched nodes' center. Each time reading a $\varcc$, SG-Reg predicts a global transformation $\vartT^B_A$ without an initial transformation. 

\subsubsection{SG-Reg deployment}
We implement the network modules in SG-Reg using LibTorch\footnote{https://pytorch.org/cppdocs/}, enabling us to call the neural network from a C++ executable program. We integrate SG-Reg with other SLAM modules. Each network block in SG-Reg is deployed separately, allowing the SLAM system to match semantic nodes with or without fusing shape features. 

\subsubsection{SLAM integration}
As shown in Fig. \ref{fig:multi-agent}, we use VINS-Mono\cite{qin2018vins} to compute visual-inertial odometry. RAM-Ground-SAM\cite{zhang2023ram,liu2023grounding,xiong2024efficientsam} processes the RGB sequences and save their prediction before the experiment.
FM-Fusion\cite{liu2024fmfusion} reads the prepared data from RAM-Ground-SAM and runs in real time.
\vspace{-0.3cm}

\section{Experiment}
We execute a series of experiments to assess our method. These experiments reveal the following advantages of our approach, which align with our declared novelties in Sec-\ref{sec-intro}:
\begin{itemize}
    \item \textit{Scene graph matching.} Improvement is achieved through the triplet-boosted GNN and shape integration discussed in \ref{sec:scannet}.
    \item \textit{Generalization.} Our method's registration capability is validated on a cross-domain dataset (Sec-\ref{sec:rio}) and within a real-world SLAM system (Sec-\ref{sec:two_agent}).
    \item \textit{Efficiency}. \revise{Encoding the sparse scene representation requires far fewer GPU resources} in Sec-\ref{sec:rio}, enhances communication efficiency in Sec-\ref{sec:two_agent}, and accelerates inference speed in Sec-\ref{sec-runtime}.
    \item \textit{Self-supervised training}. Our training approach, explained in Sec-\ref{sec:setup}(4), involves data automatically generated from posed RGB-D sequences.
\end{itemize}
\vspace{-0.3cm}

\subsection{Set-up}\label{sec:setup}
\subsubsection{Benchmarks}
We run our method on two public datasets: ScanNet\cite{dai2017scannet} and 3RScan\cite{wald2019rio}.
\begin{table}[ht]
    \centering
    \begin{tabular}{c c c}
        \toprule
        & Train using & Train using \\
         & {3RScan-GT} & {ScanNet-Mapping} \\
        \hline
        3RScan-GT evaluation & Sec. \ref{sec:3rscan} & - \\
        ScanNet-Mapping evaluation & - & Sec. \ref{sec:scannet} \\
        \textbf{3RScan-Mapping} evaluation & - & {Sec. \ref{sec:rio}}  \\
        \bottomrule
    \end{tabular}
    \caption{Dataset splits for training and evaluation.}
    \label{tab:dataset}\vspace{-0.3cm}
\end{table}

The 3RScan dataset provides ground-truth scene graph annotations\cite{sarkar2023sgaligner}\cite{xie2024sgpgm}, which we refer to as 3RScan-GT. In addition, considering the difference between ground-truth annotations and real-world data, we use semantic mapping\cite{liu2024fmfusion} method to automatically label the two datasets and refer to them as 3RScan-Mapping and ScanNet-Mapping. The division of the training and testing sets follows the original settings of the datasets.
As shown in Table. \ref{tab:dataset}, we train the baseline method and our method using 3RScan-GT and ScanNet-Mapping separately. We evaluate their cross-domain performance in Sec. \ref{sec:rio}.

Beyond reconstructed scenes, we evaluate our method in a customized two-agent SLAM benchmark. Our scene graph registration modules are deployed into a SLAM system and it registers scene graphs between two agents in a coarse-to-fine strategy. The benchmark is based on RGB-D and inertial sequence data that are collected in the real-world environment. 

\subsubsection{Evaluation metrics}
Node recall (NR) and node precision (NP) evaluate scene graph matching performance. If the intersection over the unit (IoU) between the nodes' point cloud is higher than a threshold $\tau_{iou}=0.3$, we treat it as a true positive (TP) node pair. Otherwise, it is a false positive (FP) pair. Then, we can compute the NR and NP, similar to the recall and precision in image matching task \cite{sarlin20superglue}. In the registration task, we follow GeoTransformer\cite{qin2022geometric} and use Inlier Ratio (IR) and Registration Recall (RR) to evaluate the accuracy. With the predicted transformation, if the root mean square error (RMSE) between the aligned point cloud is within a threshold (\textit{i.e.,} RMSE$<$0.2m), we treat the prediction as a successful registration. Registration recall is the portion of the successful registration. 

\newlength{\figwidth}
\newlength{\subfigwidth}
\setlength{\subfigwidth}{0.6\columnwidth}

\subsubsection{Baselines}
We compare our registration performance against SG-PGM\cite{xie2024sgpgm} and GeoTransformer\cite{qin2022geometric}.  
In two-agent SLAM, our approach is compared with Hydra\cite{hughes2022hydra,hughes2024foundations} and HLoc\cite{sarlin2019coarse}. 
Hydra and HLoc executes offline in the benchmark.

\subsubsection{Training data}\label{sec:train_data}
In ScanNet-Mapping data, each ScanNet sequence is segmented into multiple sub-sequences. We select a pair of the sub-sequences as a source and a reference sequence. A random transformation in 4-DoF is incorporated into each pair of the scene graphs. We run FM-Fusion on each sub-sequence. Hence, we can have a large number of scene graph pairs for training and evaluation. Since the ground-truth annotation from ScanNet is not used, we claim a \textbf{self-supervised scene graph training} is used in our method. The training method can be extended to other indoor RGB-D SLAM data, and it does not require the ground-truth semantic annotation.
In 3RScan-GT data, we directly applied the data generation from SGAligner\cite{sarkar2023sgaligner}. 

\subsubsection{Implementation}
We train our network using an Adam optimizer with a learning rate of $0.01$. The training data involves $1990$ pairs of scene graphs reconstructed from ScanNet-Mapping and the 3RScan-GT dataset. The pre-train of the shape network takes $64$ epochs. The rest of the training takes $80$ epochs. We run the training process and all experiments on a desktop computer with an Intel-i7 CPU and a Nvidia RTX-3090 GPU.\vspace{-0.3cm}

\subsection{3RScan Benchmark}\label{sec:3rscan}
This section evaluates the differences in accuracy between different methods using completely accurate training and validating annotations. We use the officially published version of SG-PGM\cite{xie2024sgpgm} and train and validate our model and SG-PGM using $2,178$ pairs of scene graphs from 3RScan-GT. We make two adaptions in our trained SG-PGM. Firstly, we use BERT to encode their semantic labels, which is identical to our method. The original version of SG-PGM generates a semantic histogram from each of its semantic labels and constructs a semantic feature vector. Since Bert is a stronger semantic encoder, we apply it in SG-PGM to ensure a fair comparison. Secondly, SG-PGM reads the relationship labels of the edges. We assign all of the relationship labels to \textit{none}. This is because the scene graphs in 3RScan-Mapping do not provide any relationship labels, and we set them to none in the two versions of the training.\vspace{-0.2cm}

\begin{table}[ht]
    \centering
    \begin{tabular}{c c c c c}
        \toprule
         & NR(\%) & NP(\%)& IR(\%) & RR(\%) \\
        \hline
         SG-PGM & 63.5 & 45 & 15.7 & 79.4 \\
         Ours & \textbf{85.2} & \textbf{58.1} & \textbf{22.5} & \textbf{82.0} \\
         \bottomrule
    \end{tabular}
    \caption{Train and evaluation on 3RScan-GT.}\label{tab:3rscangt_eval}
    \vspace{-0.3cm}
\end{table}
As shown in Table. \ref{tab:3rscangt_eval}, our node recall and node precision are significantly higher than SG-PGM. Our registration recall is slightly higher than that of SG-PGM. This reflects that we maintain a decisive advantage in semantic node matching. However, our advantage in registration recall is less significant. This is probably because using ground-truth semantic annotations will improve the accuracy of node matching. So, our method cannot fully demonstrate its robustness and superiority. 

For scene graph registration, real-world data always contain segmentation variances and false predicted semantic labels, which greatly affect accuracy. Therefore, we believe training on 3RScan-GT cannot reflect the actual usability of the scene graph registration method but shows an upper bound in accuracy. The experimental results of this section are only used as a reference to prove that our method is also superior to the baseline method in an ideal application scenario.
\vspace{-0.3cm}

\subsection{Cross-Domain Benchmark}\label{sec:rio}
Deep learning methods are often limited by their generalization performance. Considering that our method aims to be applied in the long-term use of robots, we believe it is necessary to evaluate the cross-dataset performance. In this section, we evaluate in detail the generalization, accuracy, and efficiency of our method and two baselines on a validation dataset that is different from their training dataset.

\subsubsection{Baseline Set-up}
We use the officially published version of GeoTransformer\cite{qin2022geometric} and SG-PGM\cite{xie2024sgpgm}. We train them using ScanNet-Mapping and evaluate them in 3RScan-Mapping to test the generalization ability in the different data domains. We make similar modifications to the realization of SG-PGM as in Sec. \ref{sec:3rscan}. We keep all the parameters in the 3D backbone identical, such as voxel size and point feature dimension. We also turn off the ICP\cite{segal2009generalicp} refinement for all the methods to ensure a fair comparison. 

\newlength{\legendwidth}
\setlength{\subfigwidth}{0.66\columnwidth}
\setlength{\legendwidth}{1.95\columnwidth}

\begin{figure*}[ht]
    \centering
    \begin{subfigure}[h]{\legendwidth}{
        \includegraphics[width=\legendwidth]{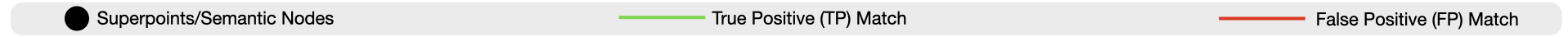}}
    \end{subfigure}

    \begin{subfigure}[h]{\subfigwidth}{
        \includegraphics[width=\subfigwidth]{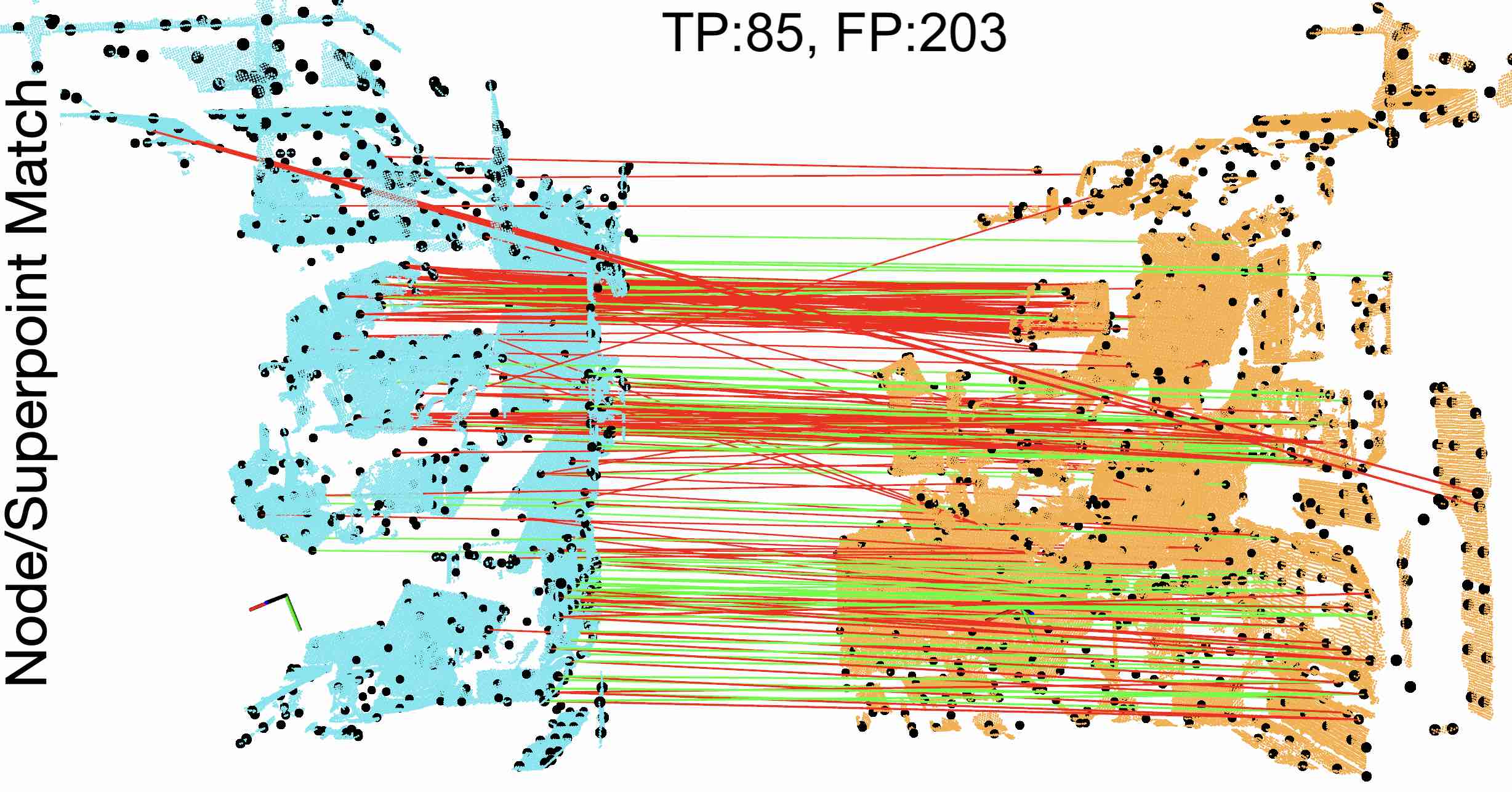}}
    \end{subfigure}
    \begin{subfigure}[h]{\subfigwidth}{
        \includegraphics[width=\subfigwidth]{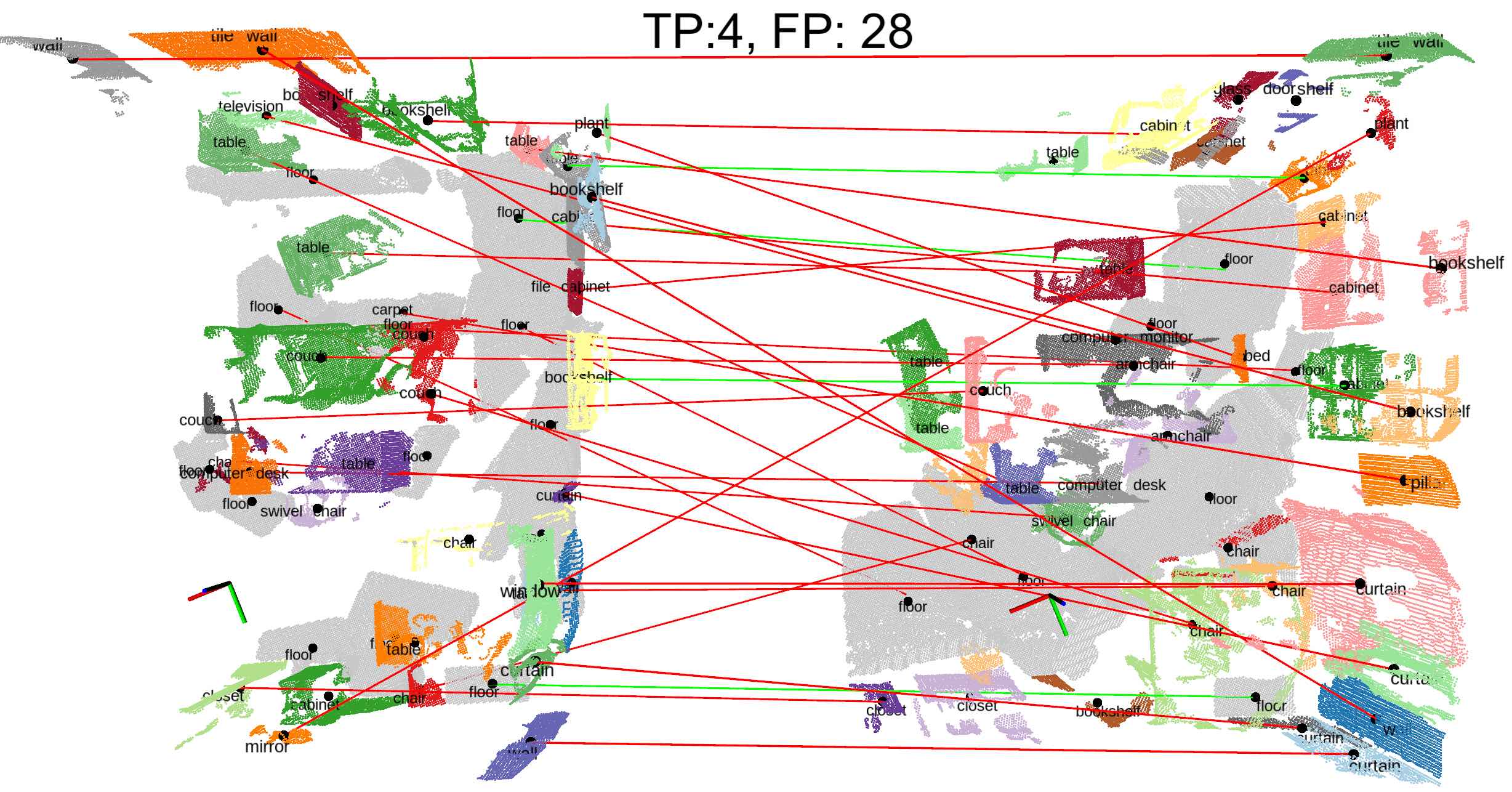}}
    \end{subfigure}
    \begin{subfigure}[h]{\subfigwidth}{
        \includegraphics[width=\subfigwidth]{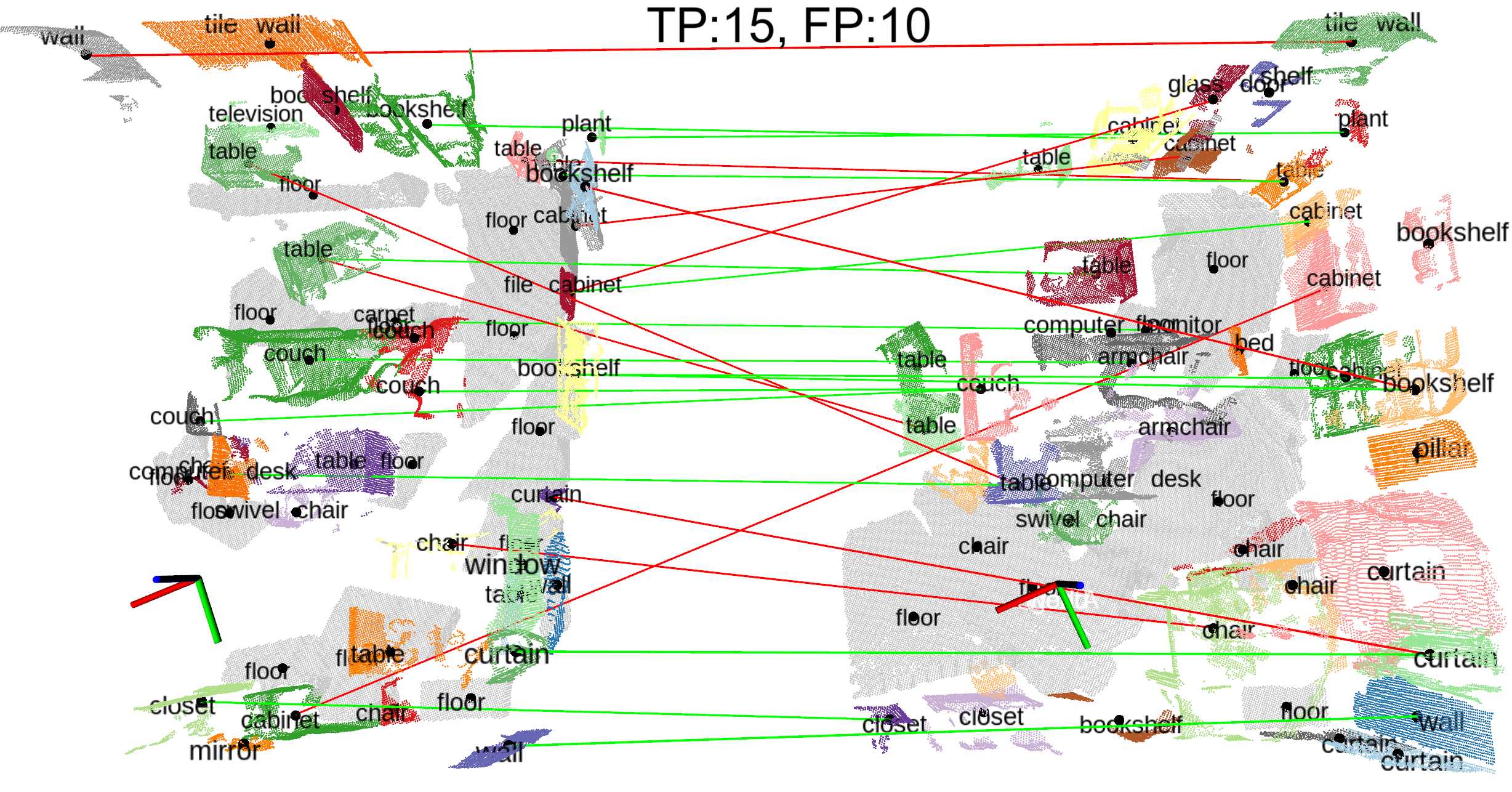}}
    \end{subfigure}
    \begin{subfigure}[h]{\subfigwidth}{
        \includegraphics[width=\subfigwidth]{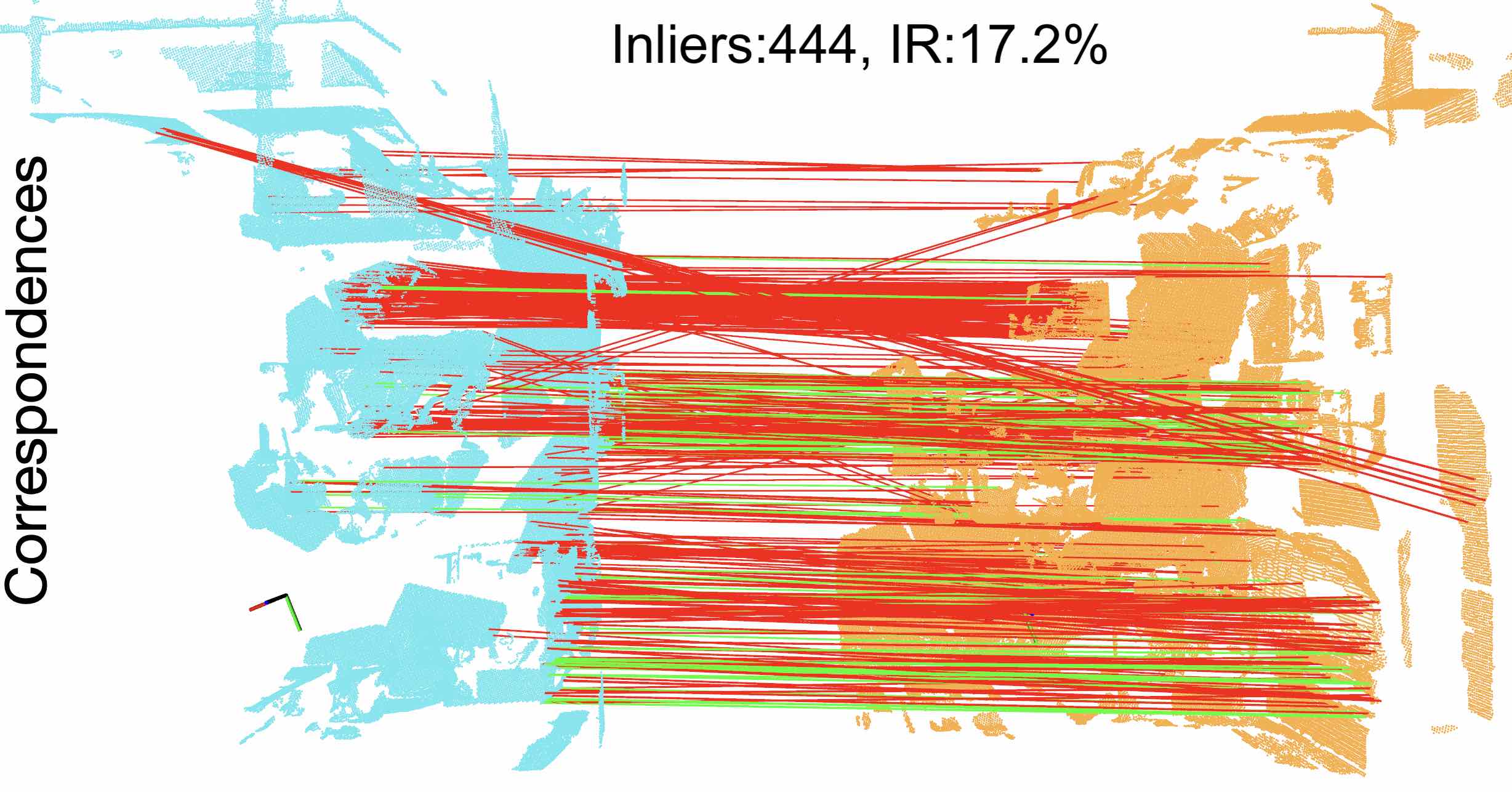}}\vspace{-0.2cm}
    \end{subfigure}
    \begin{subfigure}[h]{\subfigwidth}{
        \includegraphics[width=\subfigwidth]{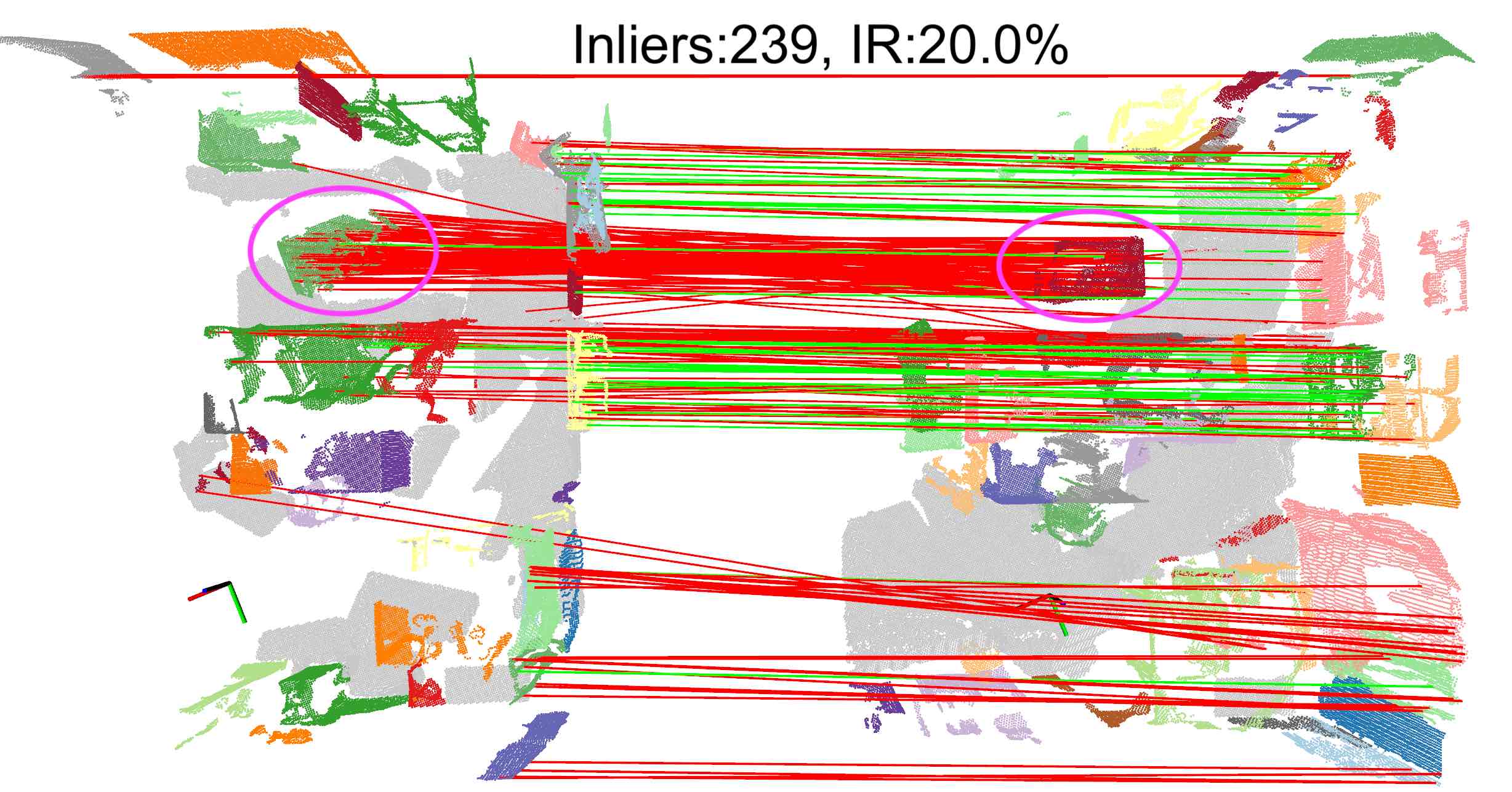}}\vspace{-0.2cm}
    \end{subfigure}
    \begin{subfigure}[h]{\subfigwidth}{
        \includegraphics[width=\subfigwidth]{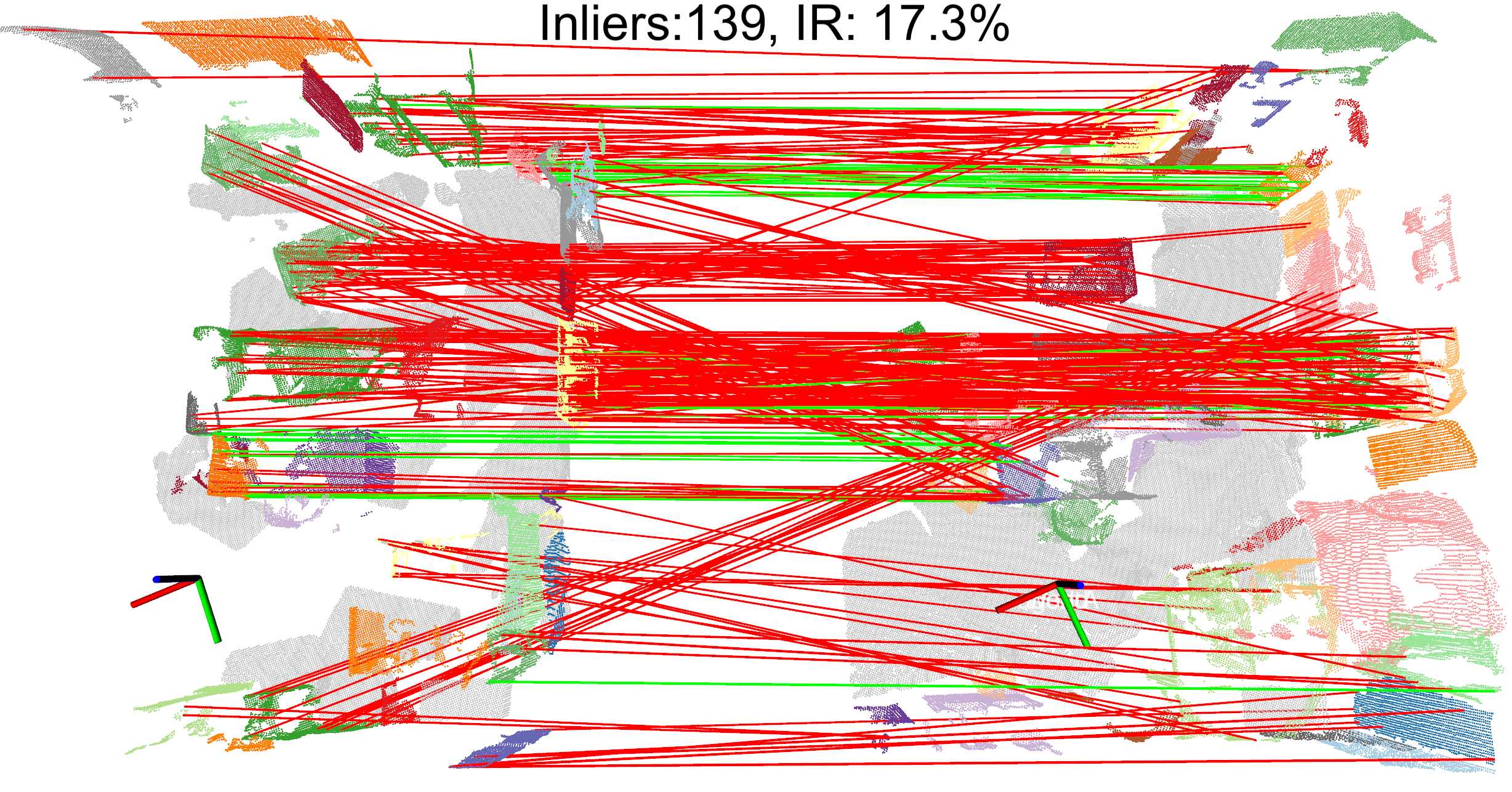}}\vspace{-0.2cm}
    \end{subfigure}
    \begin{subfigure}[h]{\subfigwidth}{
        \includegraphics[width=\subfigwidth]{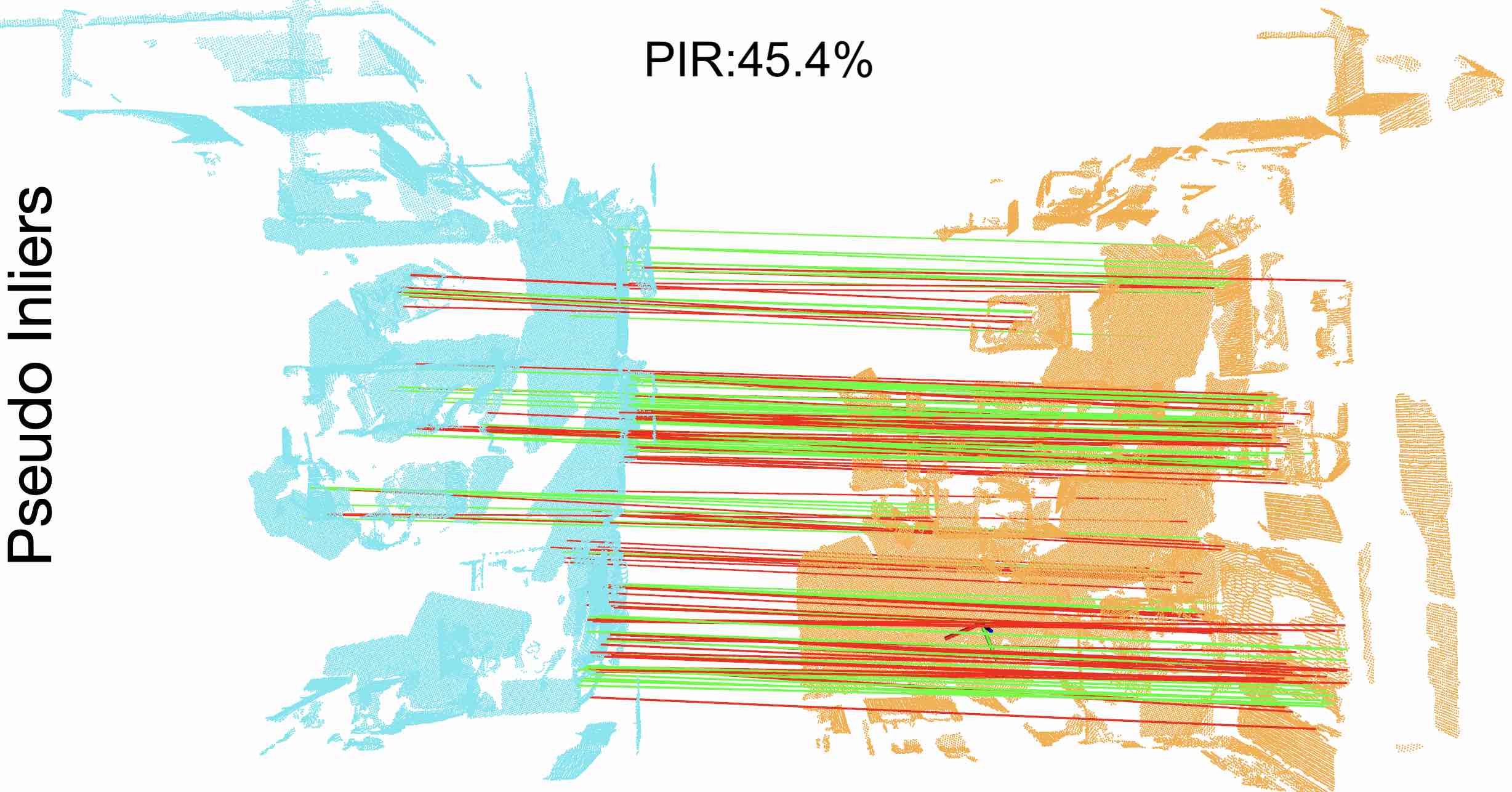}}\vspace{-0.2cm}
    \end{subfigure}
    \begin{subfigure}[h]{\subfigwidth}{
        \includegraphics[width=\subfigwidth]{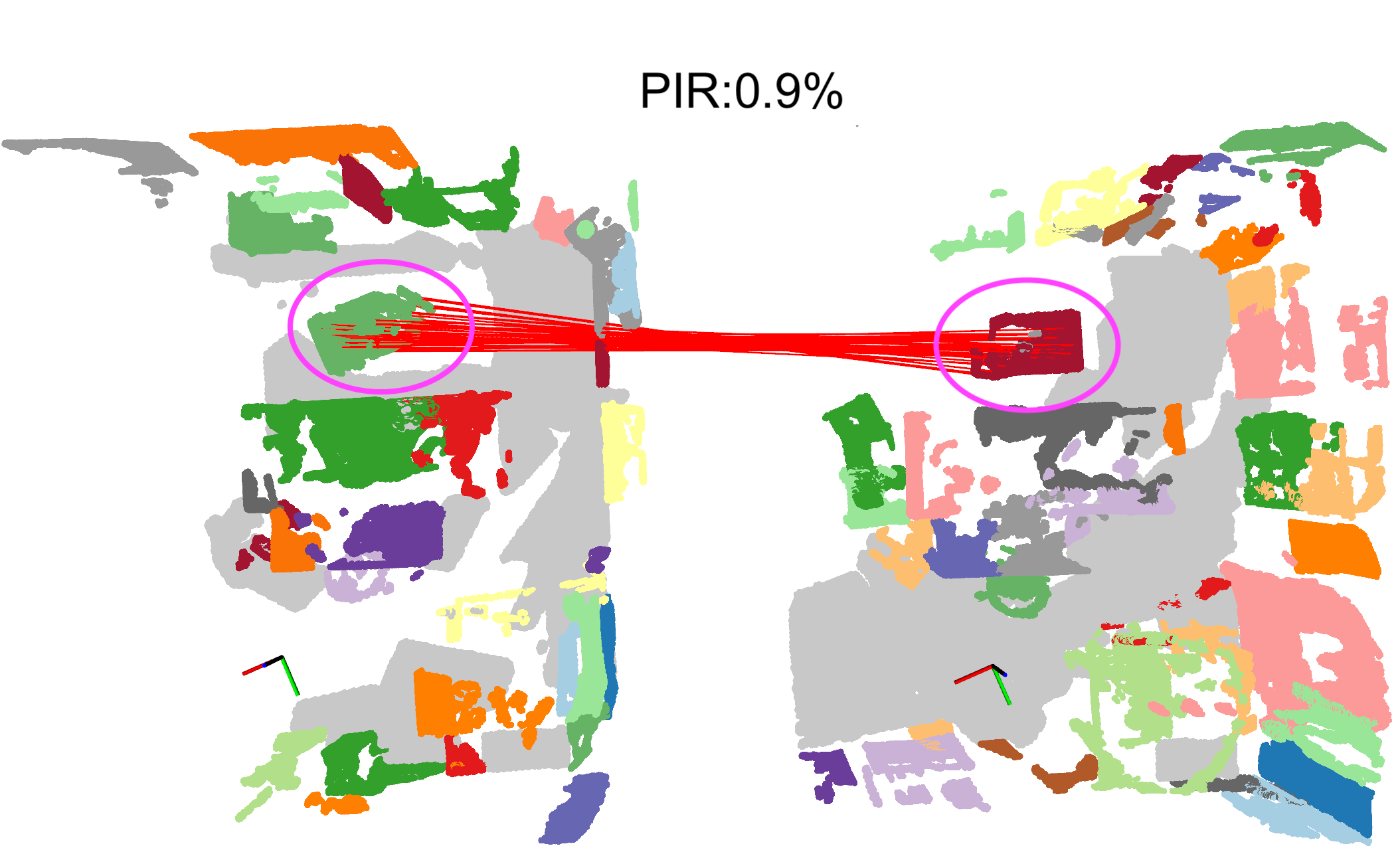}}\vspace{-0.2cm}
    \end{subfigure}
    \begin{subfigure}[h]{\subfigwidth}{
        \vspace{+0.2cm}\includegraphics[width=\subfigwidth]{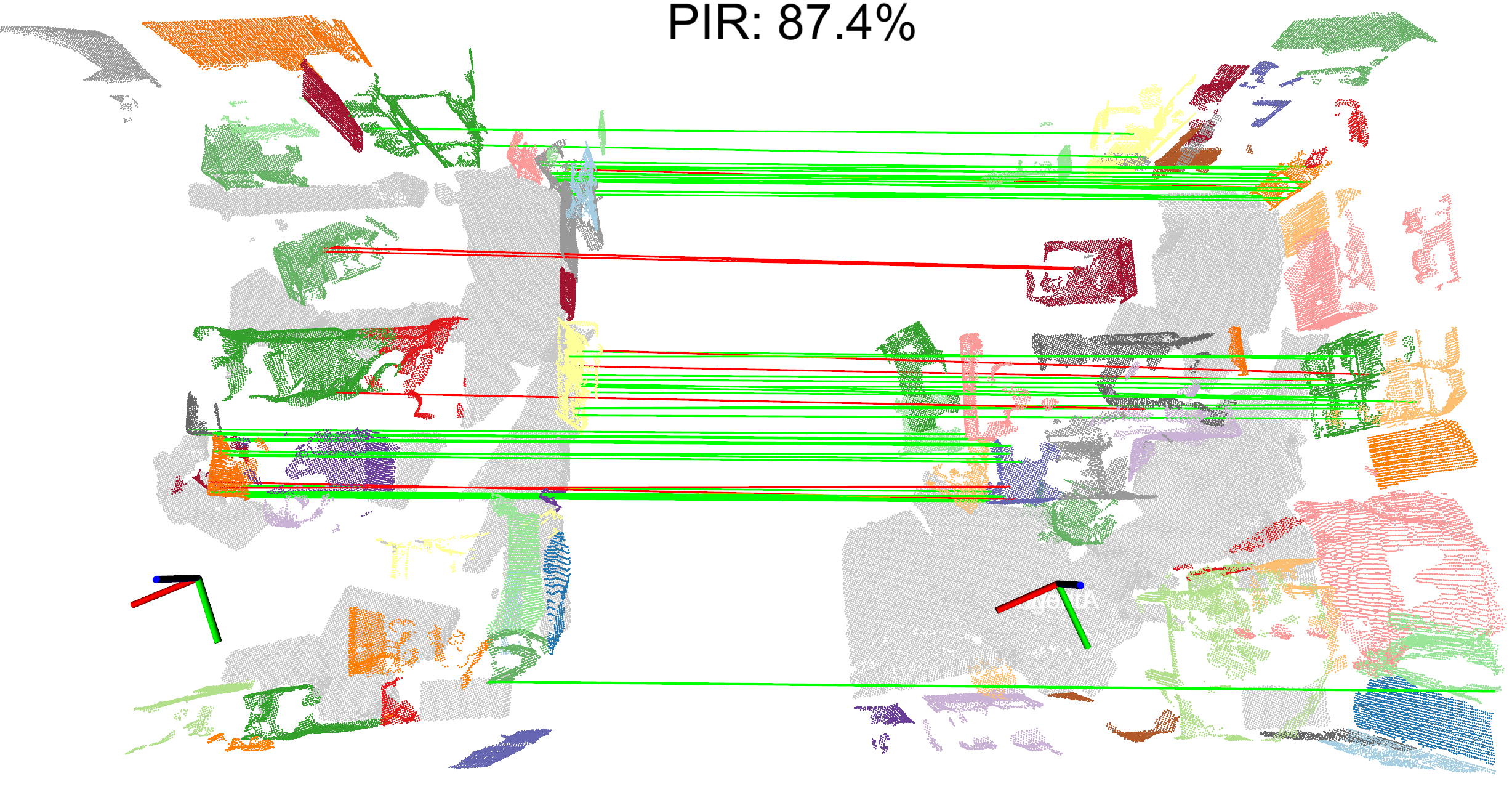}}\vspace{-0.2cm}
    \end{subfigure}
    \begin{subfigure}[h]{\subfigwidth}{
        \hspace{+0.1cm}\includegraphics[width=1.03\subfigwidth]{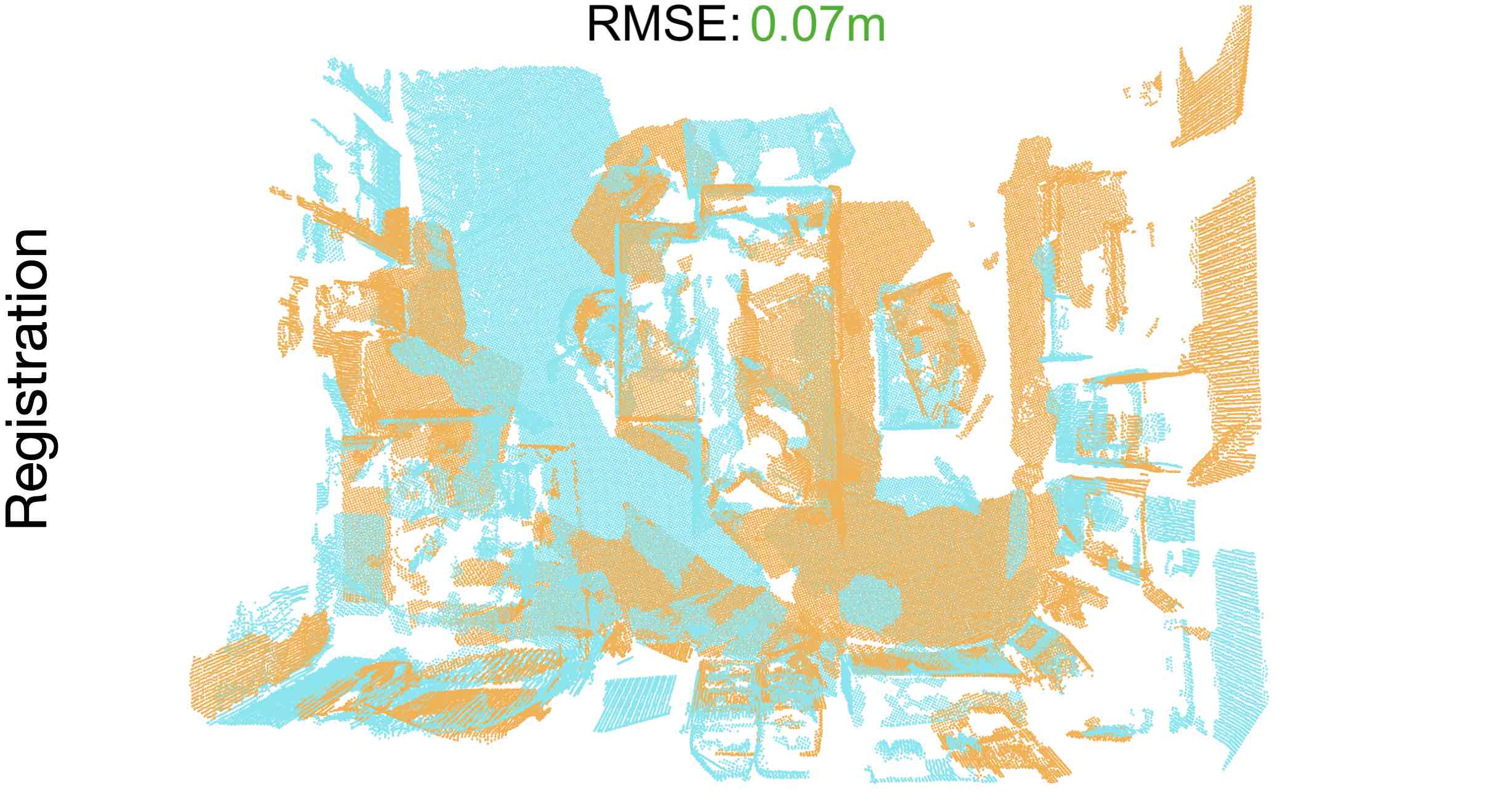}}\vspace{-0.1cm}
        \caption{GeoTransformer}
    \end{subfigure}
    \begin{subfigure}[h]{\subfigwidth}{
        \hspace{+0.4cm}\includegraphics[width=0.82\subfigwidth]{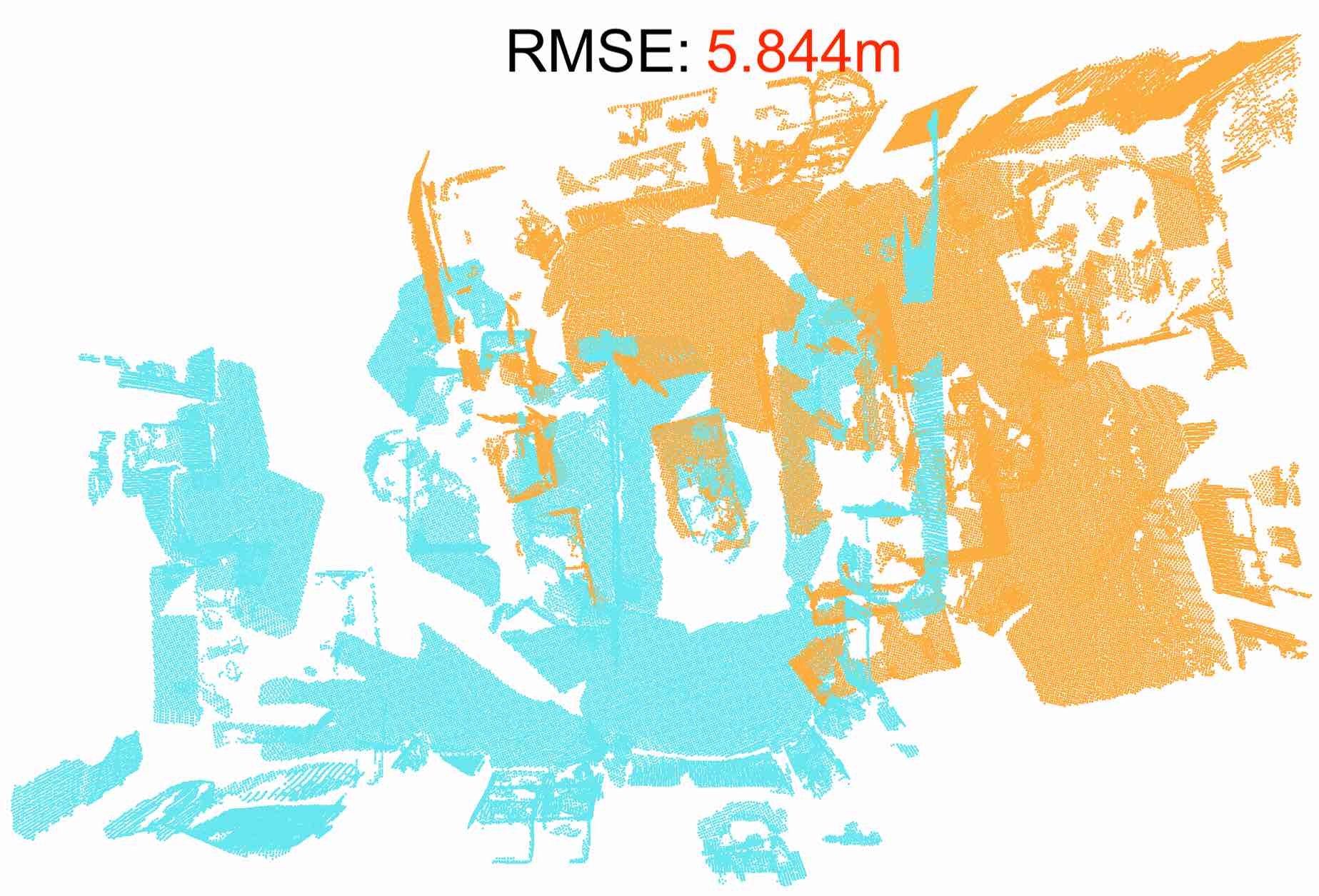}}\vspace{-0.2cm}
        \caption{SG-PGM}
    \end{subfigure}
    \begin{subfigure}[h]{\subfigwidth}{
        \hspace{+0.5cm}\includegraphics[width=0.82\subfigwidth]{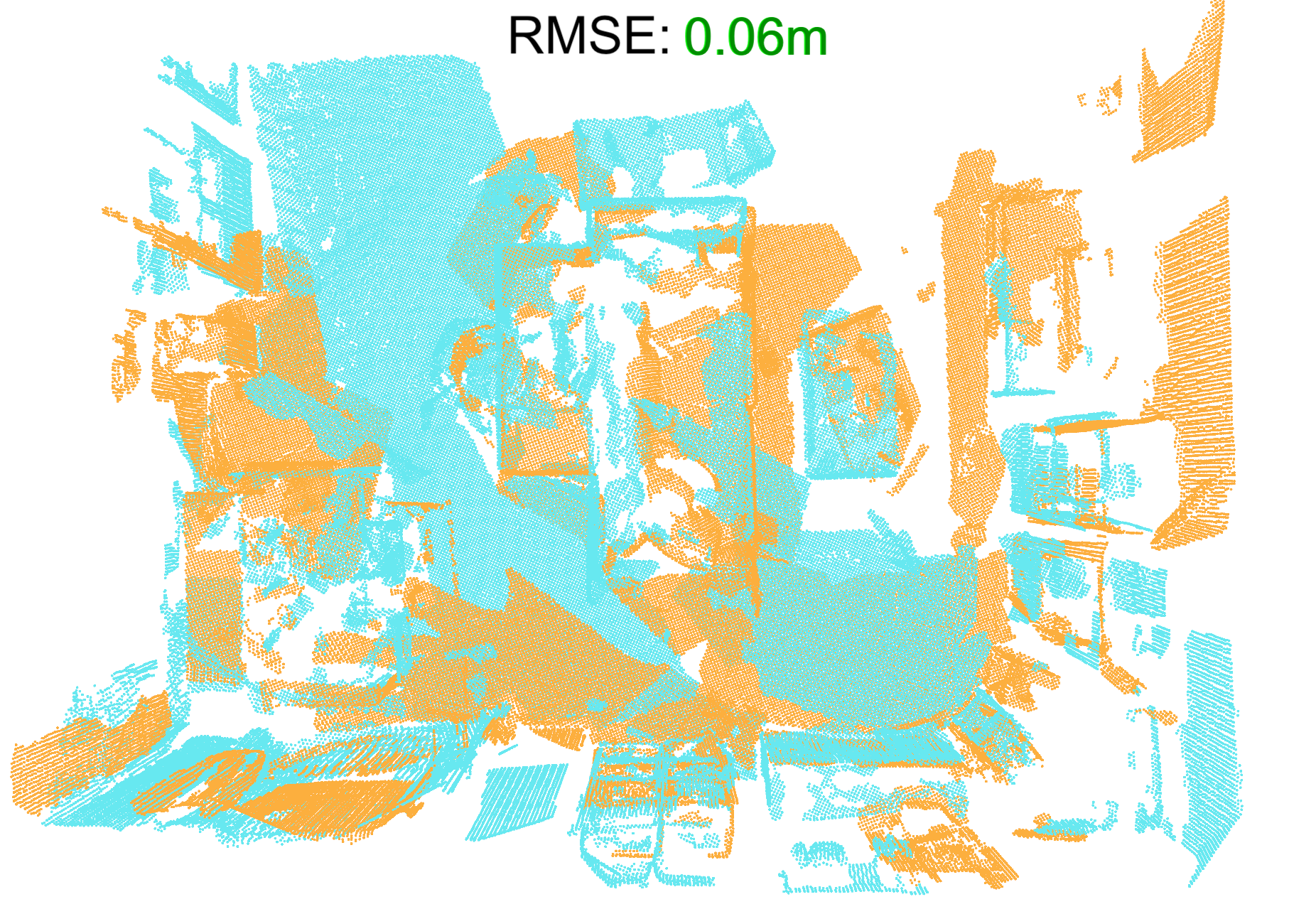}}\vspace{-0.2cm}
        \caption{SG-Reg(\textbf{Ours})}
    \end{subfigure}
    \vspace{-0.2cm}
    \caption{Registration results in a 3RScan median scene. In SG-PGM results, the adversarial outliers are highlighted in purple circles. 
    We annotate the corresponding evaluation metrics in each sub-figure. The RMSE from a successful registration is highlighted in green, while the RMSE from a failed pair is in red.}\label{fig:rio-res}\vspace{-0.5cm}
\end{figure*}

\subsubsection{Registration Accuracy}\label{sec:3rscan_reg}
We split the 100 pairs of scenes from 3RScan-Mapping into three sets: 76 small pairs, 13 medium pairs, and 11 large pairs. \revise{The grouping is based on the maximum number of points in each pair. We summarize the scene points range of each split, as shown in TABLE \ref{tab:splits}.}
\begin{table}[h]
    \centering
    \begin{tabular}{c c c c}
        \toprule
        & Small & Median & Large \\
        \hline
        Points in thousand(k) & 61k to 133k & 146k to 210k & 210k to 330k \\
        \bottomrule
    \end{tabular}
    \caption{Scene points number in each split. All the scene points are downsampled at $2$cm voxel size.}\label{tab:splits}
    \vspace{-0.3cm}
\end{table}

As the median-size scene shown in Fig. \ref{fig:rio-res}, \revise{the semantic scene graphs in 3RScan-Mapping exhibit significant semantic noise, including over-segmented objects and noisy point cloud reconstruction. Our method generates more correct node matches and fewer false node matches than SG-PGM. It demonstrates preciser and more robust scene graph matching performance than SG-PGM. 
Compared with GeoTransformer, our semantic nodes are much sparser than the superpoints from GeoTransformer. 
Thanks to the strong descriptiveness features, our node matching precision is higher than the superpoint matching precision in GeoTransformer.}

To discuss the registration performance, we introduce the definition of \textbf{pseudo inliers}, the correspondences after outlier pruning and being treated as inliers by the estimator. As introduced in Section-\ref{sec-pestimator}, we prune the outliers in point correspondences via maximum clique. Similarly, GeoTransformer and SG-PGM prune outliers via a local-to-global registration \cite{qin2022geometric}, based on a weighted SVD\cite{besl1992method}. The baseline works and SG-Reg all have pseudo inliers.
We compute the pseudo inlier ratio (PIR), the ratio of true inliers within the pseudo inliers. 
As shown in Fig. \ref{fig:rio-res}, \revise{our IR and PIR are higher than those of GeoTransformer, demonstrating more accurate point matching. In the final registration result, both SG-Reg and GeoTransformer successfully align the scenes.}

However, in SG-PGM, the presence of adversarial outliers leads to registration failure. As shown in Fig.\ref{fig:rio-res}, these outliers are geometrically consistent and thus more challenging to remove. Despite efforts to prune outliers, these adversarial outliers persist in the pseudo inliers. While adversarial outliers also appear in GeoTransformer and our result, SG-PGM is particularly susceptible as it cannot effectively prune them due to its high outlier ratio. 
\revise{Additionally, SG-PGM's outlier pruning method, which relies on a batch of weighted SVD \cite{besl1992method}, may impede its ability to eliminate adversarial outliers. It searches for a mini-batch of correspondences that contains the largest number of pseudo inliers. Consequently, this mini-batch, which includes the adversarial outliers, is likely to receive the highest confidence score, leading to a false SVD prediction.}

\begin{table}[ht]
    \centering
    \begin{tabular}{c c c c c c}
        \toprule
        & & NR(\%) & NP(\%) & IR(\%) & RR(\%)\\
        \hline
        \multirow{3}{*}{Small} & GeoTransformer & \underline{47.9} & \textbf{47.9} & \underline{14.5} & \underline{76.3}\\
        & SG-PGM & {35.2} & 22.5 & {5.5} & {40.8} \\
        & Ours & \textbf{\revise{66.8}} & \underline{\revise{40.6}} & \textbf{\revise{20.4}} & \textbf{\revise{80.3}}\\
        \hline
        \multirow{3}{*}{Median}& GeoTransformer & {25.6} & \underline{27.1}& \underline{10.6} & \underline{61.5} \\
        &SG-PGM& \underline{26.8} & 15.7 & {3.9}& 7.7 \\
        & Ours & \revise{\textbf{58.8}} & \revise{\textbf{36.9}} & \textbf{\revise{19.9}} & \textbf{\revise{76.9}}\\
        \hline
        \multirow{3}{*}{Large} & GeoTransformer & -&-&-&-\\
        &SG-PGM & 26.2 & 12.0 & 1.6 & 27.3\\
        & Ours & \textbf{\revise{51.9}} &  \textbf{\revise{32.8}} & \textbf{\revise{14.4}} & \textbf{\revise{72.7}} \\
        \hline
        \multirow{3}{*}{\textbf{Overall}} & GeoTransformer& \underline{43.8} & \textbf{41.9} & \underline{13.9} & \underline{74.2} \\ 
        & SG-PGM & {32.9} & 20.0 & {5.0} & 35.0 \\
        & {Ours}& \textbf{\revise{64.9}} & \underline{\revise{38.9}} & \textbf{\revise{19.5}} & \textbf{\revise{79.0}} \\ 
        \bottomrule
    \end{tabular}
    \caption{Scene graph registration performance. In each split, we highlight the \textbf{best} metric and the \underline{second best} metric.
    }\label{tab:3rscan}
    \vspace{-0.3cm}
\end{table}

Then, we summarize the quantitative results in Table. \ref{tab:3rscan}. 
As shown in Table. \ref{tab:3rscan}, our overall performance is much better than SG-PGM. We outperform SG-PGM in each level of matches and the final registration results. Considering that our method and SG-PGM have adopted a strategy of explicitly fusing node features and shape features, this significant accuracy gap further demonstrates the scientific validity of our encoding method.
\revise{It also validates that our method is able to register scene graphs across data distribution, demonstrating strong generalization ability.}

Compared with GeoTransformer, \revise{our method achieves slightly better performance, improving registration recall by 4.8\% and the inlier ratio by 5.6\%. However, in median-sized scenes, we significantly outperform GeoTransformer, achieving a 15.4\% higher registration recall. This demonstrates that the primary advantage of incorporating semantic information for registration lies in achieving higher recall in larger scenes. In larger scenes, frequent similar geometric structures exist, such as point clouds from multiple chairs or numerous books. Relying solely on 3D point features to register large scenes is extremely challenging. By incorporating semantic information, we enhance the descriptive ability of the encoded 3D features.}

\begin{table}[ht]
    \centering
    \begin{tabular}{c c c c}
        \toprule
        & GeoTransformer & SG-PGM & \textbf{Ours} \\
        \hline
        Inlier Rate (\%) & \underline{14.2} & {5.0} & \textbf{\revise{19.5}} \\
        Pseudo Inlier Rate (\%) & \underline{31.5}(+17.3) & 18.7(+13.7) & \textbf{\revise{50.3}}(+30.8) \\
        \bottomrule
    \end{tabular}
    \caption{Analysis point correspondences and their pseudo inliers. \revise{The improvements after prune outliers are highlight as a precision gain.}}\label{tab-pir}
    \vspace{-0.3cm}
\end{table}
We also report the pseudo inliers to support our improvement. {As shown in TABLE \ref{tab-pir}, our PIR is higher than that of SG-PGM and GeoTransformer. A higher PIR increases the likelihood that the pose estimator will predict an accurate transformation $\vartT^B_A$. Additionally, we note that each method in TABLE \ref{tab-pir} has a PIR greater than its IR. However, our PIR has improved by 30.8\%, which represents a larger gain than the baseline. This suggests that SG-Reg generates fewer adversarial outliers, allowing for more effective pruning of outliers and resulting in a greater precision gain after outlier removal.}

\subsubsection{\revise{Computational efficiency}}

\begin{figure}[ht]
    \centering
    \begin{subfigure}[h]{\columnwidth}
       \includegraphics[width=\columnwidth]{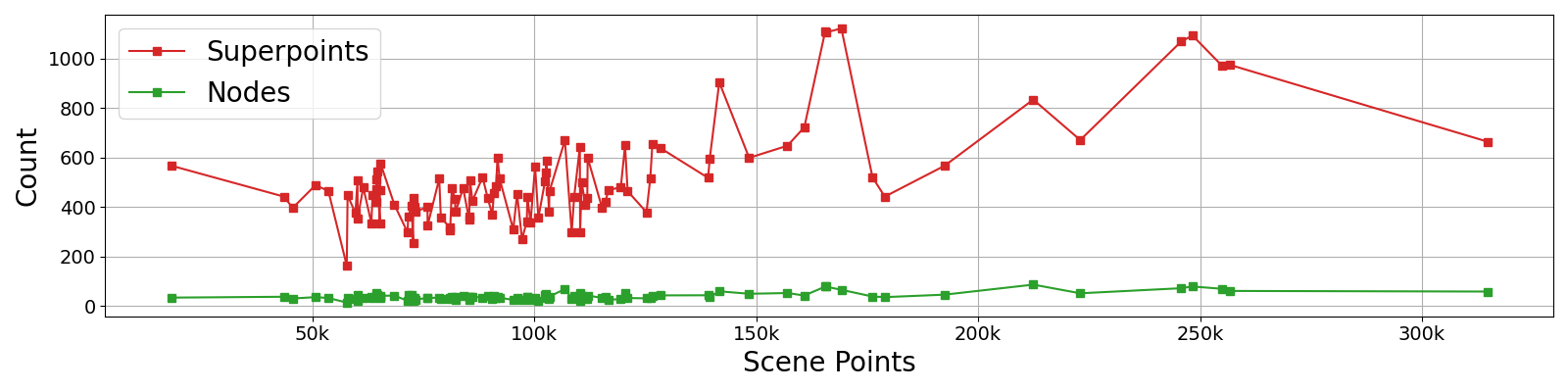}\vspace{-0.3cm}
       \caption{}
    \end{subfigure}
    \begin{subfigure}[h]{\columnwidth}
        \includegraphics[width=\columnwidth]{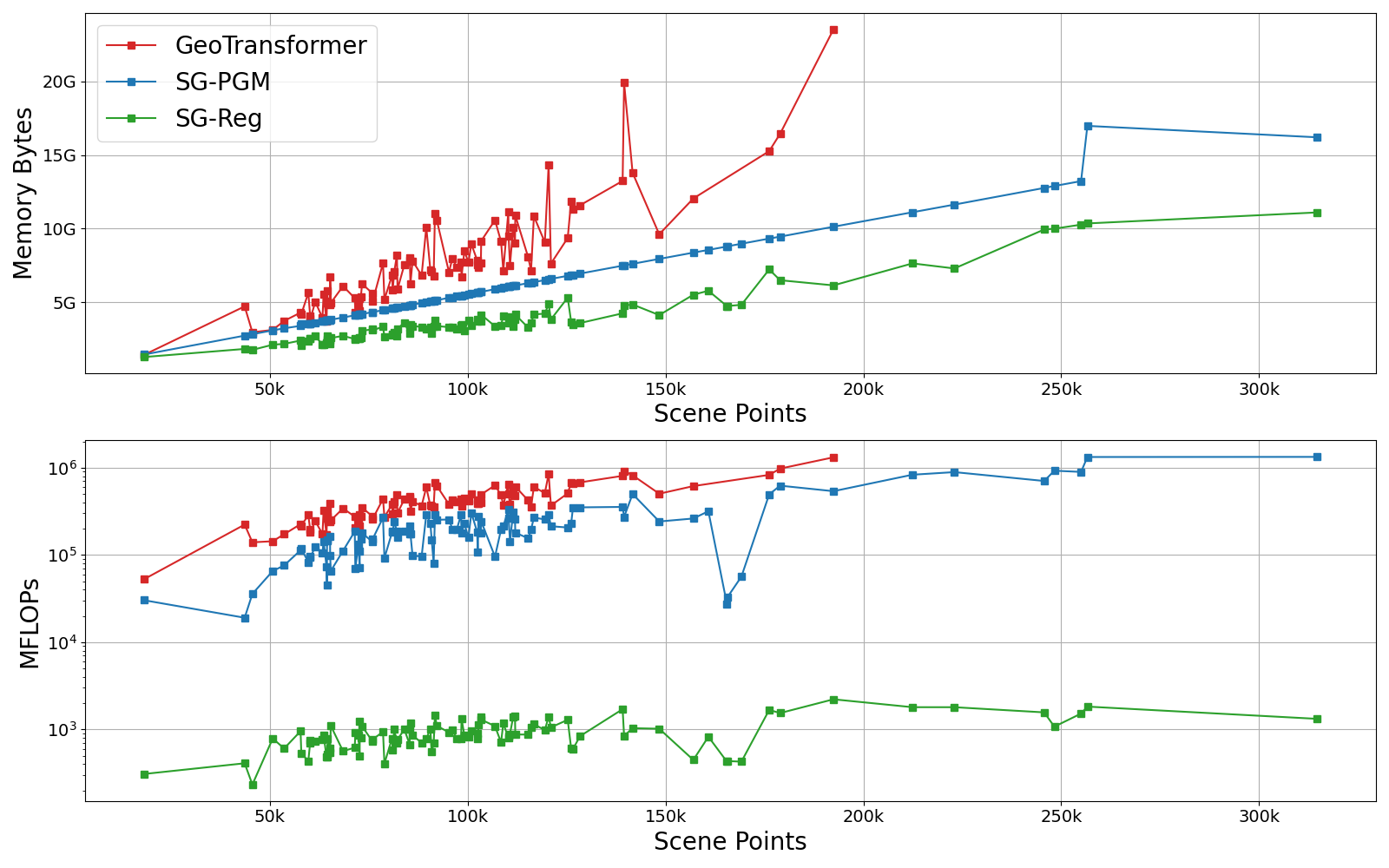}\vspace{-0.3cm}
        \caption{}
    \end{subfigure}\vspace{-0.2cm}
    \caption{\revise{(a) Sparsed 3D representations from the 3RScan scenes in the cross-domain benchmark. (b) The reserved memory and computational complexity on a Nvidia RTX-3090 GPU. The MFLOPS are shown in exponential scales.
    }}\label{fig:memory}\vspace{-0.3cm}
\end{figure}
\revise{SG-Reg requires significantly fewer floating-point operations per second (FLOPS) and consumes less GPU memory compared to baseline models. This enhanced computational efficiency arises from its ability to encode scenes at a higher compression ratio. As illustrated in Fig. \ref{fig:memory}(a), SG-Reg reduces a scene to dozens of semantic nodes, while GeoTransformer compresses it into hundreds of superpoints. Given that attentional operations \cite{vaswani2017attention} on an input feature set have a computational complexity of $O(n^2)$, where $n$ is the number of input features, performing attentional operations on semantic nodes is far more computationally efficient than on superpoints.}

To verify our analysis, we design a study that runs a single GNN layer from GeoTransformer, SG-PGM and SG-Reg. We conduct the study in all of target scenes. 
\revise{As shown in in Fig. \ref{fig:memory}(b), SG-Reg consumes the least GPU memory and requires far fewer FLOPS than the two baseline models. Even for large scenes exceeding 210K points, our reserved GPU memory remains below 11GB, and the FLOPS are below $2,215$ MFLOPS. In contrast, GeoTransformer requires more than $1,000,000$ MFLOPS, significantly higher than ours. Besides, GeoTransformer cannot inference the large scenes due to memory limitations; it exceeds the maximum GPU memory capacity of 24GB on the Nvidia RTX-3090. SG-PGM, our closest baseline, exhibits efficiency that falls between SG-Reg and GeoTransformer, as it inherits the superpoint encoding from GeoTransformer, necessitating more GPU resources than our approach.}

\vspace{-0.3cm}\subsection{ScanNet Benchmark}\label{sec:scannet}
We further evaluate our performance in ScanNet-Mapping evaluation split, which includes $218$ pairs of scene graphs. In the ScanNet benchmark, apart from overall accuracy, we focus on exploring the factors that influence scene graph matching, including the effects from GNN backbone, shape fusion strategy, and perception field of the semantic nodes. 

\setlength{\subfigwidth}{0.6\columnwidth}
\begin{figure*}[ht]
    \centering
    \includegraphics[width=1.8\columnwidth]{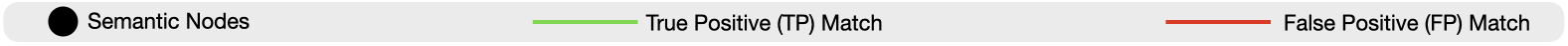}
    \includegraphics[width=\subfigwidth]{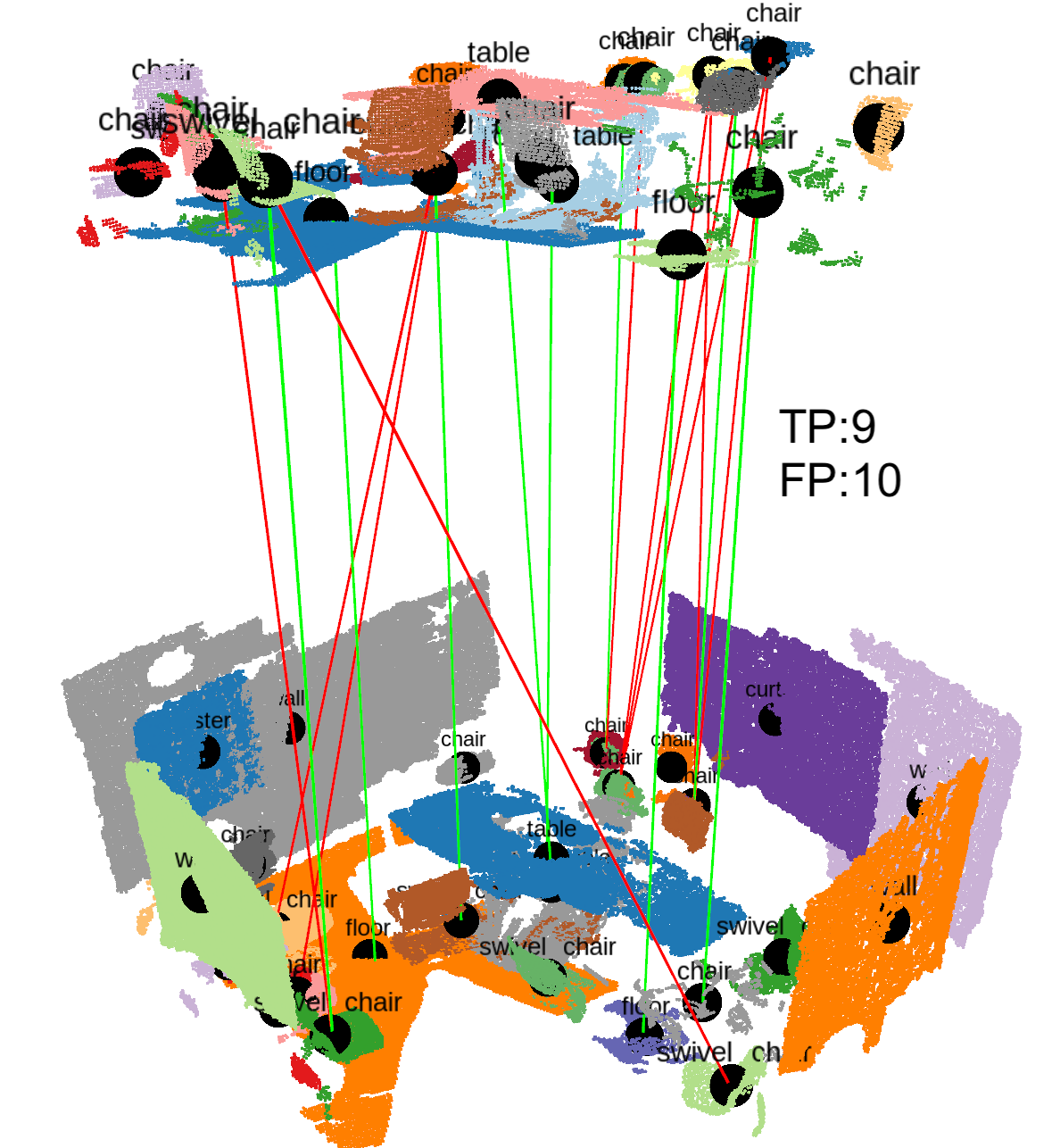}
    \includegraphics[width=\subfigwidth]{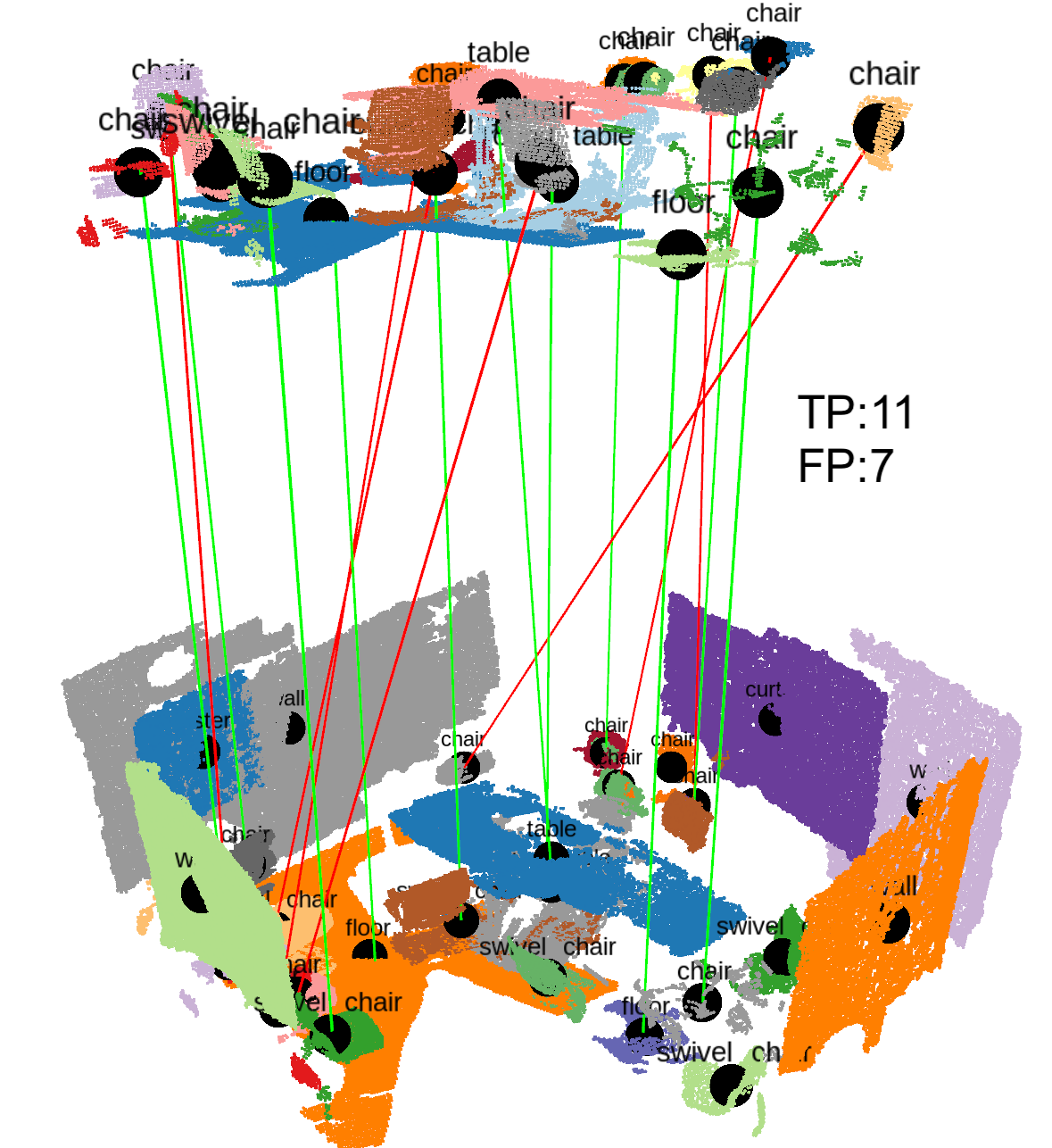}
    \includegraphics[width=\subfigwidth]{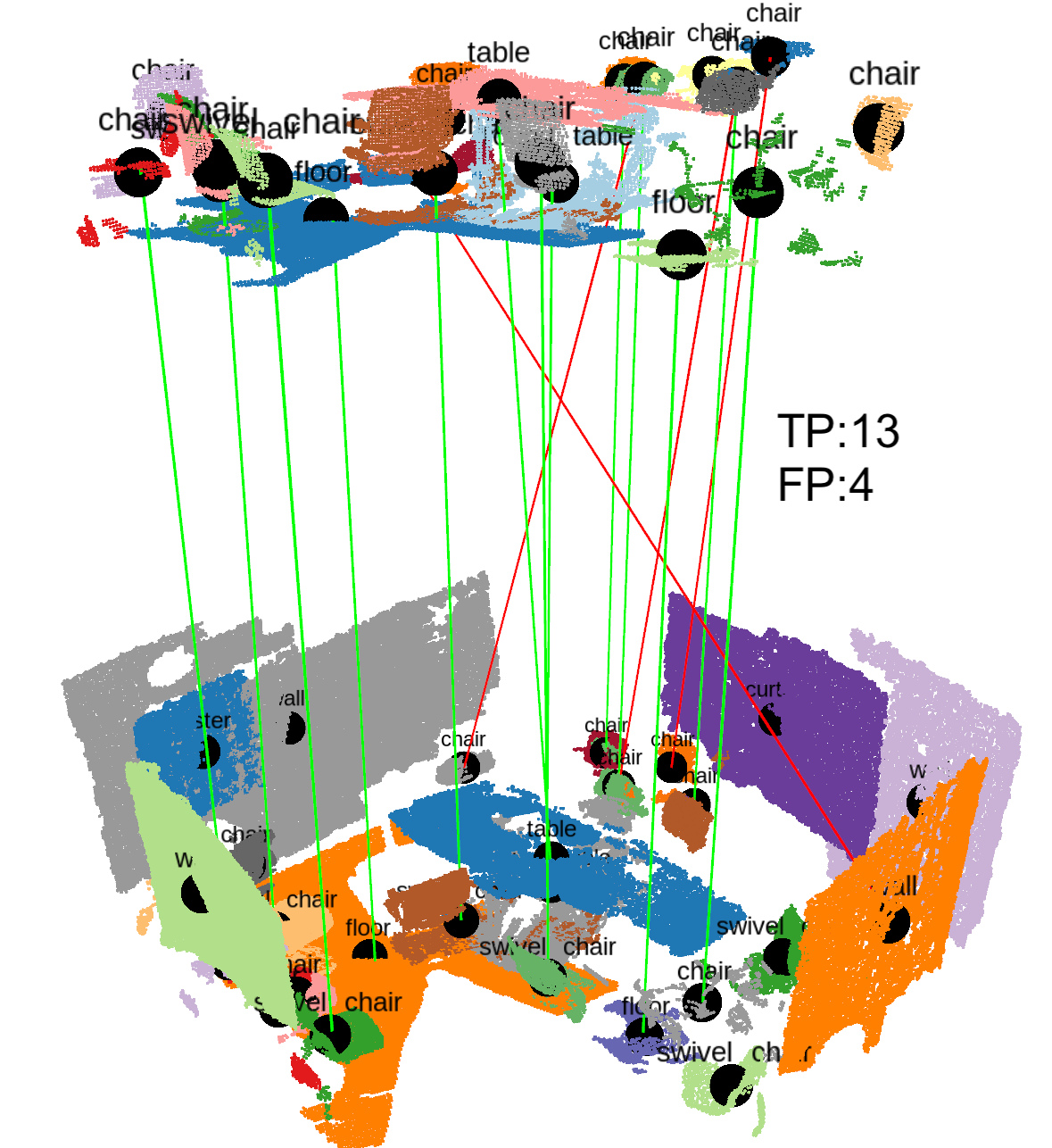}
    \makebox[\subfigwidth]{\small (a) Vanilla GAT}
    \makebox[\subfigwidth]{\small (a) RPE GAT}
    \makebox[\subfigwidth]{\small (a) Triplet-boosted GNN (Ours)}
    \vspace{-0.2cm}\caption{Visualize the scene graph match result from different GNN backbone. Results are from ScanNet \textit{scene0655\_02}.}\label{fig:scannet-res}\vspace{-0.5cm}
\end{figure*}

\subsubsection{Registration comparison}
As shown in Table. \ref{tab:scannet_sgpgm}, without considering cross-domain performance, our node recall and node precision are largely higher than SG-PGM. Our registration recall is slightly higher than SG-PGM. We report the source domain results to show each method's upper bound.

\begin{table}[ht]
    \centering
    \begin{tabular}{c c c c c}
        \toprule
         &  NR(\%) & NP(\%) & IR(\%) & RR(\%)\\
         \hline
        SG-PGM & $64.6$ & $53.3$ & ${29.1}$ & $94.0$ \\
        Ours & \revise{$\mathbf{86.2}$} & \revise{$\mathbf{63.1}$} & \revise{$\mathbf{63.4}$} & \revise{$\mathbf{99.1}$} \\
         \bottomrule
    \end{tabular}
    \caption{Evaluate registration in ScanNet-Mapping.}
    \label{tab:scannet_sgpgm}\vspace{-0.3cm}
\end{table}

\subsubsection{Triplet-boosted GNN evaluation}
We demonstrate that the triplet-boosted GNN can learn the node features with stronger spatial awareness. Three baseline GNN backbones are used as baselines: vanilla GAT\cite{brody2021gatv2}, the RPE GAT used in LightGlue\cite{lindenberger2023lightglue}, and Geometric Transformer \cite{qin2022geometric}. We apply each GNN backbones to replace the triplet-boosted GNN we use. All methods are trained and evaluated on ScanNet. To focus on evaluating the influence of GNN, we skip the shape fusion module in all the baseline methods and our method. We directly use the node features after the GNN to match the semantic nodes.

In the office scene depicted in Fig. \ref{fig:scannet-res}, the triplet-boosted GNN generates more TP matches and fewer FP matches. The scene features multiple objects within the same semantic categories, positioned closely together. The vanilla GAT only considers a semantic topological relationship. It struggles with these ambiguous objects, resulting in the highest number of FP matches. In contrast, the RPE GAT, which is boosted by a variant relative position information, performs slightly better than the vanilla GAT. \revise{Furthermore, since the triplet feature is invariant to a global transformation in 4DoF, the triplet-boosted GNN outperforms the RPE GAT.}

\begin{table}[ht]
    \centering
    \begin{tabular}{c c c}
        \toprule
        & Node Rec.(\%) & Node Pre.(\%)\\
        \hline
        Vanilla GAT & 72.9 & 47.7 \\
        RPE GAT & 75 & 52.7  \\
        Geometric Transformer & 71.1 & 46.2  \\
        Triplet-boosted GNN (\textbf{Ours}) & \textbf{76.8} & \textbf{57.5} \\ 
        \bottomrule
    \end{tabular}
    \caption{\revise{Evaluation of GNN layer on the ScanNet-Mapping split.} To focus on evaluating the GNN layers, the shape features are not fused in this study.}\label{tab:scannet}\vspace{-0.3cm}
\end{table}
Next, we analyze the quantitative results. As shown in Table. \ref{tab:scannet}, our method achieves the highest node recall and node precision in scene graph matching. It proves that triplet-boosted GNN can learn the local topological relationships while distinguishing their spatial distribution. 
On the other hand, the other GNN backbones perform less satisfactorily. Interestingly, we find the geometric transformer performs even poorer than a vanilla GAT. In point cloud registration tasks, geometric transformer \cite{qin2022geometric} is used to encode superpoints and perform very well. However, in scene graph registration tasks,  semantic nodes have a larger geometric variance. Many of them are partially observed, with some being over-segmented. This may introduce extra noise to a node center, which is later propagated into its learned geometric embedding. As a result, geometric transformer performance is degenerated.
\subsubsection{Evaluate Shape Fusion}\label{sec:eval_shape}

\graphicspath{{./assets}}

\begin{figure}[h]
    \centering
    \begin{subfigure}[h]{0.8\columnwidth}
        \includegraphics[width=\columnwidth]{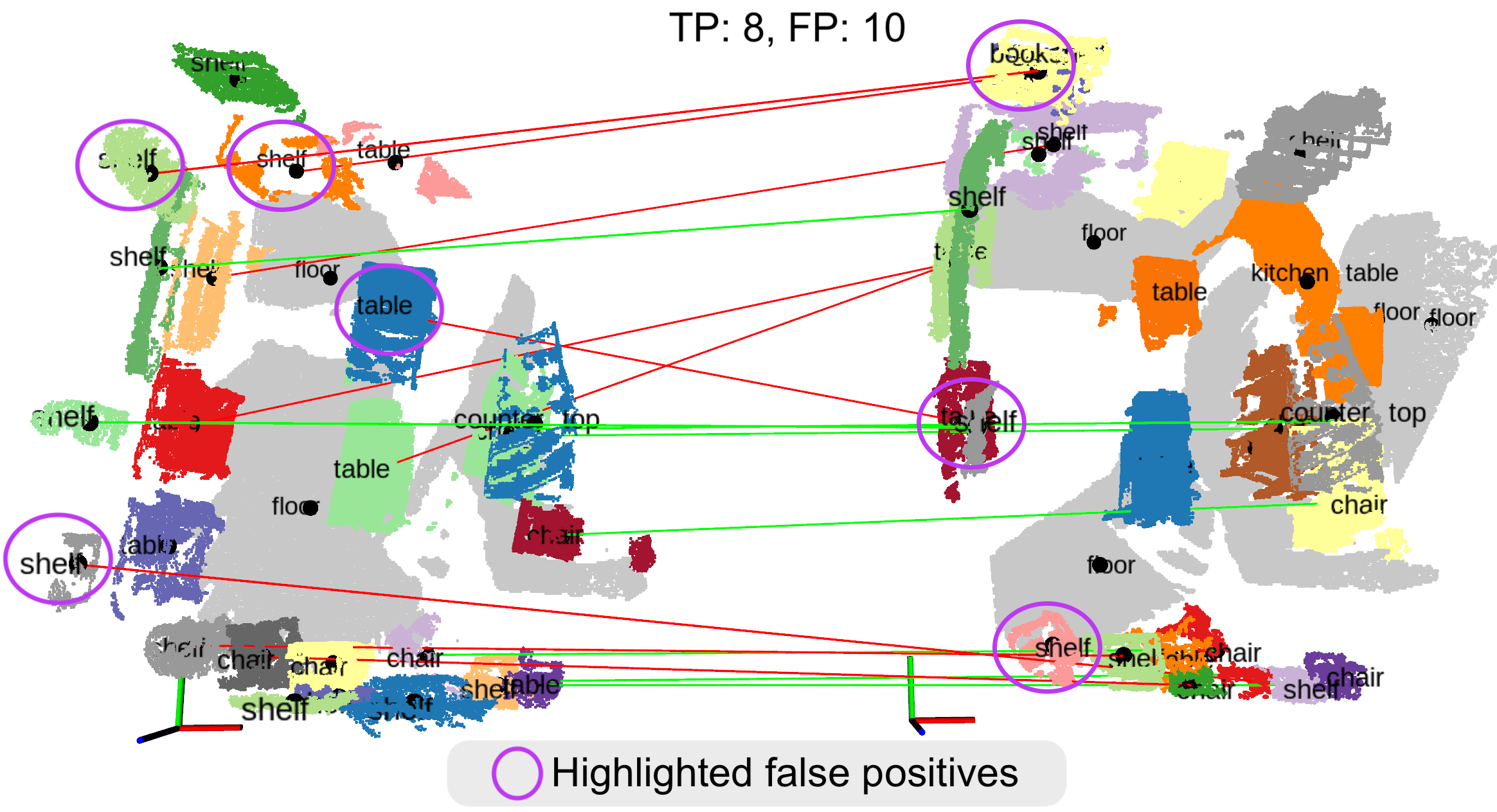}\vspace{-0.2cm}
        \caption{No shape fusion}
    \end{subfigure}
    \begin{subfigure}[h]{0.8\columnwidth}
        \includegraphics[width=\columnwidth]{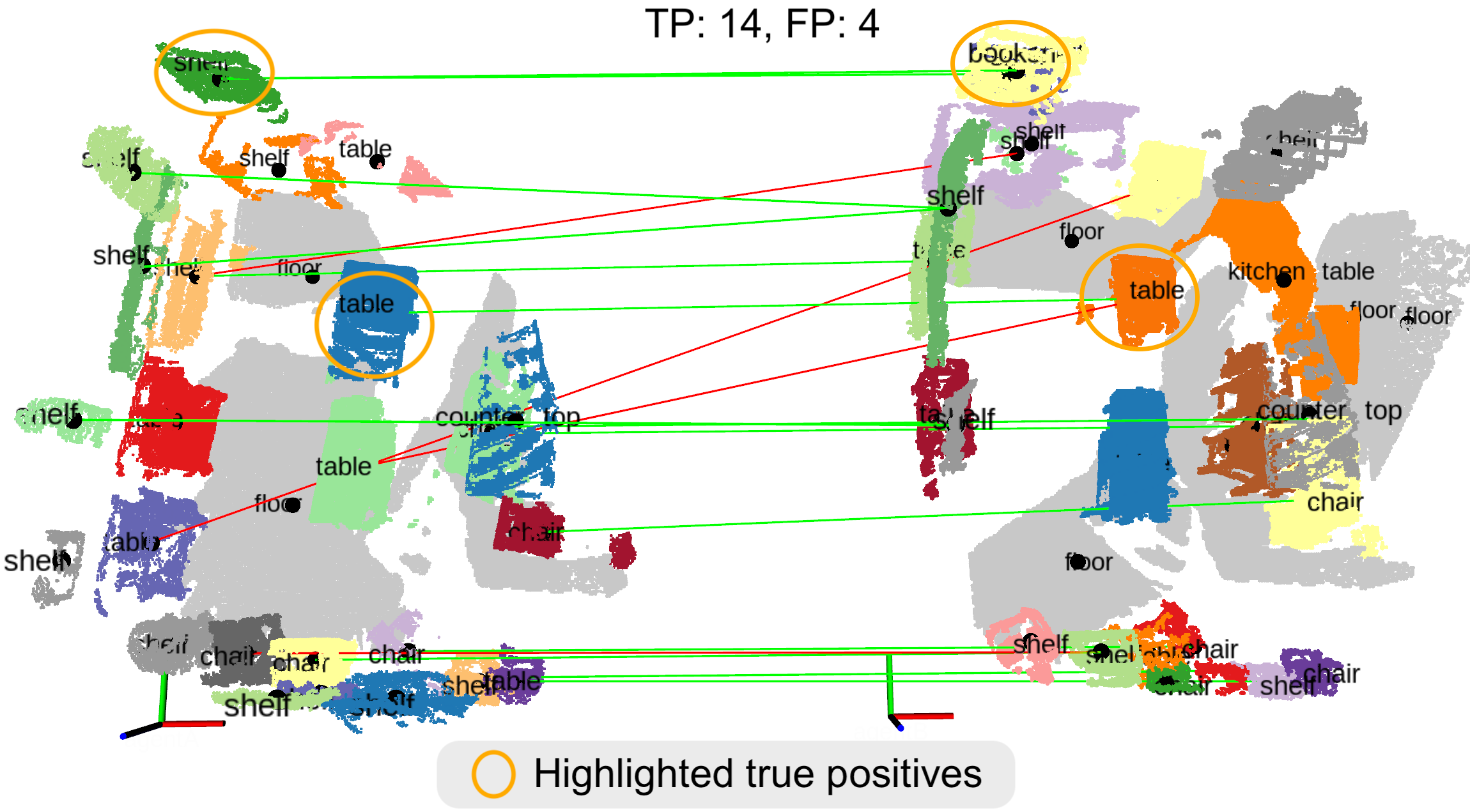}\vspace{-0.2cm}
        \caption{Late fusion}\vspace{-0.2cm}
    \end{subfigure}
    \caption{\revise{Scene graph matching at a ScanNet scene. We highlighted a few false positives that are rejected after the fusion, and a few true positives that are matched after the fusion.}}\vspace{-0.4cm}
    \label{fig:fusion}
\end{figure}

We run an ablation study to evaluate the impact of shape feature. In the no fusion setting, the node features $\{^1\varxxX^A,^1\varxxX^B\}$ after the triplet-boosted GNN are used to search node matches. In the early fusion setting, shape features are fused into node features $[^0\varxxX^{A/B} \Vert \varffeat^{A/B}]$ before the triplet-boosted GNN. While in late fusion, they are fused into node features $[^1\varxxX^{A/B} \Vert \varffeat^{A/B}]$ after the GNN.

We visualize the scene graph matching results using different fusion approaches in Fig. \ref{fig:fusion}. \revise{As shown, a few shelves and a table are incorrectly matched in the no shape fusion mode but are rejected after shape fusion due to their differing shapes. Additionally, some objects are correctly matched after shape fusion, which were missed in the no fusion mode due to semantic noise. This demonstrates how the fused shape features enhance the performance of scene graph matching.}


We summarize the quantitative results of the fusion approaches in Table \ref{tab:eva-shape}. As shown, the late fusion strategy outperforms the others in graph matching. This suggests that while the shape feature is less critical for learning the graph topology, it becomes more significant when fused after the triplet-boosted GNN. In this work, we employ the late fusion strategy, as illustrated in our system structure in Fig. \ref{fig:pipeline}.


\begin{table}[ht]
    \centering
    \begin{tabular}{c c c}
        \toprule
         &Node Rec.(\%) &Node Pre.(\%)\\
        \hline
        No fusion & 76.8 & 57.5 \\ 
        Early Fusion & 79.8(+3.0) & 60.7(+3.2) \\
        Late Fusion (\textbf{Ours}) & \revise{$\mathbf{{86.2}}$(+9.4)} & \revise{$\mathbf{63.1}$(5.6)} \\ 
        \bottomrule
    \end{tabular}
    \caption{Ablation study on the fusion strategy. \revise{The improvements relative to the no fusion mode are annotated.}}\label{tab:eva-shape}\vspace{-0.5cm}
\end{table}

\vspace{-0.3cm}
\subsection{Two-agent SLAM Benchmark}\label{sec:two_agent}
The following experiment estimates the relative localization between agent-A and agent-B. The agents have their local coordinates system, denoted as $O^A$ and $O^B$. At a query frame, whose frame index is $q$, the agent pose estimation in its local coordinate is $\vartT^A_{q}$ or $\vartT^B_{q}$. 
\revise{The benchmark focuses on registering the local coordinates between two agents and predicting a relative transformation $\vartT^B_A$, where $\vartT^B_A \in SE(3)$.}

\subsubsection{Data collection}
We collect the experiment data using a visual-inertial hardware suite\cite{luqi2022tunnels}. It has an intel realsense D-515 camera to capture RGB-D sequence and a DJI A3 flight controller that provides IMU data streaming. 
The experiment involves $10$ pairs of RGB-D sequences inside an academic building. Among them, $5$ pairs of sequences have camera trajectories in opposite directions, as the example scene shown in Fig. \ref{fig1}.
We generate ground-truth transformation $\bar{\vartT}^B_A$ by manually aligning the reconstructed point cloud, following with a ICP refinement.

\subsubsection{Evaluation metrics} \revise{With the predicted transformation $\vartT^B_{A}$ and ground-truth transformation $\bar{\vartT}^B_A$, we compute their \textit{relative translation error} (RTE) and the \textit{relative rotation error} (RRE). If RTE$<0.2m$ and RRE$<5^\circ$, we call it a successful registration.}


\subsubsection{Using Hydra and HLoc}
\revise{When using Hydra and HLoc in agent-A, we assume it has received all the keyframes broadcasted from agent-B between timestamp $[t_0,t_q]$, where $t_q$ is the timestamp at the query frame. The image keyframes in agent-A and agent-B are $\{\mathcal{I}^A_{i}\}_{i}$ and $\{\mathcal{I}^B_{i}\}_{i}$, where the frame index $i \in [0,q]$.}

In the Hydra implementation, \revise{we mainly utilize its loop closure detection module\footnote{https://github.com/MIT-SPARK/Hydra}. We input our scene graph into Hydra, which is used to construct the hand-crafted scene graph descriptor, as introduced in \cite{hughes2022hydra} and \cite{hughes2024foundations}. The scene graph sent to Hydra is identical to the one sent to our method. Hydra retrieves matched frames using the handcrafted descriptor during the global matching step, identifying the top-10 images from the set $\{\mathcal{I}^B_{i}\}_{i}$. It then verifies the matched images using DBoW2. In the local matching step, we implement the image registration module from Kimera\footnote{https://github.com/MIT-SPARK/Kimera-VIO}. Specifically, it performs brute-force matching of ORB features between the query image $\mathcal{I}_q^A$ and Hydra's candidate images. The candidate image with the highest number of matched ORB features is selected as the final matched image $\mathcal{I}_l^B$. Subsequently, it uses the query depth image $\mathcal{D}^A_q$ and the image correspondences to solve a RANSAC-based Perspective-n-Point (PnP) problem\cite{brachmann2017dsac}, predicting a relative transformation $\vartT^{l}_{q}$. By utilizing their local pose estimation $\vartT^A_{q}$ and $\vartT^B_{l}$, the implemented Hydra predicts a $\vartT^B_A$.}

We noticed that the latest version of Hydra\cite{hughes2024foundations} introduces a GNN-based scene graph descriptor. However, it does not publish the source code or its model weight. So, we can only choose the hand-crafted descriptor to compare.

In the implemented HLoc\footnote{https://github.com/cvg/Hierarchical-Localization}, \revise{it runs NetVLAD to retrieve the top-$k$ matched images and we set $k=10$.
Next, it runs LightGlue to match features in each of the retrieval images. HLoc selects the candidate image with the most matched features as the final loop image $\mathcal{I}^B_l$. Similar to the pose estimation in Hydra,
we take the image matching result and the depth image $\mathcal{D}^A_{q}$ to run RANSAC-based PnP\cite{brachmann2017dsac}. With their local pose estimation, we predict the relative transformation $\vartT^B_A$.}

We run Hydra and HLoc in an offline setting. They can access the same data streaming from the two agents as our method. 
In designing the communication strategy for HLoc and Hydra, we follow the communication mechanism in $D^2$SLAM\cite{xu2022dslam}, which relies on image matching modules to detect loop closure. We assume each agent incrementally broadcasts its live image and live image features(\textit{i.e.} NetVLAD, DBoW.). And we record the bandwidth that belongs to each type of message.

\subsubsection{Registration as a rigid body}
It is important to note that \revise{this benchmark focuses on evaluating the registration between agents while neglecting the pose drift within the trajectory of a single agent. In a previous multi-agent benchmark\cite{chang2023hydramulti}, the system relies on Hydra to detect loop closures across agents. The loop closure measurements are combined with odometry constraints to construct a pose graph, which is then further optimized. In this study, we evaluate only the registration performance, treating each local trajectory or scene graph as a rigid body representation.}

\newcommand*{\varii}{\ensuremath{\mathcal{I}}}

\begin{figure*}[t]
    \centering
    \includegraphics[width=1.95\columnwidth]{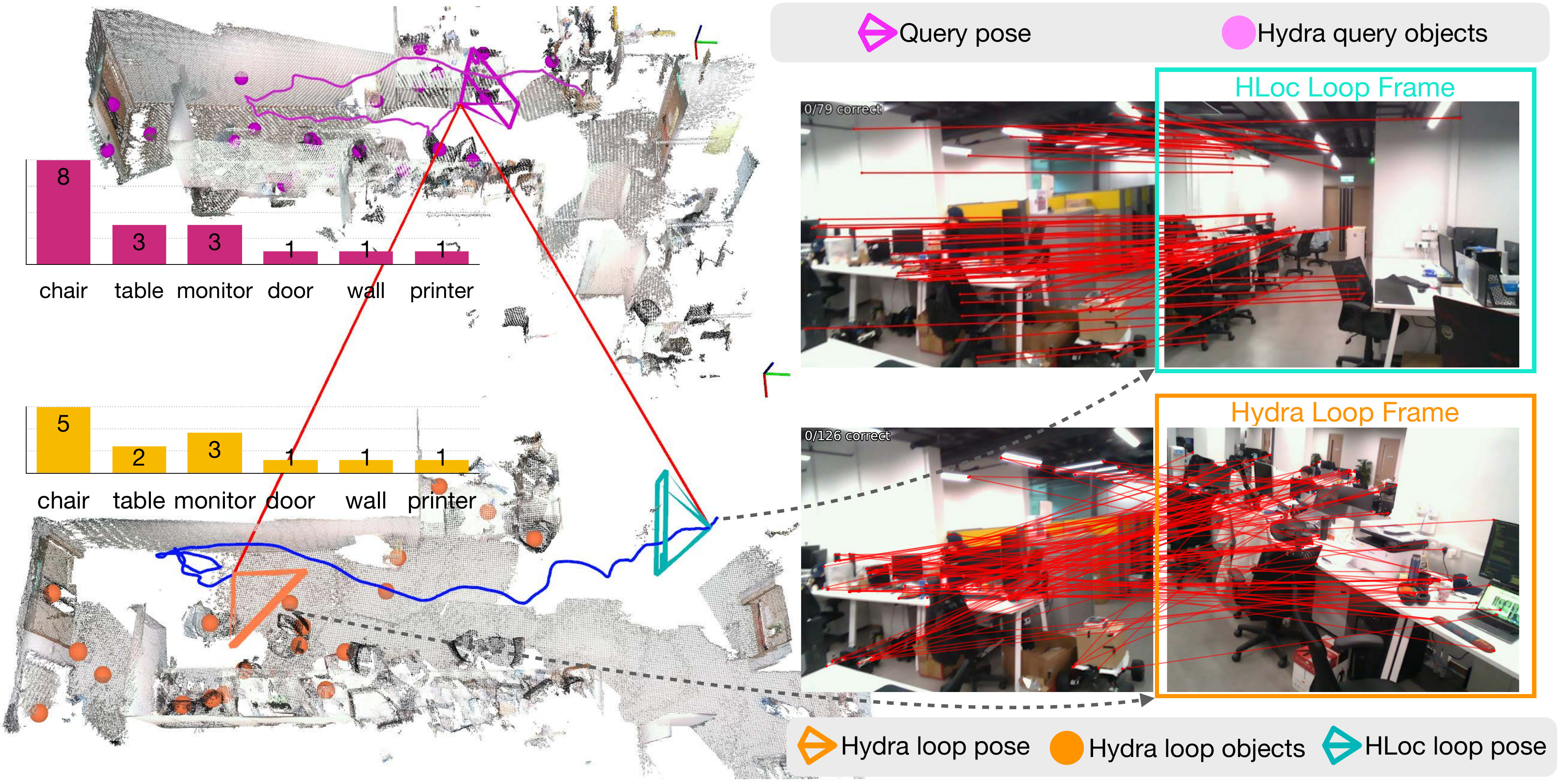}\vspace{-0.2cm}
    \caption{\revise{Loop closure visualization of Hydra and HLoc at a sampled query frame. On the left, 3D maps from the two agents are rendered, along with the poses of the query frame and the loop frames. Additionally, the semantic objects used by Hydra are rendered on the 3D maps, while their semantic histograms are separately displayed. On the right, the image matching results for the query frame are presented. At the query frame, HLoc achieves RTE at $15.1$m and Hydra achieves RTE at $17.4$m, both fail the registration.}}\label{fig:da-sglue}
    \vspace{-0.5cm}
\end{figure*}

\subsubsection{Running rate}
We run Hydra, HLoc, and SG-Reg in every $10$ keyframe. 
\revise{To ensure a fair comparison, we standardize the number of query frames across all methods. Specifically, even if a method identifies too few correspondences, we still execute the registration process to predict a transformation. As a result, the number of query frames for each method remains nearly identical.}

\subsubsection{Robust Pose Average}
Since a single estimation at a query frame can be noisy, we run the robust pose average from Kimera-Multi\cite{tian2023kimera_multi} in multiple consecutive query frames. Specifically, the predicted poses at the query frame and the previous $X$ queried frames are collected and sent to the robust pose average function, where $X$ is the selected window size. The pose average module generates a refined pose as the final result. 
It is designed to enhance the stability of the method. If multiple relative transformations are predicted with a few outliers, pose averaging should improve the performance. On the other hand, if the predictions are robust, the robust pose average module brings limited enhancement. Thus, it can be used to verify the stability of the predicted relative transformations.

\subsubsection{Qualitative results}

We start the evaluation by visualizing a sample loop closure detection frame. \revise{As shown in Fig. \ref{fig:da-sglue}, Hydra recalls a false loop closure. The semantic histogram at the query frame and the false loop frame are closely similar, resulting in an ambiguous hand-crafted descriptor. Their DBoW descriptors are similar as well. So, Hydra matches them in the global retrieval step. Then, their ORB features are also falsely matched. It demonstrates a false loop closure in large viewpoint differences. The images observing different subvolumes create two closely similar hand-craft descriptors.}
Similarly, HLoc fails at the same query frame, as shown in Fig. \ref{fig:da-sglue}. It falsely matches an unrelated image due to its similar appearance.
On the other hand, our method searches correspondences between 3D scene graphs, \revise{which tolerates vastly larger viewpoint differences}. Once the scanned 3D scene graphs overlap, our method can successfully register the scenes. 

\setlength{\figwidth}{0.65\columnwidth}
\newcommand{\rulesep}{\unskip\ \vrule\ }

\newdimen\figrasterwd
\figrasterwd\textwidth
\setlength{\subfigwidth}{0.66\columnwidth}
\begin{figure*}[h]\centering
    \includegraphics[width=1.8\columnwidth]{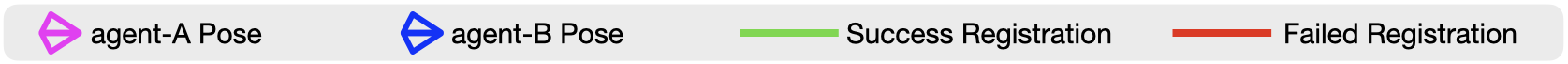}

    \raisebox{1.5\height}{\makebox[0.01\textwidth]{\rotatebox{90}{\makecell{\scriptsize Original}}}}
    \includegraphics[width=\subfigwidth]{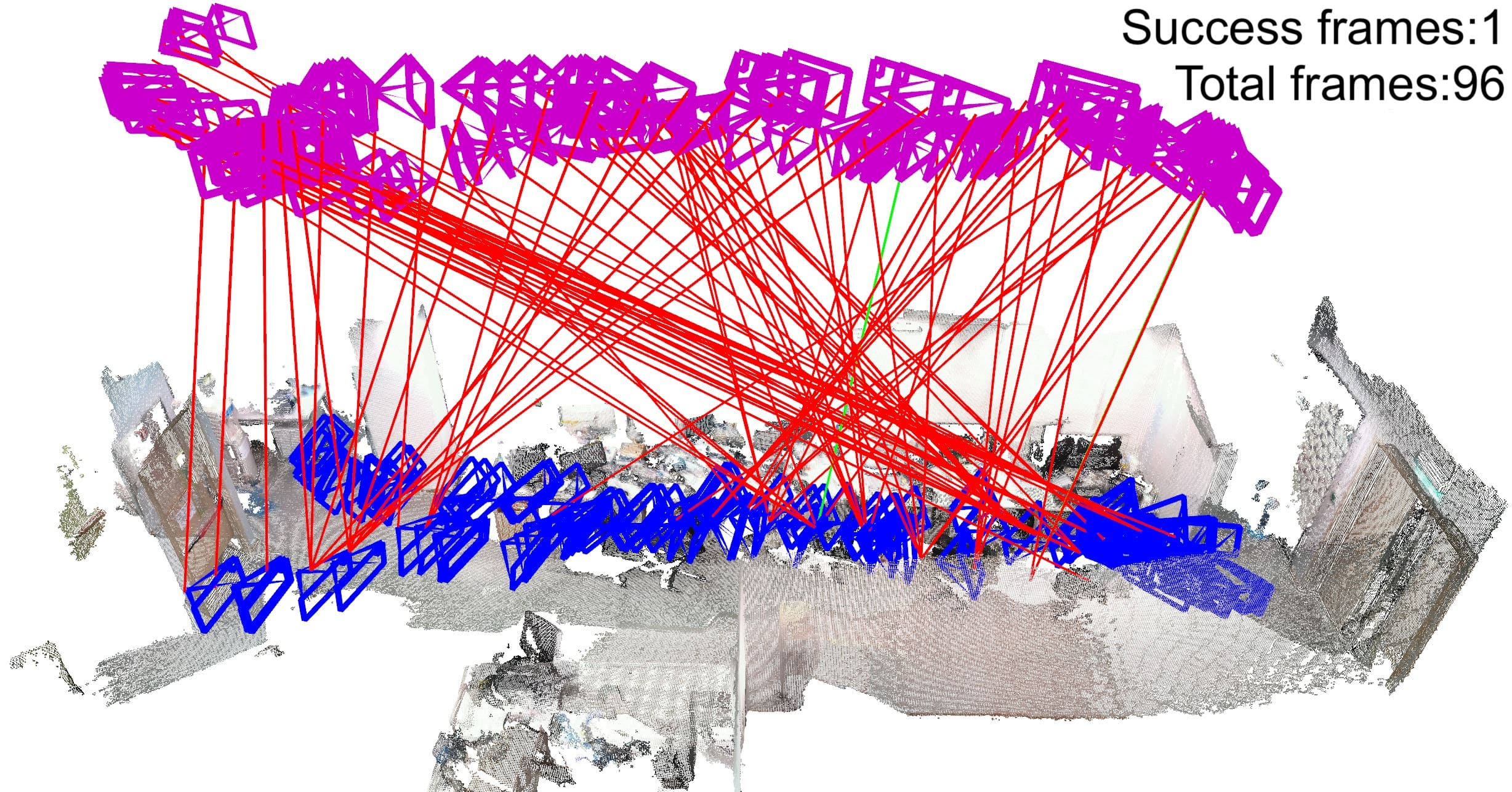}
    \includegraphics[width=\subfigwidth]{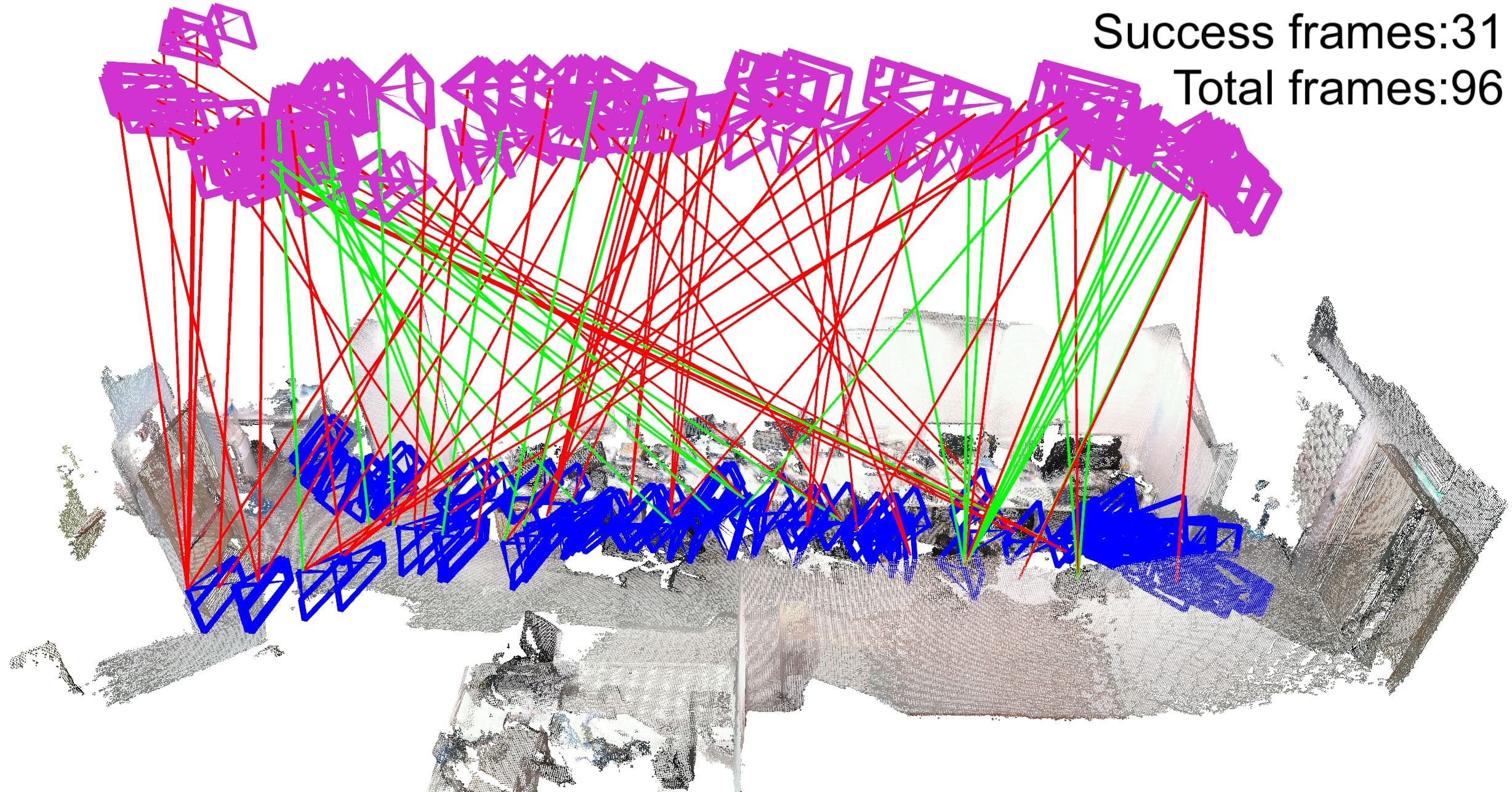}
    \includegraphics[width=\subfigwidth]{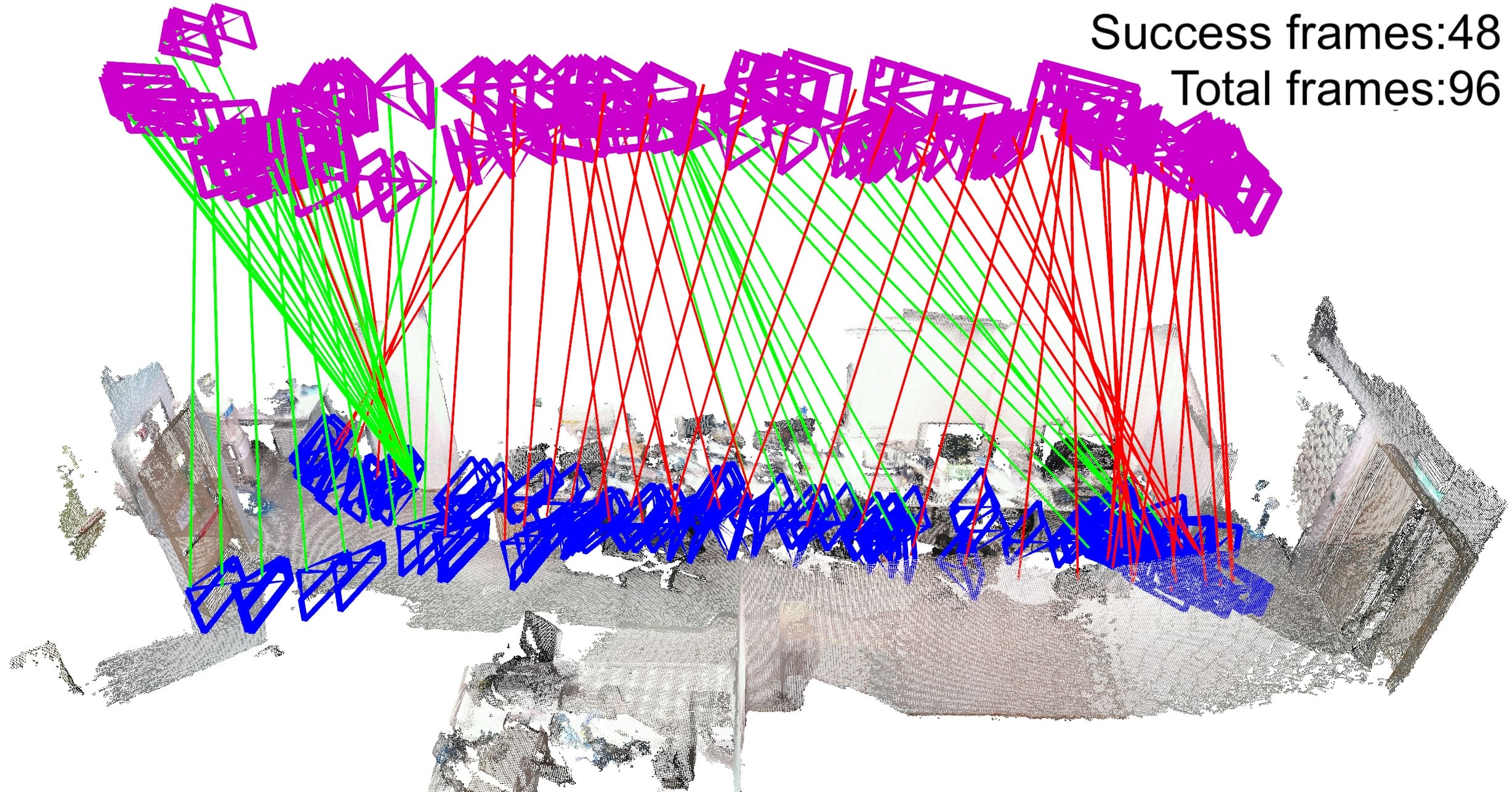}

    \raisebox{0.8\height}{\makebox[0.01\textwidth]{\rotatebox{90}{\makecell{\scriptsize Pose Avg@3}}}}
    \includegraphics[width=\subfigwidth]{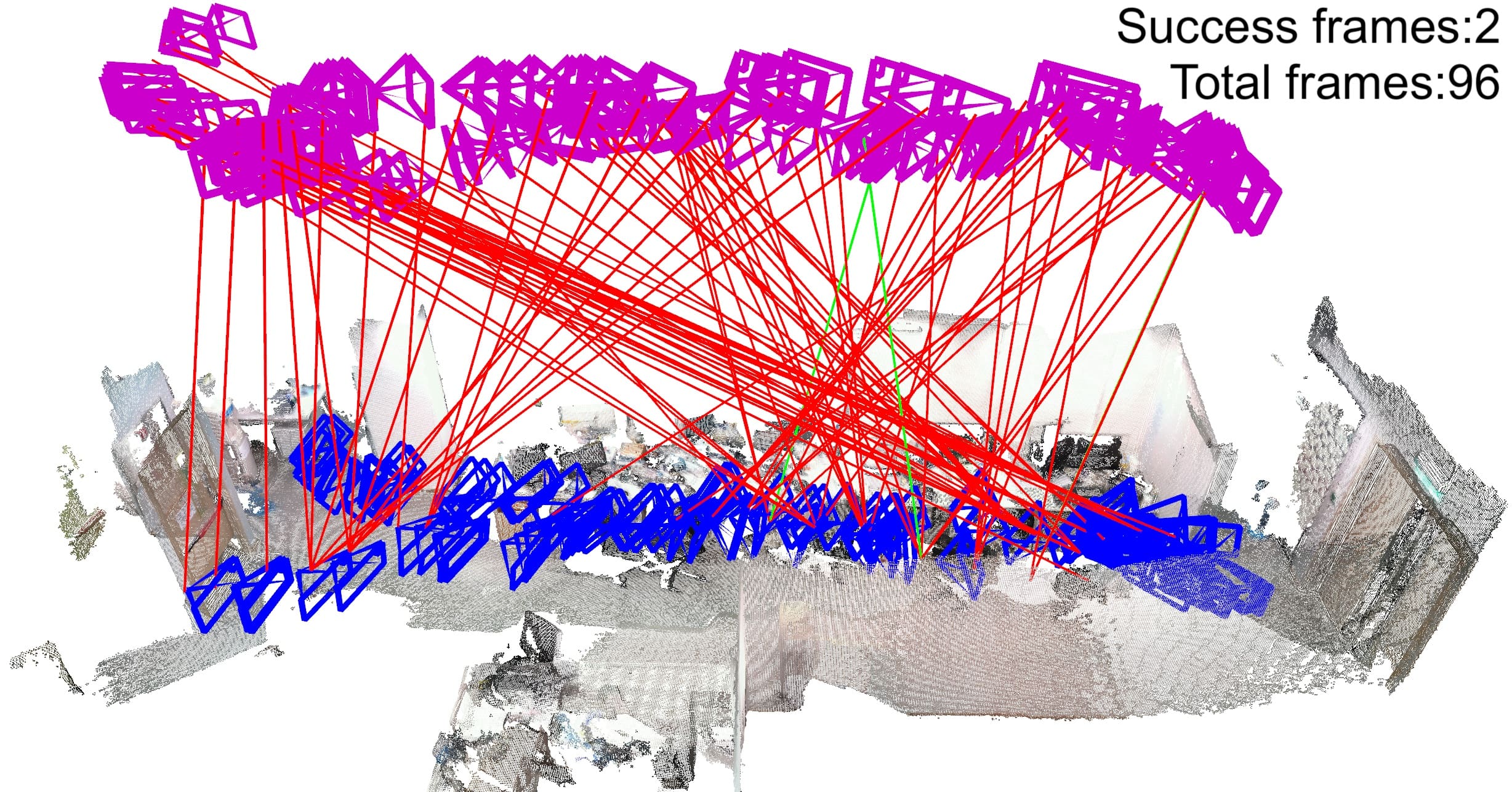}
    \includegraphics[width=\subfigwidth]{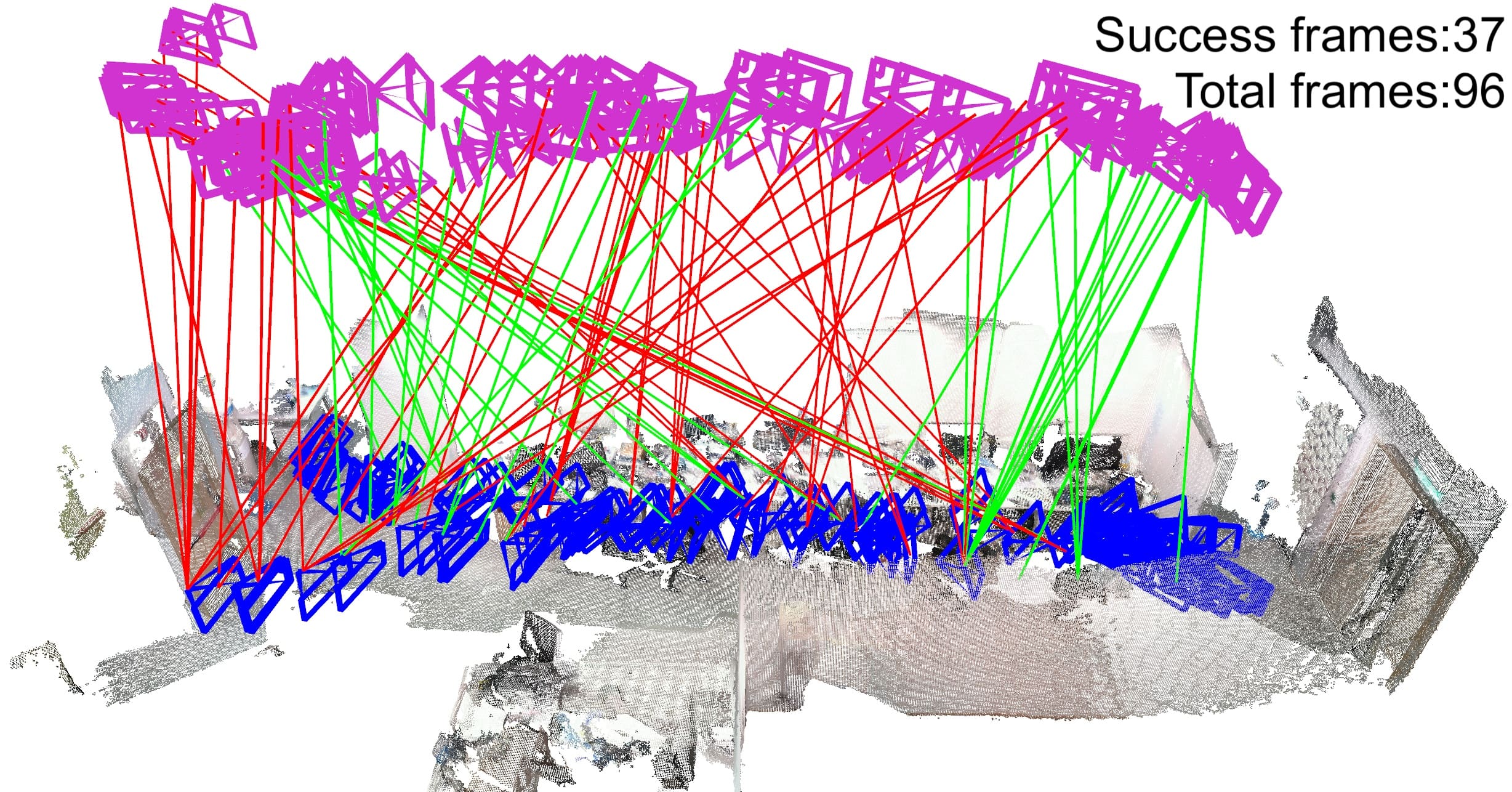}
    \includegraphics[width=\subfigwidth]{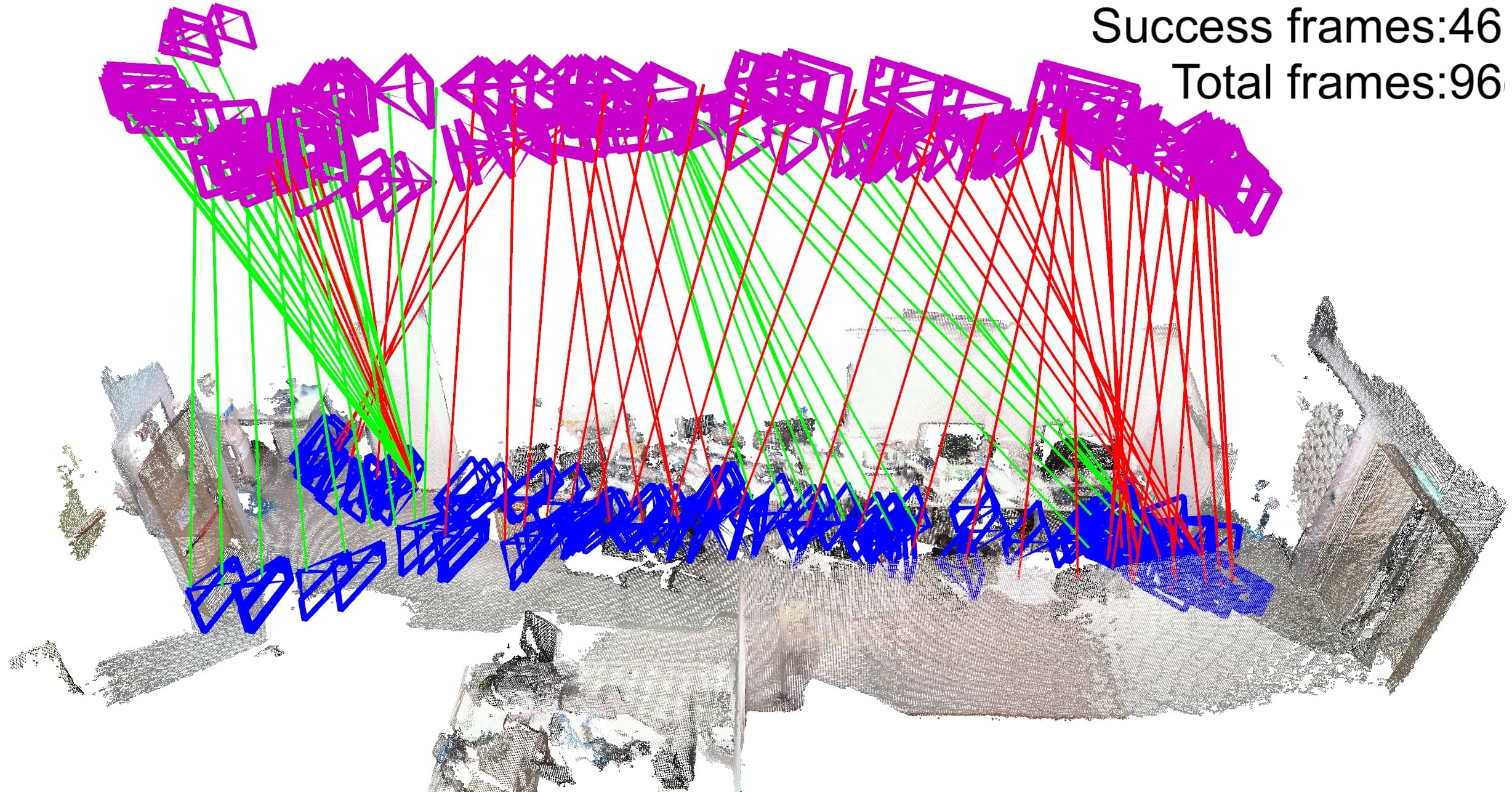}
    \makebox[\subfigwidth]{\small (a) Hydra}
    \makebox[\subfigwidth]{\small (b) HLoc}
    \makebox[\subfigwidth]{\small (c) SG-Reg (Ours)}
    \caption{Registration visualization in an office scenario. The first row shows the original results, while the second row shows the results after pose averaging using three nearest measurements. Each registration result is represented in a line. A registration is successful if its predicted camera pose has a $\text{RTE}<0.2m$ and $\text{RRE}<5^\circ$. For a better visualization, we align the camera poses with ground-truth transformation and incorporate a vertical translation}
  \label{fig:swarm-result}\vspace{-0.5cm}
\end{figure*}
We further visualize the registration results on a 3D point cloud map with sequences of camera poses. As shown in Fig. \ref{fig:swarm-result}, our method registers the scenes at a higher success rate than Hydra and HLoc. 
Hydra fails most of the query frames due to its hand-crafted semantic descriptor and the classical ORB-based image matching. The hand-crafted semantic descriptor can not capture the spatial distribution of the objects. So, it frequently matches the false positive images. Moreover, their ORB-based image matching suffers from a significant viewpoint difference. Thus, most of its querying frames fail.

HLoc's success rate is much higher than that of Hydra. However, it still generates around $60\%$ failed registration frames. The two agents move in vastly different trajectories, generating significant different viewpoints. NetVLAD and LightGlue are both affected by the viewpoint differences. So, their success rate is less satisfied.

Another interesting phenomenon is that our query frames with success registration are spatially close to each other, as shown in Fig \ref{fig:swarm-result}. This is because our method requires a longer initialization. At the beginning of the sequence, our integrated scene graphs have very few semantic nodes and a low overlap ratio. As the scene graphs grow, they gradually have a larger intersection. Hence, nearly all of our query frames failed initially, while those in the latter half of the sequence were mostly successful. Conversely, as shown in Fig \ref{fig:swarm-result}, the successful loop closure frames in HLoc are spatially distributed in variances. 

To verify the stability of the three algorithms, we conducted a robust pose-averaging experiment to refine the predictions. According to our previous introduction, the pose averaging rejects outliers in the estimated transformation. Therefore, when the algorithm is robust enough to noise, pose averaging has a minimal impact. When the algorithm has drastic randomness, its accuracy can be improved after pose averaging. 

As shown in Fig. \ref{fig:swarm-result}, pose averaging has a minimal impact on Hydra due to its low success rate. In contrast, HLoc achieves a higher success rate after pose averaging, demonstrating that HLoc is significantly affected by random noise. This also highlights the effectiveness of the pose averaging module in estimating a reliable pose from multiple predictions with variations. However, as shown in Fig. \ref{fig:swarm-result}, pose averaging has little effect on our results. We attribute this to the inherent stability of our method: while it requires a relatively higher scene overlap and adopts a conservative approach to registration, it consistently achieves accurate and stable results without significant jitter.

Considering HLoc is a powerful baseline, we explain the main differences between HLoc and our method.
Our method first performs object-level registration based on semantic center points and further does point-level refinement on the objects. Therefore, our method utilizes significantly fewer feature points than  HLoc and occupies a much smaller transmission bandwidth. However, our method requires more data for initialization. When the scene IoU is too low, our method may be unable to find overlapping objects, leading to registration failures. On the other hand, HLoc uses more feature points and requires greater computational complexity yet increases the possibility of registration success in low repetition scenarios. 

In the following comparison with HLoc, we aim to verify whether our method can achieve similar or even better accuracy than HLoc with much less communication bandwidth and faster inference speed.

\subsubsection{Quantitative results}
As shown in Table.\ref{tab:pose_ate}, Hydra recalls very few success frames. It is because Hydra focuses more on \revise{the back-end of} scene graph optimization, which involves bundle adjustment\cite{Kimera2020Rosinol} and mesh optimization. In this benchmark, we evaluate the \revise{registration accuracy at a single loop closure measurement}, which makes Hydra less competitive. Specifically, Hydra suffers from limited descriptiveness due to its hand-craft semantic descriptor, frequently false retrieval images from DBoW2, and inaccurate image matching by the brute-force method. 

HLoc is a competitive baseline with which to compare.
Surprisingly, our success rate is higher than that of HLoc before conducting pose averaging. After the pose averaging refinement, as explained earlier, HLoc improves while our method remains unaffected. Therefore, HLoc shows a better success rate. However, our method always maintains an advantage in accuracy. This means that our method tends to be conservative but precise. HLoc performs registration even when the IoU is small, which may result in matching results with significant bias.

\begin{table}[ht]
    \centering
    \begin{tabular}{m{2em}|c c|c c|m{3em} m{2.5em}}
        \toprule
        & \multicolumn{2}{c|}{Original} & \multicolumn{2}{c|}{Pose Avg@$3$} & \multicolumn{2}{c}{Pose Avg@$5$}\\
         &  SR (\%) & RTE(m) & SR (\%) & RTE(m) & SR (\%) & RTE(m) \\
        \hline 
         Hydra & 3.7 & $0.108$ & 4.7 & 0.111 & 4.1 & 0.113\\
         HLoc & ${31.7}$ & $0.118$ & $\mathbf{34.1}$ & $0.121$ & $\mathbf{35.6}$ & $0.119$\\
         \textbf{Ours} & $\mathbf{32.4}$ & $\mathbf{0.106}$ & $32.0$  & $\mathbf{0.105}$ & $31.5$ & $\mathbf{0.107}$ \\
        \bottomrule
    \end{tabular}
    \vspace{+0.1cm}
    \caption{Evaluate registration success rate (SR). We report the averaged RTE from the succeed registration. Pose Avg@$X$ refers to the robust pose average using $X$ frames.}\label{tab:pose_ate}\vspace{-0.4cm}
\end{table}

\begin{figure}[ht]
    \centering
    \includegraphics[width=0.9\columnwidth]{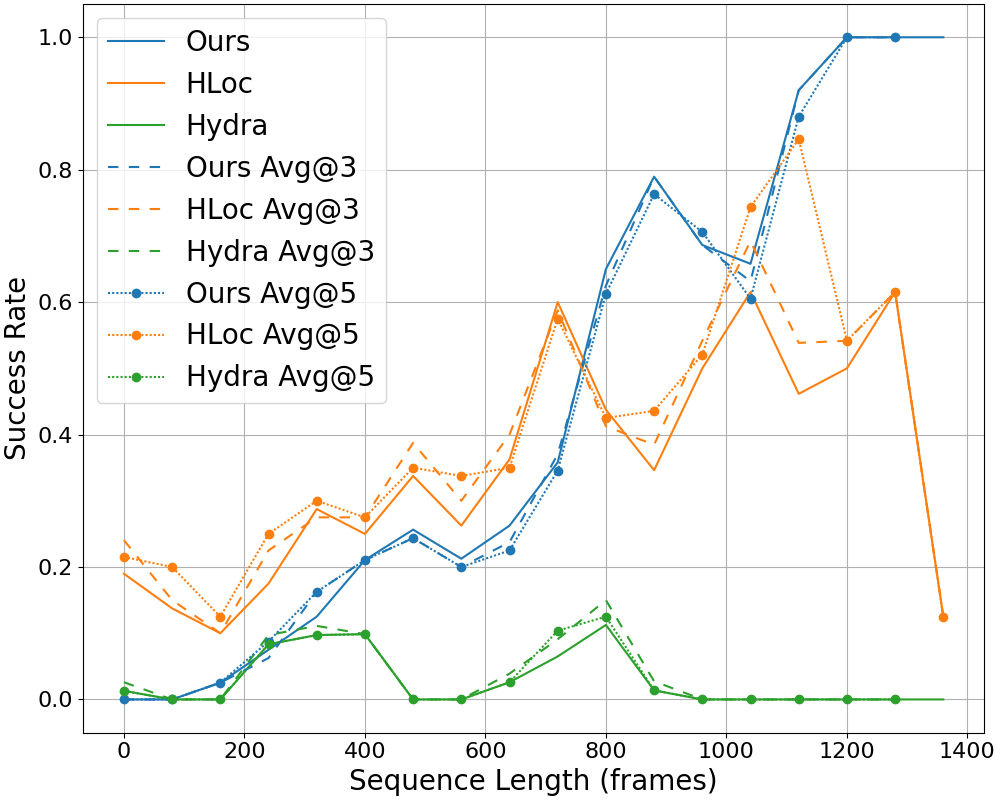}
    \caption{The Loop closure success rate over the sequence length. Avg@$X$ refers to the robust pose averaging in $X$ consecutive loop frames.}\label{fig:sucess_rate}\vspace{-0.5cm}
\end{figure}

The experiment results also align with our initial estimation of the advantages and disadvantages of HLoc and our method. As shown in Fig. \ref{fig:sucess_rate}, HLoc can provide registration earlier and at a higher registration success rate when the scene overlap is low. However, it is easily affected by noise and can hardly cope with large changes in perspective. Our method requires a long initialization time, but it can ensure the stability of registration accuracy, pose a sustained growth in success rate, and is robust to significant relative viewpoints. Considering that our method uses much less bandwidth, its drawbacks are completely acceptable.


Meanwhile, we point out that SG-Reg is not adversarial with image registration methods. In future works, they can be combined into a unified factor graph to optimize.

\subsubsection{Communication Bandwidth}
A significant advantage of our method is the communication bandwidth. As shown in Table \ref{tab:comm}, our method only takes $3.9\%$ of the communication bandwidth that HLoc would take at each query frame. Even if we exclude the RGB images from HLoc, we require less than $11\%$ bandwidth of theirs. The enhanced communication efficiency is because of the sparsified scene representations. We broadcast the encoded node features in the coarse message, which requires an extremely low data bandwidth. We publish a coarse message in every registration frame and only publish a dense message if the other agent requests it. It reduces the amount of dense messages that need to be published.


\begin{table}[ht]
    \centering
    \begin{tabular}{c c c c c c}
        \toprule
         & Message & Number & Dimension & Bandwidth \\
         \hline
         \multirow{4}{*}{Hydra} & SG-Descriptor & 1131 & 339 & $1,497$ kB\\
           & ORB & 622k & 32 & $77,746$ kB \\
           & RGB & 1,131& $640\times480$ & $1,017,900$ kB \\
           & \underline{Overall} & - & - & 970 kB/Frame \\
         \hline
        \multirow{4}{*}{HLoc} & NetVLAD& 1131 & 4096 &$18,096$ kB \\ 
        & SuperPoint & 460k & 259 & $463,864$ kB\\
        & RGB & 1,131 & $640\times480$ & $1,017,900$ kB\\
        & \underline{Overall} & - & - & $1326.1$ kB/Frame \\
        \hline
        \multirow{3}{*}{\textbf{Ours}} & Nodes & $17$k & 132 & $8,472$ kB \\
         & Points & $2918$k & 4 & $40,347$ kB \\
         & \underline{Overall} & - & - & $\mathbf{52}$ \textbf{kB/Frame} \\
        \bottomrule
    \end{tabular}
    \caption{Communication bandwidth comparison. Our messages contain the semantic nodes and points. Their message format is introduced in Sec \ref{sec-multiagent}. We report the accumulated number of each representation. The corresponding averaged bandwidth is in the \underline{overall} rows.}
    \label{tab:comm}\vspace{-0.3cm}
\end{table}

To \revise{further evaluate the influence of communication rate and registration accuracy, we run an ablation study. It defines a minimum interval time for publishing dense messages. We change the interval time and run the experiment in 4 settings. Results from each setting are summarized in Table. \ref{tab:dense_msg}.} 
\begin{table}[h]\centering
    \begin{tabular}{c c c c c}
        \toprule
         & \multicolumn{2}{c}{\underline{Averaged interval time(s)} } & \multirow{2}{*}{Success rate} & \multirow{2}{*}{BW} \\
         & Coarse message & Dense message & & \\
         \hline
        A & 0.98 s & None & 34.9\% & 10.1 KB\\
        B & 0.98 s & 9.5 s & 57.8\% & 109.1 KB \\
        C & 0.98 s & 5.4 s & 62.3\% & 163.6KB\\
        D & 0.98 s & 3.4 s & 63.3\% & 223.7 KB\\
        \bottomrule
    \end{tabular}
    \caption{\revise{Ablation study of communication strategy. Setting A turns off the dense message broadcast and fully relies on the coarse messages. BW refers to the averaged bandwidth over the query frames.}}\label{tab:dense_msg}\vspace{-0.3cm}
\end{table}

As \revise{shown in Table \ref{tab:dense_msg}, from setting A to setting D, the success rate gradually increases as the interval time between publishing dense messages decreases. This demonstrates that dense messages significantly improve the likelihood of successful registration. However, the success rate approaches its optimal level at a certain communication rate, beyond which further improvements are marginal. In this experiment, we selected the setting of Group C for its optimal balance between registration success and communication efficiency.}

\subsubsection{Coarse-to-fine Vs dense matching} \label{sec:dense_msg}

\revise{To investigate how the coarse scene graph matching improves registration performance, we conducted an ablation study by disabling {coarse-to-fine matching} and directly applying {dense matching}. Specifically, the dense matching method brute-force matches point features between the two scenes, while the coarse-to-fine method (ours) matches point features only between matched semantic nodes.}

In \revise{the \underline{coarse-to-fine point matching}, we select the point features from the matched nodes $\varpmM$ and build $\{ \Tilde{\varzZ}^A,\Tilde{\varzZ}^B\}$. We then compute the similarity matrix:}\vspace{-0.2cm}
\begin{equation}\label{eq:similarity}
    \mathbf{S}^Z = g(\Tilde{\varzZ}^\varaA,\Tilde{\varzZ}^\varbB)
    \vspace{-0.2cm}
\end{equation}
\revise{where $g(\cdot)$ is the similarity function in equation(\ref{eq:dotprod}). In \underline{dense matching}, it concatenates the point features in all the semantic nodes to construct $\{ \Tilde{\varzZ}^A,\Tilde{\varzZ}^B\}$ and run equation (\ref{eq:similarity}). They build the point features $\{ \Tilde{\varzZ}^A,\Tilde{\varzZ}^B\}$ differently, where we illustrate their computational complexity in Table \ref{tab:brute_dense}.}

\begin{table}[h]
    \centering
    \begin{tabular}{m{1em} c c c}
        \toprule
        & \multicolumn{2}{c}{{Input Dimensions}} & {Dot Product} \\
         & $\Tilde{\varzZ}^A$ & $\Tilde{\varzZ}^B$ & Operations \\
         \hline
         DM & $(|\mathcal{A}| \cdot K_p) \times d_z$ & $(|\mathcal{B}| \cdot K_p)\times d_z$  & $(|\mathcal{A}|\cdot K_p) \cdot (|\mathcal{B}| \cdot K_p)$\\
         CFM & $|\mathcal{M}|\times K_p \times d_z$ & $|\mathcal{M}|\times K_p \times d_z$ & $|\mathcal{M}|\cdot K_p \cdot K_p$ \\
        \bottomrule
    \end{tabular}
    \caption{\revise{Analysis complexity in computing the similarity matrix. DM denotes dense matching, while CFM denotes coarse-to-fine matching.}}\label{tab:brute_dense}\vspace{-0.3cm}
\end{table}
As \revise{shown in TABLE \ref{tab:brute_dense}, dense matching requires heavier dot product operations. To verify the analysis, we select a scene to perform dense matching. The result shows that the dense matching results in a high outlier ratio and requires higher computation FLOPS than the coarse-to-fine matching. Please check our supplementary material for details.

}

\vspace{-0.4cm}
\subsection{Runtime Analysis }\label{sec-runtime}
\subsubsection{Runtime in 3RScan}

\begin{table}[ht]
    \centering
    \begin{tabular}{c c c c}
        \toprule
         Steps & GeoTransformer & SG-PGM & \textbf{Ours} \\
         \hline
         Point\&Shape Backbone & 23.6 & 60.1& 20.3 \\
         GNN\&Attention Layers & 71.2 & 59.0 & 2.45 \\
         Match Layers & 91.7  & 52.7  &  28.6 \\
         Pose Estimator & 15.7 & $10.9$ & $90.5^\dagger$ \\
         \textbf{Total}& 202.3 & $182.7$& $\mathbf{142.7}$ \\
         \bottomrule
    \end{tabular}
    \caption{\revise{Compare runtime in millisecond (ms) on 3RScan benchmark.} We mark our pose estimation time using $^\dagger$ because it runs on CPU, while the baselines estimate pose on GPU. }\vspace{-0.3cm}\label{tab:rio_runtime}
\end{table}
As shown in Table. \ref{tab:rio_runtime}, our GNN and match layers are significantly faster than the baseline methods. It benefits from our sparsified scene representation. Our triplet-based GNN learns on the semantic nodes, while the geometric transformer learns on the densely distributed superpoint. So it runs much faster. This enhanced GNN speed is consistent with our FLOPs analysis in Fig. \ref{fig:memory}(b). 

One limitation is that our pose estimator is relatively slower than the baseline. Their estimator runs faster because they run weighted SVD\cite{besl1992method} on GPU. They organize the point correspondences in an array of patches. An SVD thread solves a patch of correspondences. Since the patches are gathered into a batch, solving it on GPU is very fast. On the other hand, our pose estimator handles a global set of point correspondences, which is processed on the CPU. Since the correspondences are dense, solving the maximum clique and the robust estimator can be relatively slow.
Despite the slower runtime in pose estimation, our total registration is still faster than the baseline methods. If we only consider the network inference, our inference is $3$ times faster than the baselines.



\subsubsection{Runtime in SLAM}
In the two-agent SLAM, we save our median results and rerun each step on the PyTorch platform. Then, we record the time consumption on PyTorch and compare it with HLoc. It is designed to compare the inference time in the same platform since our network is deployed by Libtorch in the SLAM system. Also, as HLoc runs offline, it ensures we compare with HLoc fairly.

As shown in Table. \ref{tab:sgslam_time}, LightGlue is the most time-consuming step in HLoc. 
Global matching can also be slow when the sequence is long, as HLoc searches candidate images within the timestamp interval $[0, t_q]$. Consequently, longer sequences result in a greater number of images to search through. For instance, in a frame towards the end of the sequence, global matching can take $51.5$ ms, which is twice the average time shown in Table \ref{tab:sgslam_time}.

Compared to HLoc, our total computation time is significantly faster. As shown in Table \ref{tab:sgslam_time}, our total processing time is considerably less than that of HLoc, even when receiving dense messages. Within our framework, the shape network and pose estimator account for the majority of the inference time due to their inputs being dense points.
For Hydra's loop closure detection, it runs quite fast, including its semantic descriptor construction and similarity calculation. Those operations require low computing resources. Despite its fast inference time, Hydra detects too few loop closures and is a less competitive baseline method.

However, although our key modules run faster than HLoc, we point out that our method can not run in real-time yet. 
One bottleneck is the point cloud pre-processing operation, adapted from GeoTransformer\cite{qin2022geometric}. It uniformly downsamples the point cloud into multiple layers of point cloud. It searches the nearby points at each layer, generating a hierarchical point cloud aggregation structure. Due to the large size of our point cloud and it runs on CPU, it can take up to $900$ ms in one query frame. GeoTransformer and SG-PGM also run the same pre-processing step and take a similar time. Similar to them, we treat the pre-processing operation as a step in the data loader, which is not considered in the runtime. Besides, since data loading time in HLoc is not involved either, we exclude ours in Table\ref{tab:sgslam_time}, which is a fair comparison.
Considering that real-time performance is not our motivation in this paper, we leave it to be better implemented in future work. 

\begin{table*}[ht]
    \centering
    \begin{tabular}{c c|c c|c c c}
         \toprule
         \multicolumn{2}{c|}{\textbf{Hydra}{$^\dagger$}}& \multicolumn{2}{c|}{\textbf{HLoc}} &\multicolumn{3}{c}{\textbf{SG-Reg} }\\
         Steps &Time(ms) &Steps & Time(ms) & Steps & Coarse(ms) & Dense(ms) \\
        \hline
         Object\&DBoW & 0.5 & NetVLAD& 18.9 & Shape Network & - & 38.2 \\ 
         Global Match& 14.3 & Global Match& 23.5 & T-GNN & 2.1 & 2.1 \\
         ORB Extract & 23.8 & SuperPoint& 15.4 & Node Match & 1.5 & 2.0 \\
         Brutal Match& 0.9 & LightGlue& 173.6 & Point Match & - & $22.4$ \\
         RANSAC-PnP & 10.3 & RANSAC-PnP & $15.6^\dagger$ & Pose Estimator & $12.1^\dagger$ & $43.9^\dagger$ \\
         \hline
        \textbf{Total}& 49.6 & \textbf{Total}& 247 & \textbf{Total} & $15.7$ & $108.6$\\
         \bottomrule
         \multicolumn{7}{c}{(a)}
    \end{tabular}
    \begin{tabular}{c c}\vspace{+0.08cm}
        \\
        \\
        \\
        \toprule
         &{Time (ms)} \\
        \hline
        RAM$^*$\cite{zhang2023ram} & {$29.8$}\\
        GroundingDINO$^*$\cite{liu2023grounding} & {$130.2$}\\
        EfficientSAM$^*$\cite{xiong2024efficientsam} & {$43.6$}\\
        FM-Fusion\cite{liu2024fmfusion} & {$62.8$}\\
         \bottomrule
         \multicolumn{2}{c}{(b)}
    \end{tabular}
    \vspace{-0.1cm}\caption{(a) Registration time in the two-agent SLAM. We report the average time in each frame. Modules or method marked in $^\dagger$ run on CPU. (b) Semantic mapping time. FM-Fusion runs in real-time, while the foundation models$^{*}$ run offline.}
    \label{tab:sgslam_time}\vspace{-0.5cm}
\end{table*}

\subsubsection{Mapping Runtime}
We report the semantic mapping time for reference. Semantic mapping is a high-level perception function in SLAM systems. Multiple modules use it as a shared front-end \cite{werby23hovsg,gu2023conceptgraphs}. As a result, we excluded its timing from the scene graph registration timing and reported it separately.
In computing the runtime, we run all the foundation models offline and run FM-Fusion in real-time. FM-Fusion loads the pre-generated results from foundation models to construct the scene graph. As shown in Table.\ref{tab:sgslam_time}(b), our semantic mapping module can run in approximately $14$ Hz. With the foundation models, the system should be able to run at $4$ Hz. If \revise{the SLAM system requires the semantic perception running} at a slower rate \cite{hughes2024foundations}, the overall inference speed is acceptable for a semantic SLAM system. 
\vspace{-0.4cm}

\subsection{Limitations}
In the two-agent SLAM benchmark, \revise{we neglect the pose drift for each agent. Since VINS-Mono\cite{qin2018vins} provides accurate odometry over short camera trajectories ($\leq$ 150 m), we assume that the scene graph and trajectory of each agent are rigid bodies. However, a desired multi-agent SLAM system should account for pose drift for each agent and further reconstruct an optimal SLAM representation. To address this limitation, we believe that combining the correspondences from SG-Reg with the image matching constraints from HLoc for optimization is a promising direction. SG-Reg can provide an accurate initial alignment between agents, which can then be followed by scene graph optimization frameworks\cite{chang2023hydramulti,bavle2022sgraph} to correct pose drift for each agent.}

Moreover, we find some failed registration in the long-tailed scenes, such as empty rooms and scenes in symmetric layout. We provide their results in the supplementary material, which can be further improved in the future. 
\vspace{-0.3cm}

\section{Conclusion}
In \revise{this work, we propose a learning-based registration method that aligns two semantic scene graphs without requiring an initial value. The neural network encodes coarse semantic node features, which are fused from their semantic label, local topology with spatial awareness, and geometric shapes. Besides, our network encodes the dense point cloud that are belong to each semantic node. Then, the network searches for correspondences at two levels: coarse semantic nodes and dense point clouds. In the optimization stage, we employ a robust pose estimator to compute the transformation based on these correspondences. Thanks to automatic data generation, we train the network in a self-supervised manner without the need for any ground-truth semantic annotations.
In the evaluation, we compare our method with point cloud registration networks and semantic registration networks. Our approach achieves a higher recall rate, particularly in medium and large scenes. Additionally, our GPU inference requires less than $2.2$ GFLOPS, while the baseline can require up to $1,314$ GFLOPS.
We deploy SG-Reg in a two-agent SLAM system to align the coordinates between the two agents. In real-world indoor scenes, SG-Reg significantly outperforms handcrafted descriptors in Hydra. When compared to the state-of-the-art image matching network HLoc, SG-Reg achieves a similar registration success rate. However, in frames with large viewpoint differences, SG-Reg registers the scene more accurately and robustly than HLoc. Additionally, our communication module requires only 3.9\% of the bandwidth that HLoc demands.
On the other hand, we recognize the system demands heavy front-end modules to reconstruct the semantic maps. In the two-agent benchmark, we ignore the pose drift within each agent and register their maps as two rigid bodies. Despite these limitations, our system is generalizable, offers higher computational efficiency during inference, and communicates data using lower bandwidth.}
\vspace{-0.3cm}

\bibliography{lch}

\begin{IEEEbiography}[{\includegraphics[width=1in,height=1.25in,clip,keepaspectratio]{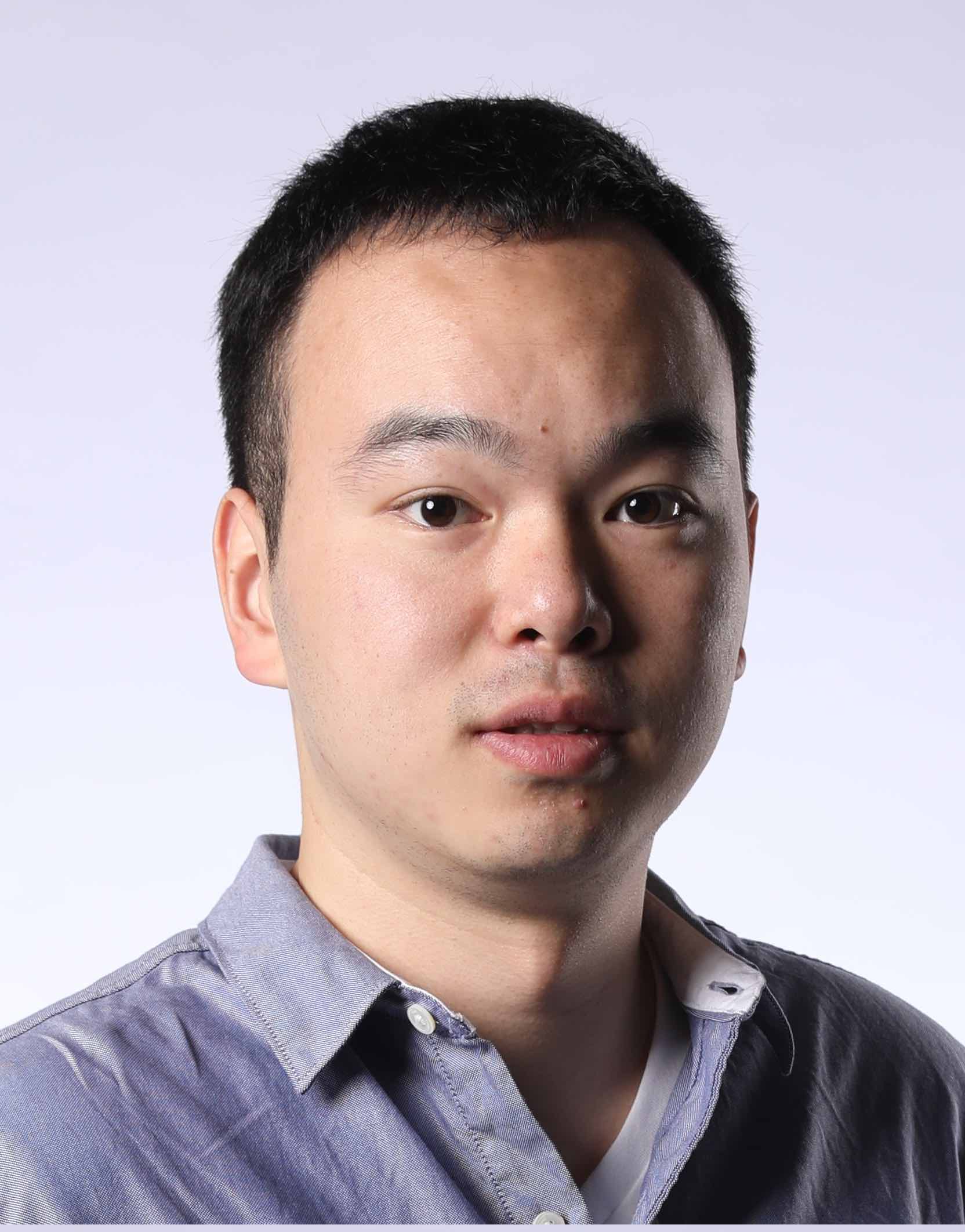}}]{Chuhao Liu} received his B.Eng degree in Electrical and Electronic Engineering from University of Nottingham UK in 2013 and M.Phil in Technology Leadership and Entrepreneurship from Hong Kong University of Science and Technology in 2020. He is currently working toward his Ph.D. degree in Electronic and Computer Engineering at the Hong Kong University of Science and Technology. His research focuses on spatial perception, localization, and mapping.   
\end{IEEEbiography}

\begin{IEEEbiography}[{\includegraphics[width=1in,height=1.25in,clip,keepaspectratio]{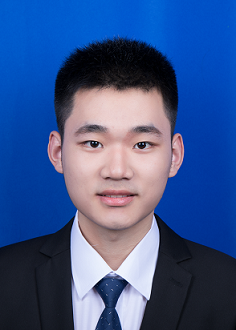}}]{Zhijian Qiao}
Zhijian Qiao received the B.Eng degree in Automation from Northeastern University, Shenyang, China, in 2019, and the M.S. degree in Control Engineering at Shanghai Jiao Tong University, Shanghai, China, in 2022. He is currently working toward the PhD degree at the Department of Electronic and Computer Engineering, Hong Kong University of Science and Technology, Hong Kong. His research interests include the areas of robotics and autonomous driving, with focus on robust state estimation and crowd-sourced mapping.
\end{IEEEbiography}

\begin{IEEEbiography}[{\includegraphics[width=1in,height=1.25in,clip,keepaspectratio]{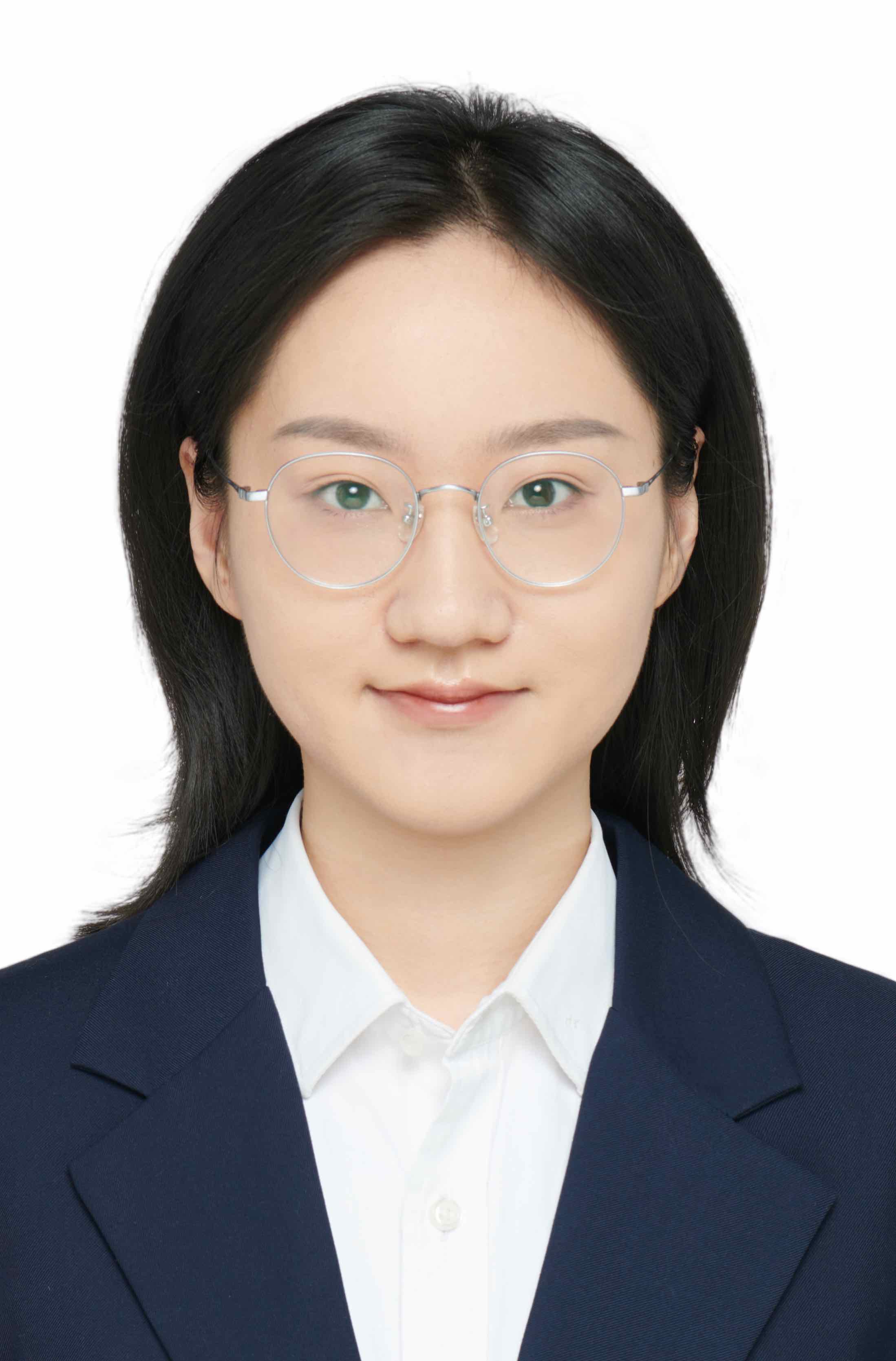}}]{Jieqi Shi} received her B.Eng degree in the School of Electronics Engineering and Computer Science from Peking University, Beijing, China in 2018, and the Ph.D. degree in Electronic and Computer Engineering at the Hong Kong University of Science and Technology in 2024. She is now an Assistant Professor at the School of Intelligence Science and Technology of Nanjing University. Her research interests lie in robotics, with a focus on perception, localization and mapping, and learning-based autonomous navigation.

\end{IEEEbiography}

\begin{IEEEbiography}[{\includegraphics[width=1in,height=1.25in,clip,keepaspectratio]{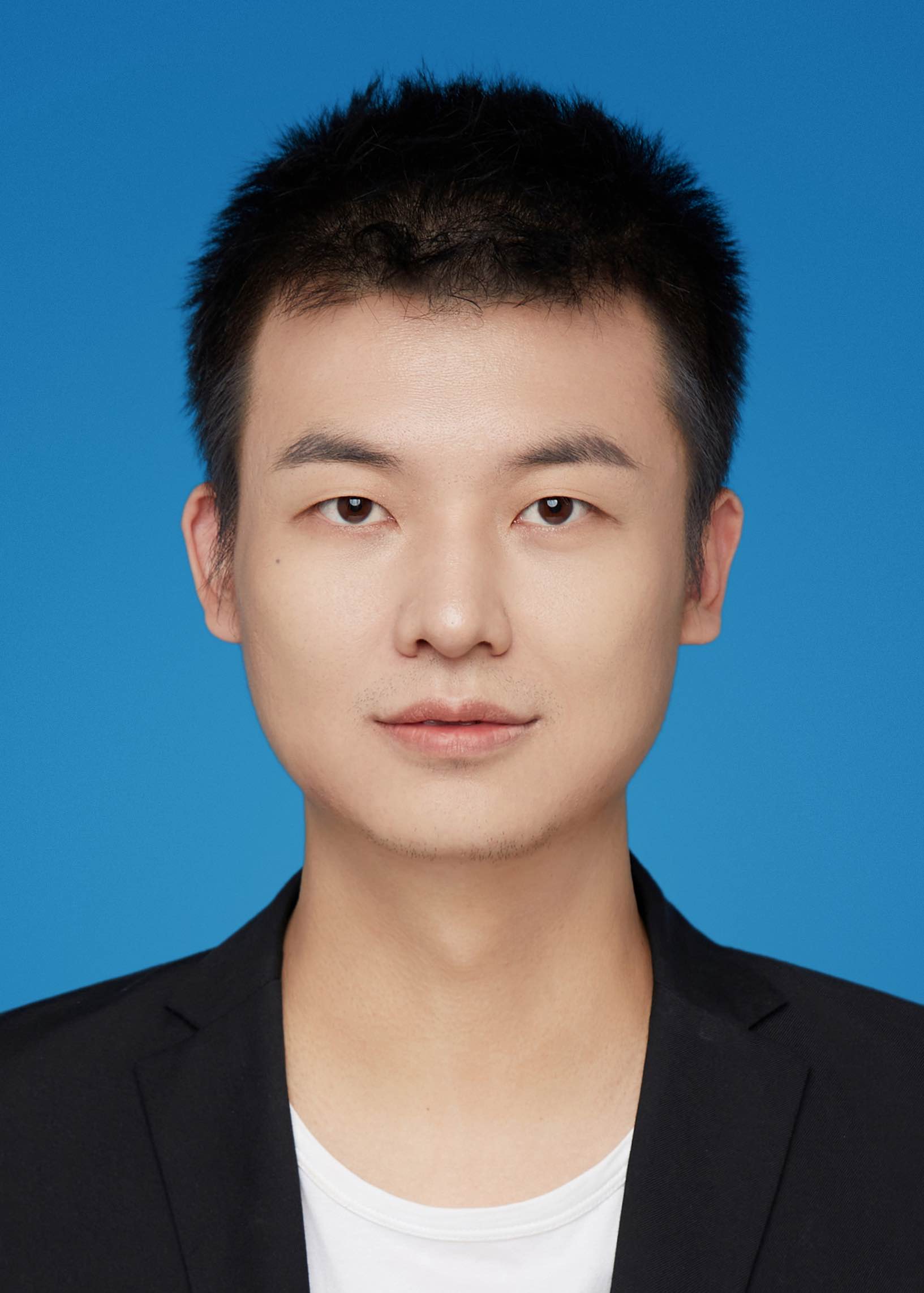}}]{Ke Wang} received his B.Eng degree in the School of Computer Science and Technology from Northwestern Polytechnical University, Xi'an, China in 2013, and the Ph.D. degree in Electronic and Computer Engineering at the Hong Kong University of Science and Technology in 2023. He is now an Assistant Professor at the School of Information Engineering of Chang'an University. His research interests lie in robotics, with a focus on localization and mapping, visual depth estimation, and V2X Cooperation perception.
\end{IEEEbiography}

\begin{IEEEbiography}[{\includegraphics[width=1in,height=1.25in,clip,keepaspectratio]{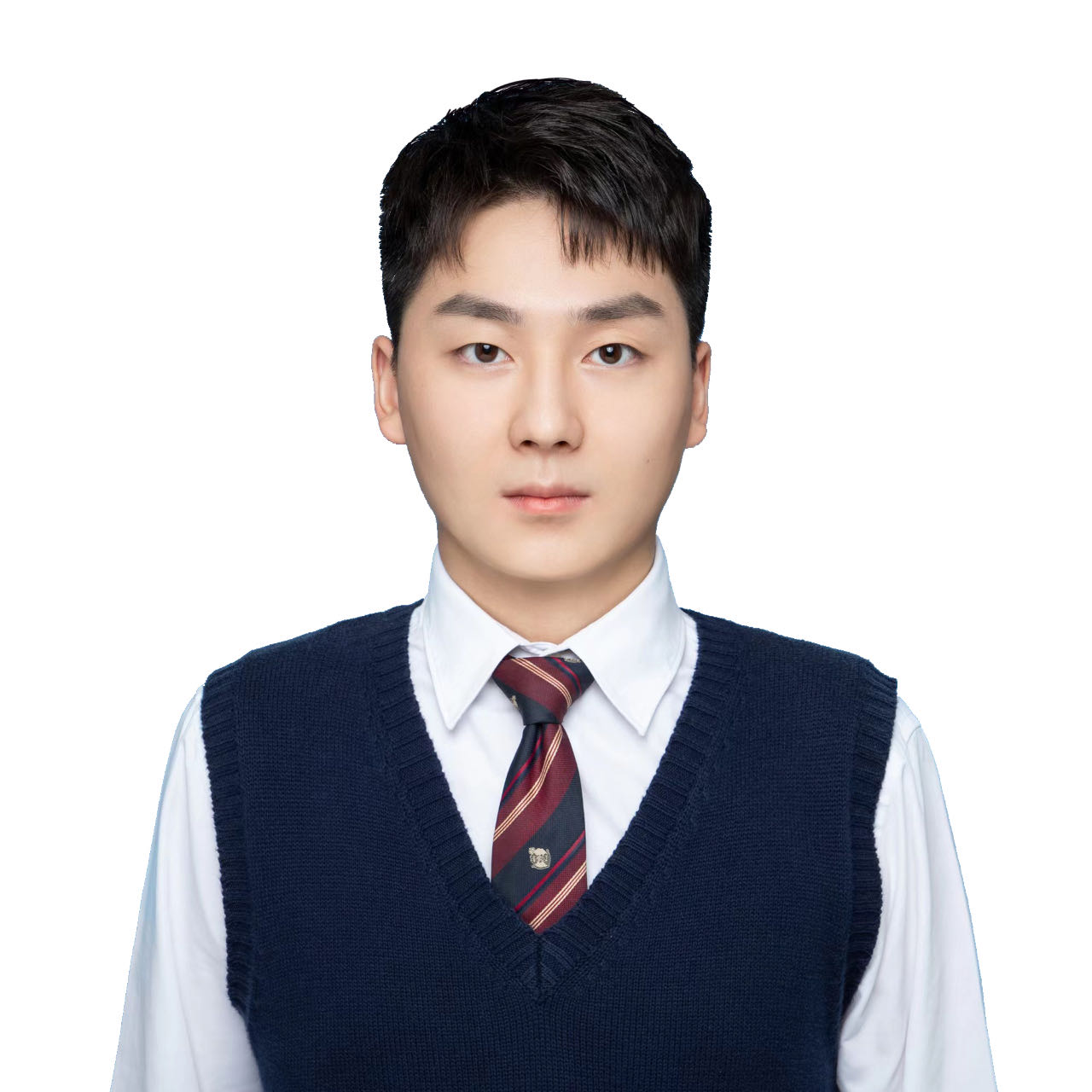}}]{Peize Liu} received the B.Sc. degree in Software Engineering  from
the University of Electronic Science and Technology of China,
Chengdu, China, in 2022. He is currently working
toward the Ph.D. degree with the Hong Kong University of Science and Technology, Hong Kong, under
the supervision of Prof. Shaojie Shen. His research interests include unmanned aerial vehicles, state estimation in aerial swarm, and swarm system.
\end{IEEEbiography}

\begin{IEEEbiography}[{\includegraphics[width=1in,height=1.25in,clip,keepaspectratio]{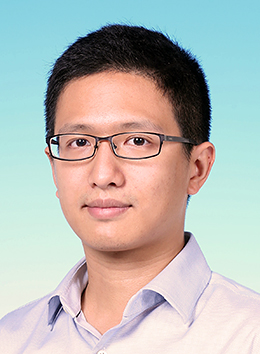}}]{Shaojie Shen} received the B.Eng. degree in electronic engineering from the Hong Kong University of Science and Technology, Hong Kong, in 2009, and the M.S. degree in robotics and the Ph.D. degree in electrical and systems engineering from the University of Pennsylvania, Philadelphia, PA, USA, in 2011 and 2014, respectively.

In September 2014, he joined the Department of Electronic and Computer Engineering, Hong Kong University of Science and Technology, as an Assistant Professor, and was promoted to Associate Professor in July 2020. His research interests include robotics and unmanned aerial vehicles, with a focus on state estimation, sensor fusion, computer vision, localization and mapping, and autonomous navigation in complex environments.
\end{IEEEbiography}

\newpage

%
\appendices
\graphicspath{{supplementary}}

\section{Scene Graph Network}



\subsubsection{Multi-stage Training}
We utilize a multi-stage training strategy to reduce data complexity and improve training efficiency. Firstly, \revise{BERT\cite{devlin2018bert} is pre-trained} and fixed. Secondly, we pre-train the shape encoder in the ScanNet dataset, ensuring the point cloud backbone (KPConv) and the object shape backbone effectively capture geometric features. The shape loss in Equation (\ref{eq:shape}) supervises the pre-training. Thirdly, we train all of the network layers end-to-end, including the triplet-GNN and the hierarchical matching layers. Since the scene graph network encodes multiple modalities, pre-training the features from semantic labels and point cloud benefits the end-to-end training.

\subsubsection{Ablation study}
\begin{table}[h]
    \centering
    \begin{tabular}{c c c c c c}
        \toprule
         & Num. of Correspondences & IR(\%) & PIR(\%) & RR(\%) \\
         \hline
        $K_p=256$  & 492 & 19.6  & 51.6 & 78.0 \\ 
        $K_p=512$  & 313 & 26.8  & 55.0 & 69.0 \\
        $K_p=1024$ & 126 & 36.9  & 55.9 &59.0\\
        \bottomrule
    \end{tabular}
    \caption{Ablation study on sampling points $K_p$ from each semantic node. We evaluate the performance in cross-domain benchmark.}
    \label{tab:ablation_kp}
\end{table}

As shown in TABLE \ref{tab:ablation_kp}, we set parameter $K_p$ to be a larger value. It is the number of points we sampled from each semantic node, as explained in Sec. \ref{sec-shape}. In a larger value of $K_p$, we have three interesting findings. Firstly, the number of correspondences declines. It is because the optimal transport algorithm\cite{Cuturi2013sinkhorn} run multiple normalization operations. A growing size of the similarity matrix suppress the average similarity scores after normalizations. Hence, fewer correspondences are detected. Secondly, the inlier ratio grows higher as it implicitly sets a higher threshold. Thirdly, the final registration recall declines. It implies that the inlier ratio is not the only key factor that decides the registration accuracy. The back-end estimator also needs a larger number of inliers to prune outliers and predict a precise transformation.

\section{Hierarchical communication}
\subsubsection{Dense matching Vs Coarse-to-fine matching}
\begin{figure}[h]
    \centering
    \includegraphics[width=0.8\columnwidth]{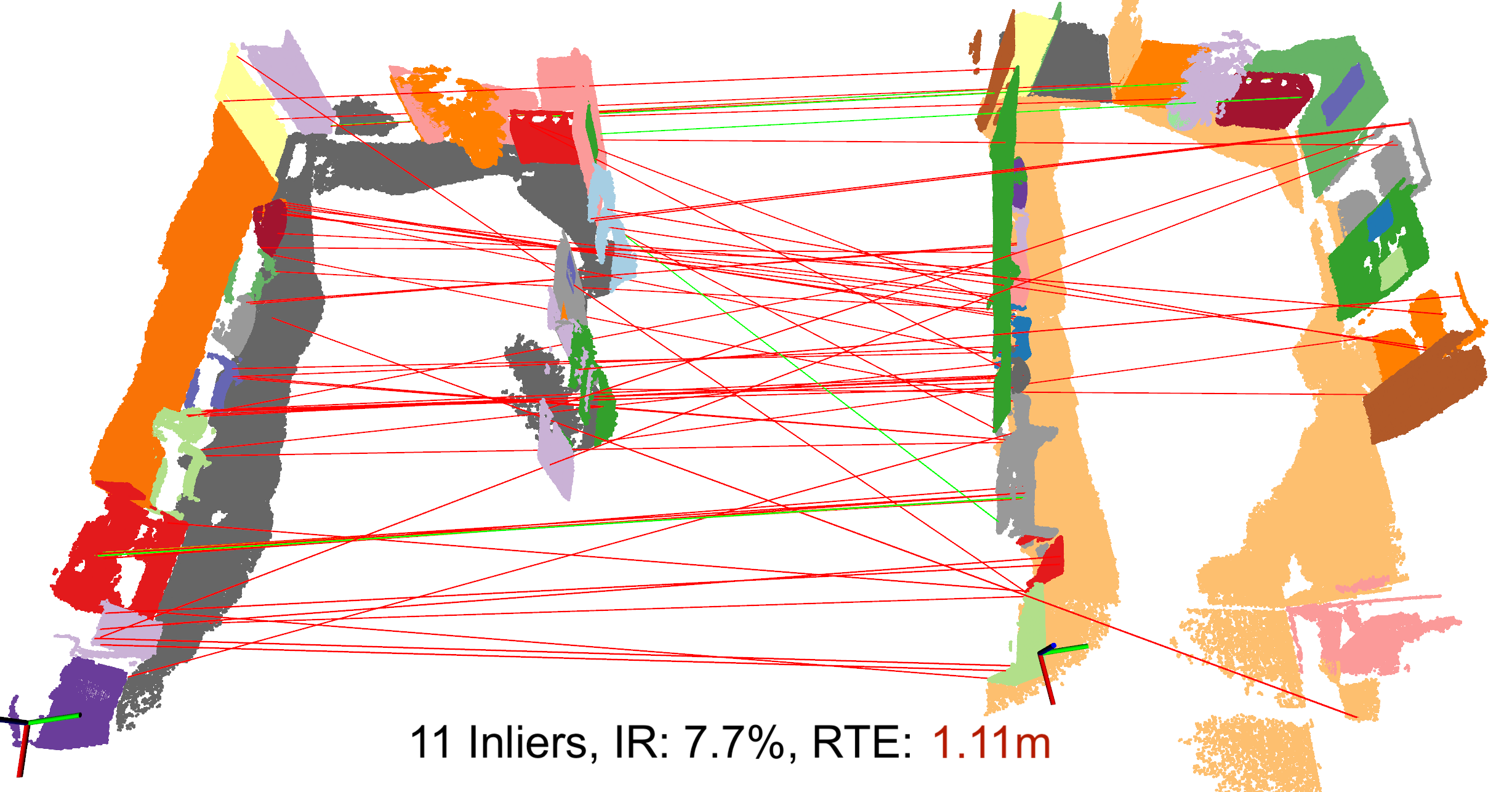}
    \makebox[0.9\columnwidth]{\small (a) Dense matching $\sigma=0.02$; $36,182$ MFLOPS. }
    \includegraphics[width=0.8\columnwidth]{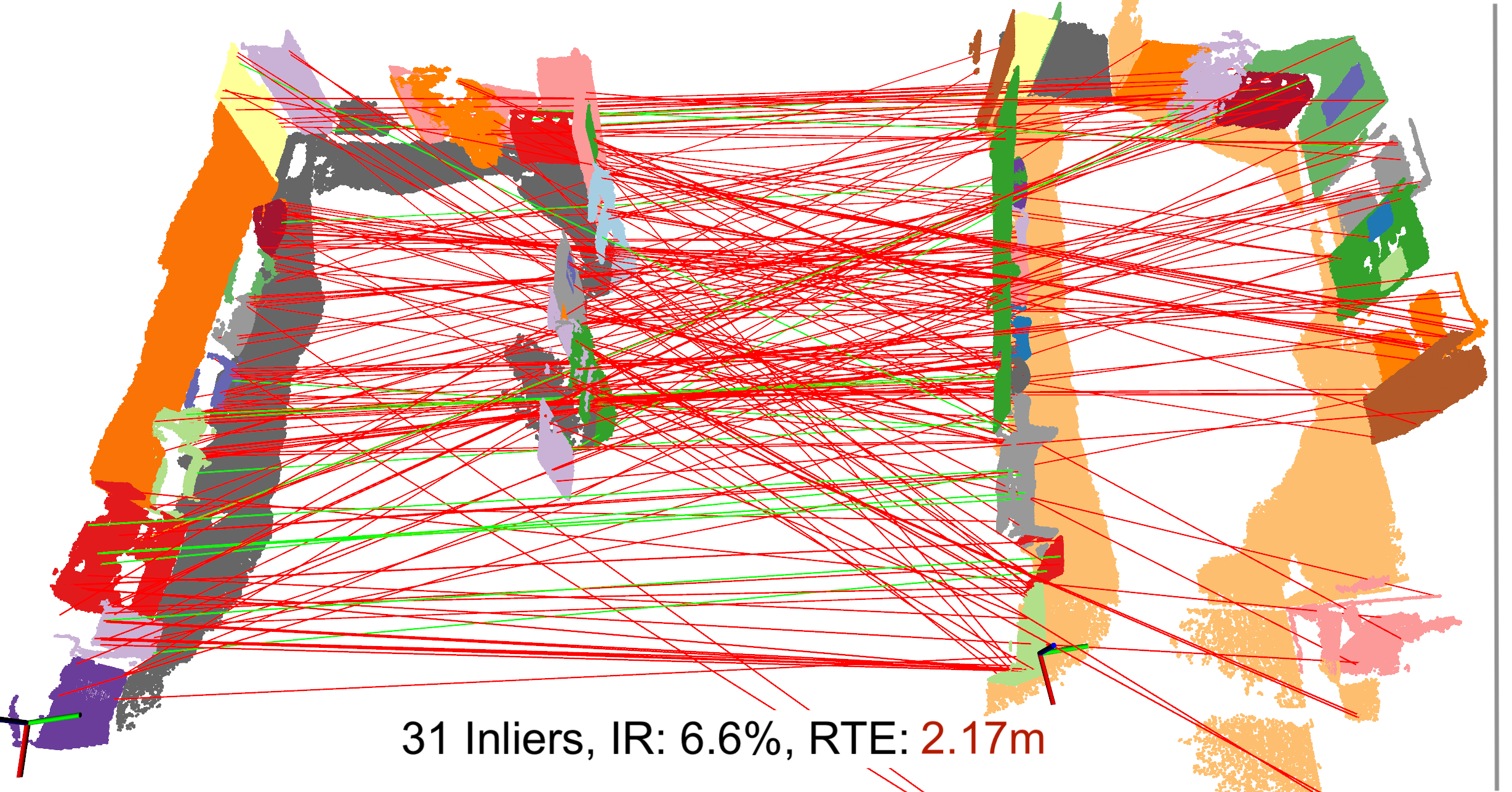}
    \makebox[0.9\columnwidth]{\small (b) Dense matching $\sigma=0.05$.}
    \includegraphics[width=0.8\columnwidth]{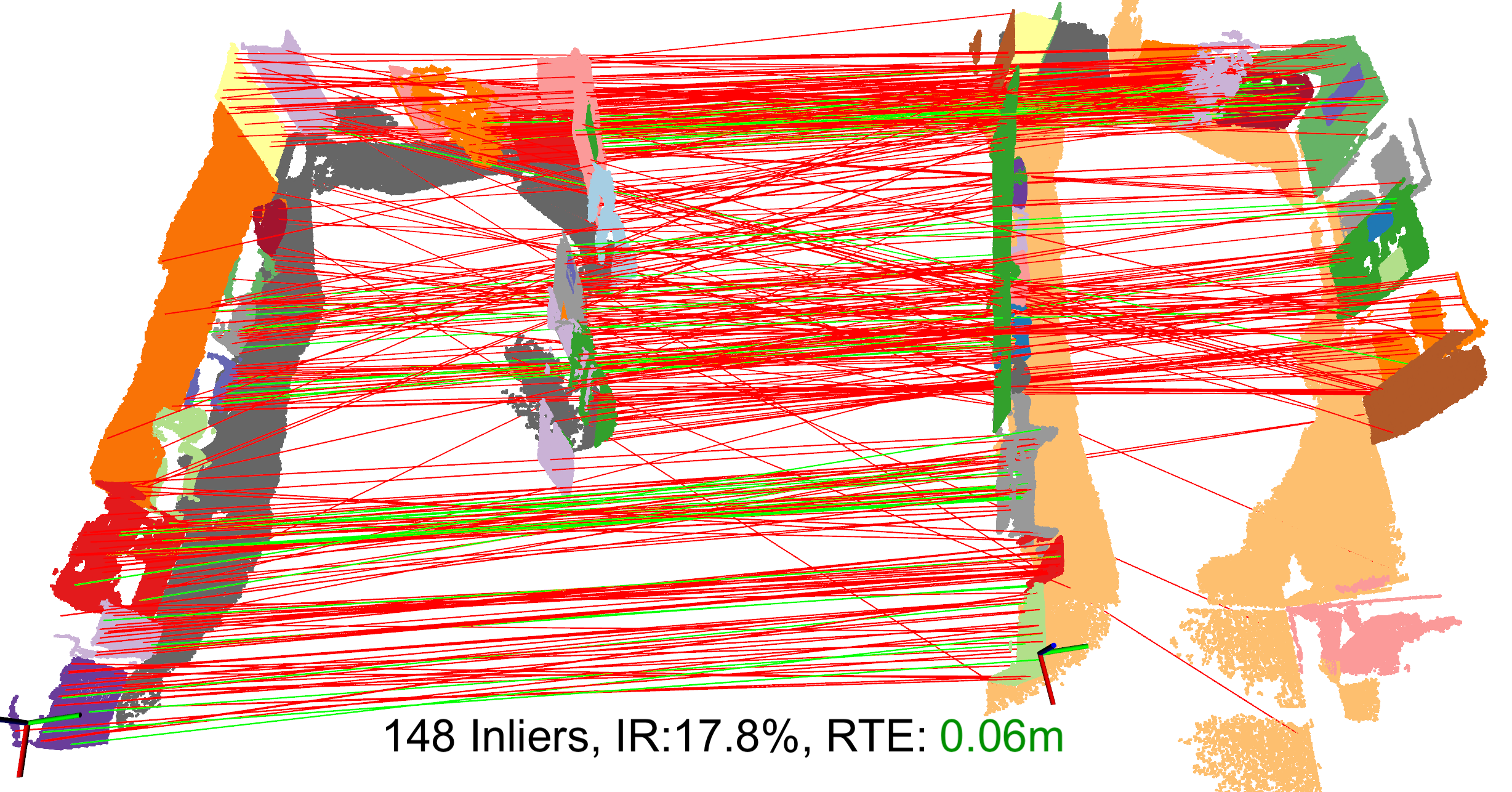}
    \makebox[0.9\columnwidth]{\small (c) Coarse-to-fine (Ours); 602 MFLOPS.}

    \caption{\revise{Dense match points and coarse-to-fine match points.}}
    \label{fig:brute_dense}
\end{figure}
We analysis the advantages of coarse-to-fine matching in Sec. \ref{sec:two_agent}(10). Compared to dense matching, it requires lower computational FLOPS and result in a higher inlier ratio. As shown in Fig. \ref{fig:brute_dense}(a), we evaluate the dense matching in a scene. Because of the large amount of similar geometric features, dense matching approach has a low inlier ratio and very few inliers. We set a higher matching threshold in Fig. \ref{fig:brute_dense}(b) and it can find a larger amount of correspondences. But the inliers ratio is still low. On the other hand, coarse-to-fine approach generates the highest inlier ratio. Its computational FLOPS is also lower, as we analyzed in Sec. \ref{sec:two_agent}(10). 

\subsubsection{How dense matching improves registration}
We provide a visualization result to further support the evaluation in TABLE \ref{tab:dense_msg}. 
\begin{figure}[b]
    \centering
    \raisebox{0.3\height}{\makebox[0.01\textwidth]{\rotatebox{90}{\makecell{\scriptsize Low-overlap}}}}
    \includegraphics[width=0.47\columnwidth]{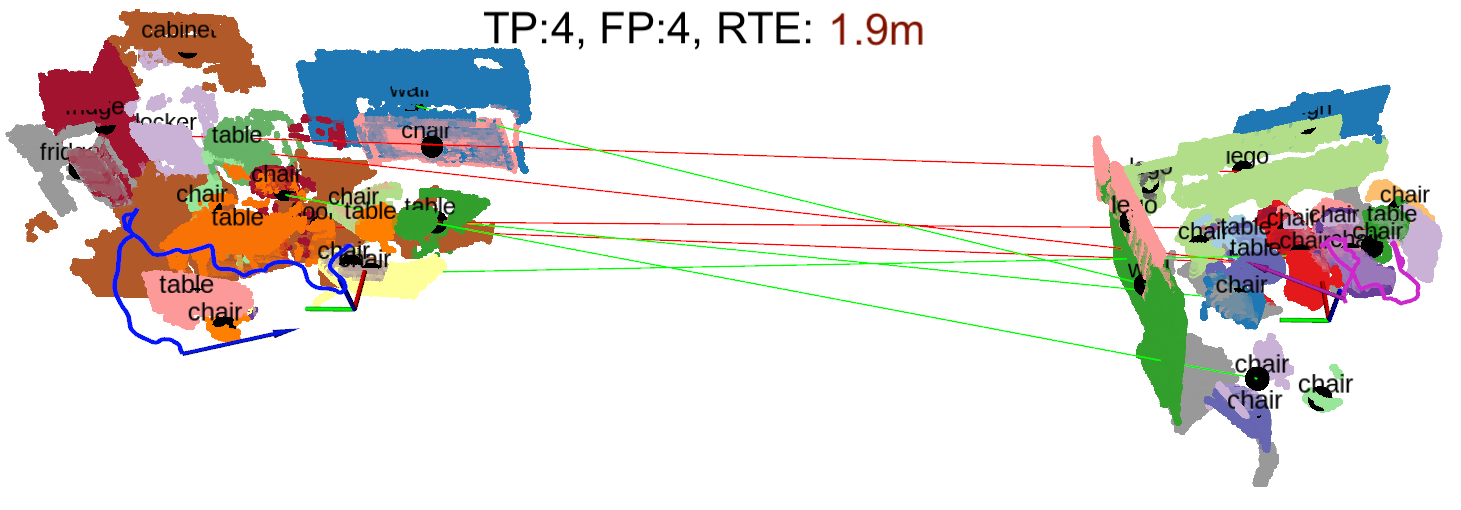}
    \includegraphics[width=0.47\columnwidth]{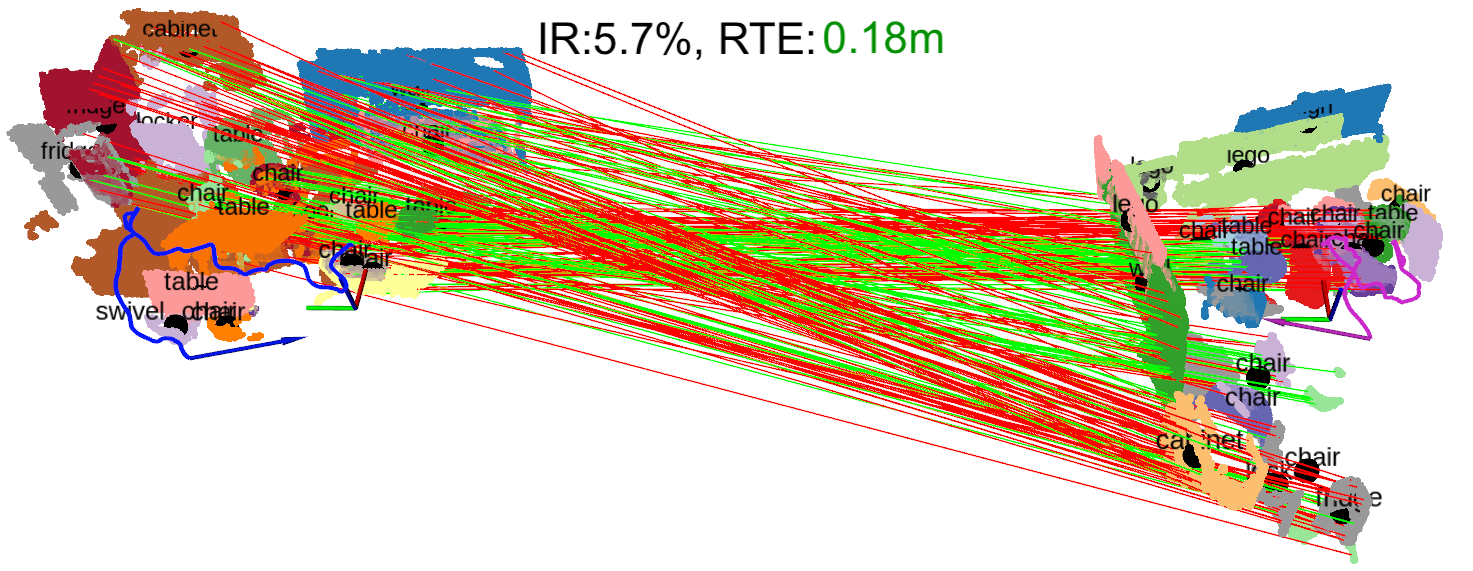}

    \raisebox{0.3\height}{\makebox[0.01\textwidth]{\rotatebox{90}{\makecell{\scriptsize High-overlap}}}}
    \includegraphics[width=0.47\columnwidth]{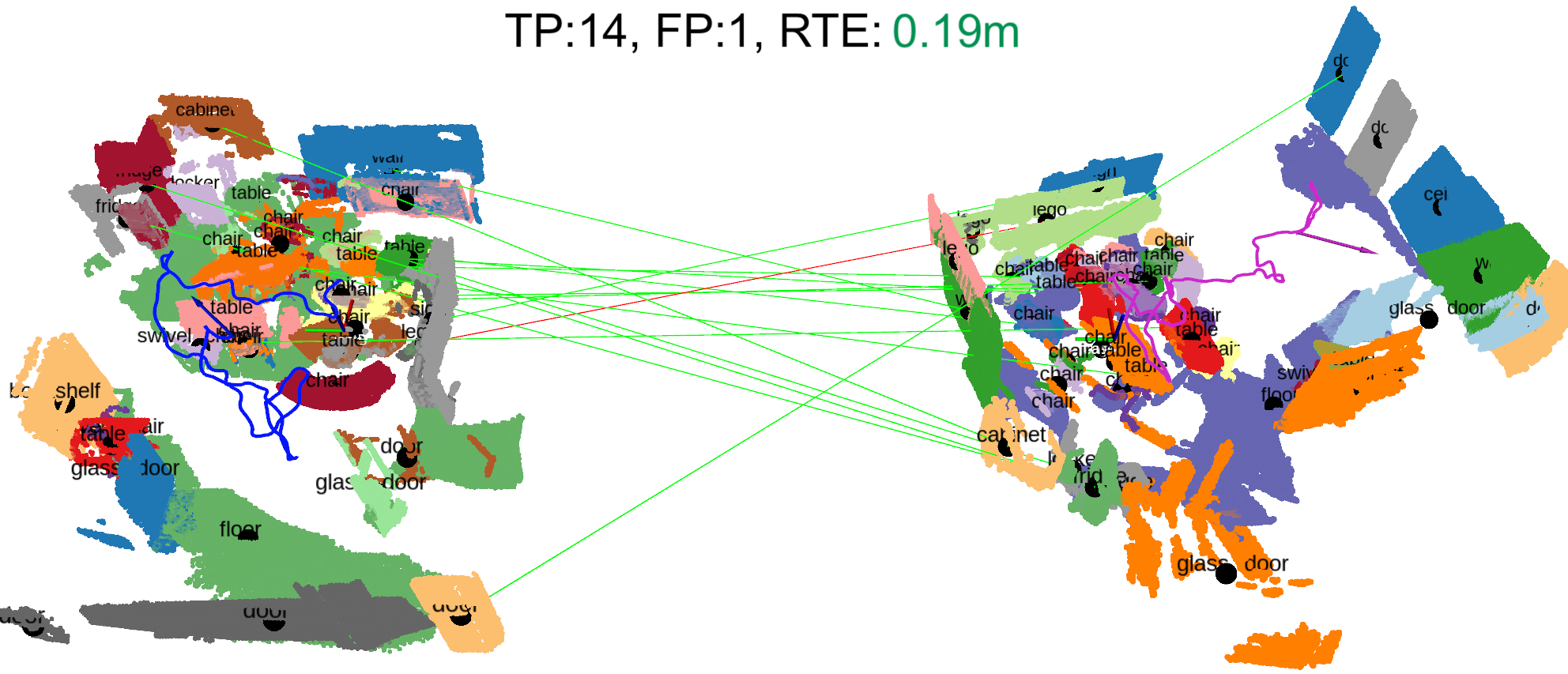}
    \includegraphics[width=0.47\columnwidth]{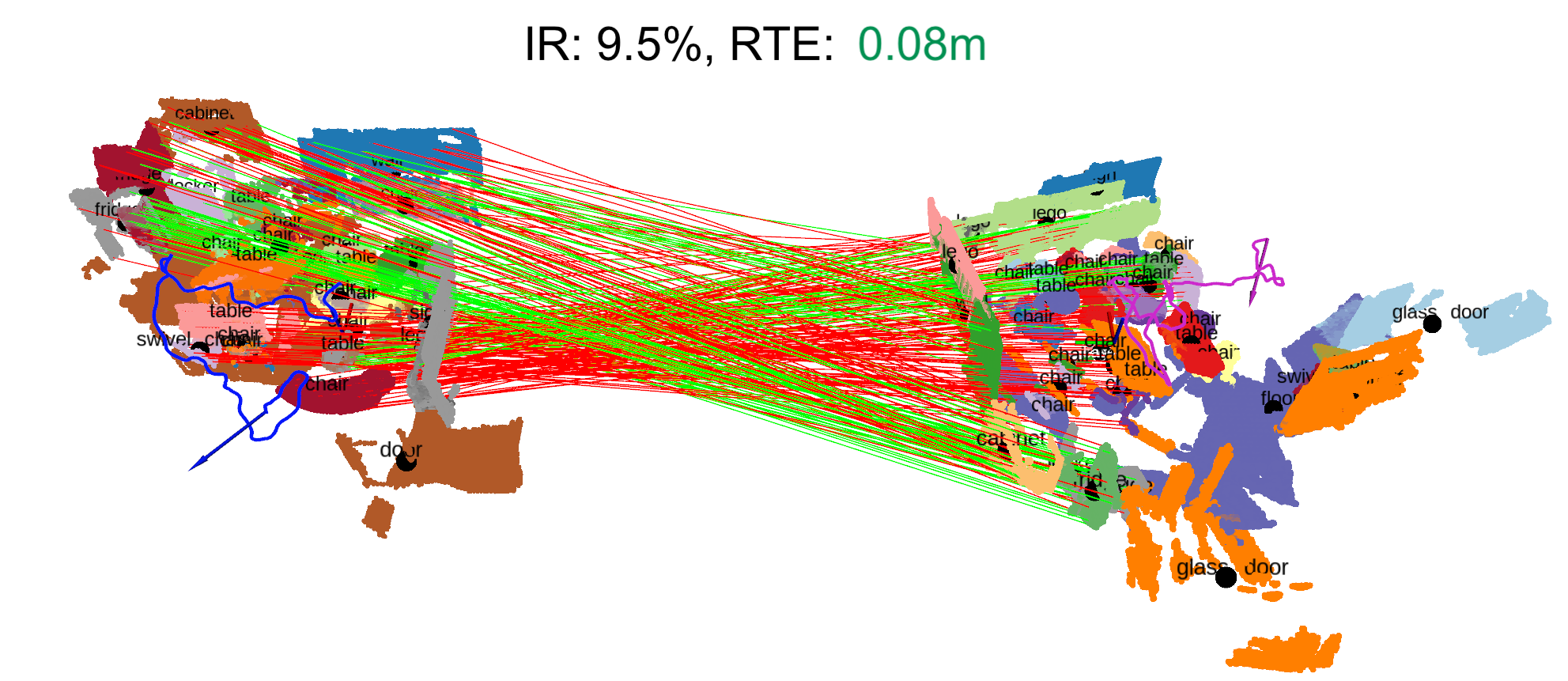}
    \makebox[0.45\columnwidth]{\small (a) Coarse message}
    \makebox[0.45\columnwidth]{\small (a) Dense message}

    \caption{The matching results at a low-overlap scene and a high-overlap scene.}
    \label{fig:dense_msg}
\end{figure}
As shown in Fig. \ref{fig:dense_msg}(a), SG-Reg relies on frames containing only coarse messages for registration. In each frame, the pose estimator uses the centers of matched semantic nodes to align the scenes. While coarse messages enable successful registration in some high-overlap scenes, they are likely to fail in low-overlap scenarios due to the limited number of true positive matches, which poses serious challenges for accurate registration. In contrast, as illustrated in Fig. \ref{fig:dense_msg}(b), dense messages significantly improves SG-Reg’s ability to register low-overlap scenes by providing more measurements utilizing point correspondences. Additionally, dense messages enhance registration accuracy in high-overlap scenes.
This further supports our claim related to TABLE \ref{tab:dense_msg}, that broadcasting dense messages improve registration accuracy.  

\section{Efficiency in high inlier ratio}
In Sec. \ref{sec-pestimator}, we mentioned that high inlier ratio requires a longer computational complexity. To clarify it, we run a simple experiment and show its results. 

\begin{figure}[h]
    \centering
    \includegraphics[width=\columnwidth]{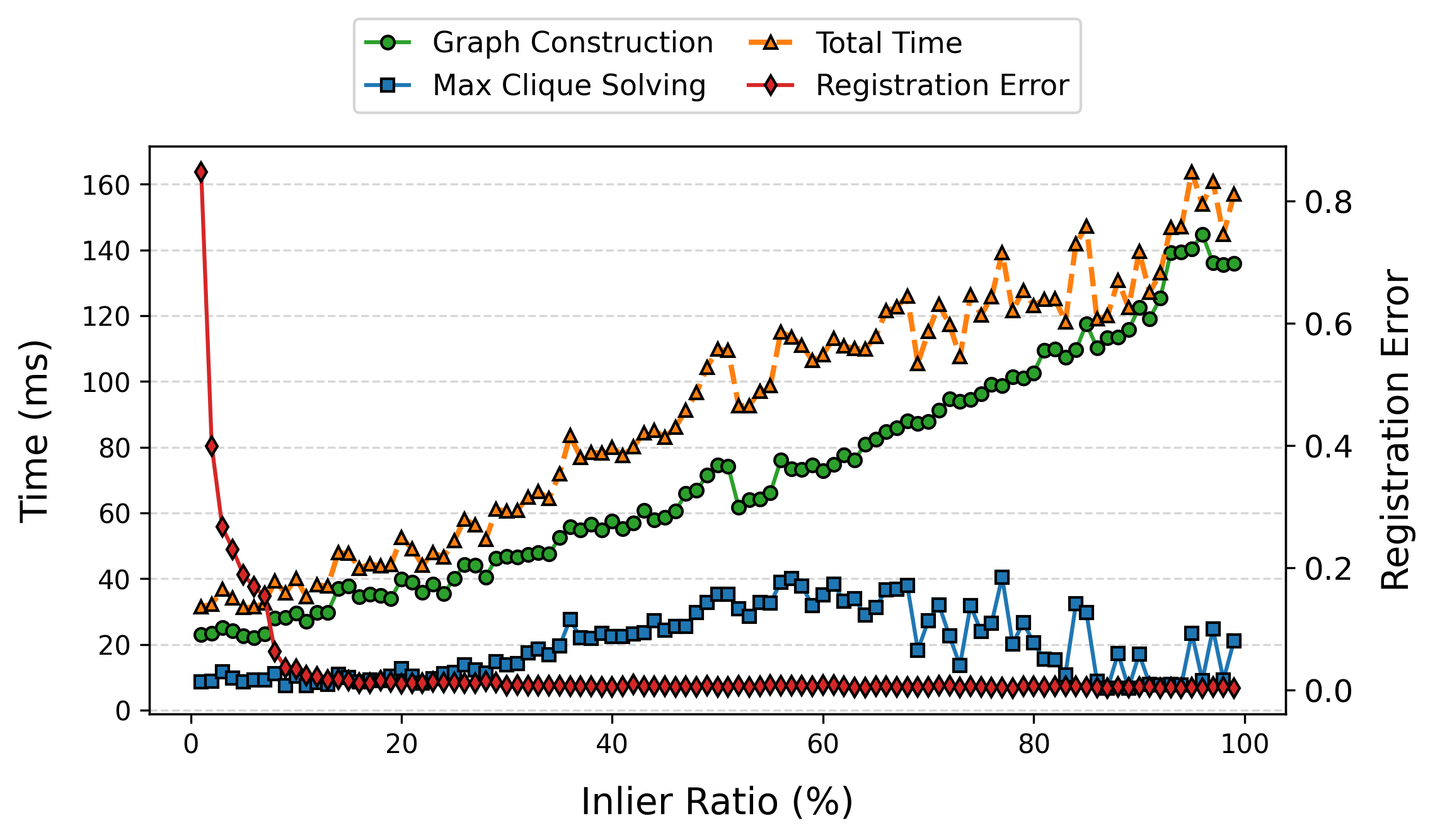}
    \caption{
        Computational profile of maximum clique-based registration under varying inlier ratios (total correspondences: 1000). Each inlier ratio was tested ten times with results averaged. 
        Graph construction time (green) grows with inlier ratio due to the addition of graph edges, while maximum clique solving time (blue) first increases then decreases as graph density enables aggressive pruning. Registration error (red) remains low at low inlier ratios, validating the robustness of maximum clique pruning even under extreme outlier rates. 
    }\label{fig:mc_time}
\end{figure}

Here, we aim to express that in scenarios with a high inlier ratio, pruning outliers via maximum clique algorithm is less efficient compared with directly employing Graduated Non-Convexity (GNC) for outlier removal and pose estimation. The reason we cite the literature \cite{bustos2019practical} is that the maximum clique solver we use in practice originates from this paper. Although, in theory, the maximum clique algorithm has exponential complexity in the worst case, the algorithm we employ is based on branch-and-bound with efficient pruning strategies. As a result, \textbf{its complexity initially increases linearly with the inlier ratio but decreases after reaching a certain threshold}. This is because, at the beginning, a higher inlier ratio leads to more edges in the graph, requiring the algorithm to explore more possibilities. However, once the inlier ratio exceeds a certain level, it provides a tight upper bound. As illustrated in the literature\cite{bustos2019practical}, a tight upper bound facilitates a more effective pruning. To validate this, we conducted an experiment on a registration example from our dataset to investigate the relationship between maximum clique search efficiency and the inlier ratio, as shown in Figure \ref{fig:mc_time}. As illustrated, the maximum clique search time aligns with our theoretical analysis.

From a practical perspective, we must also consider the time required to construct the graph. \textbf{As the number of inliers increases, more edges need to be established, which increases the graph construction time, even with parallelization strategies.} Additionally, we present the variation in registration error, demonstrating that even at low inlier ratios, the maximum clique-based outlier removal algorithm maintains robust outlier rejection. Given this finding, we adopt a practical approach: we first attempt registration using GNC. If the inlier ratio calculated from the GNC registration result is sufficiently high, we avoid using the slower maximum clique algorithm.

\section{Registrations in semantic noise}
\begin{figure*}[t]
    \centering
    \includegraphics[width=0.6\columnwidth]{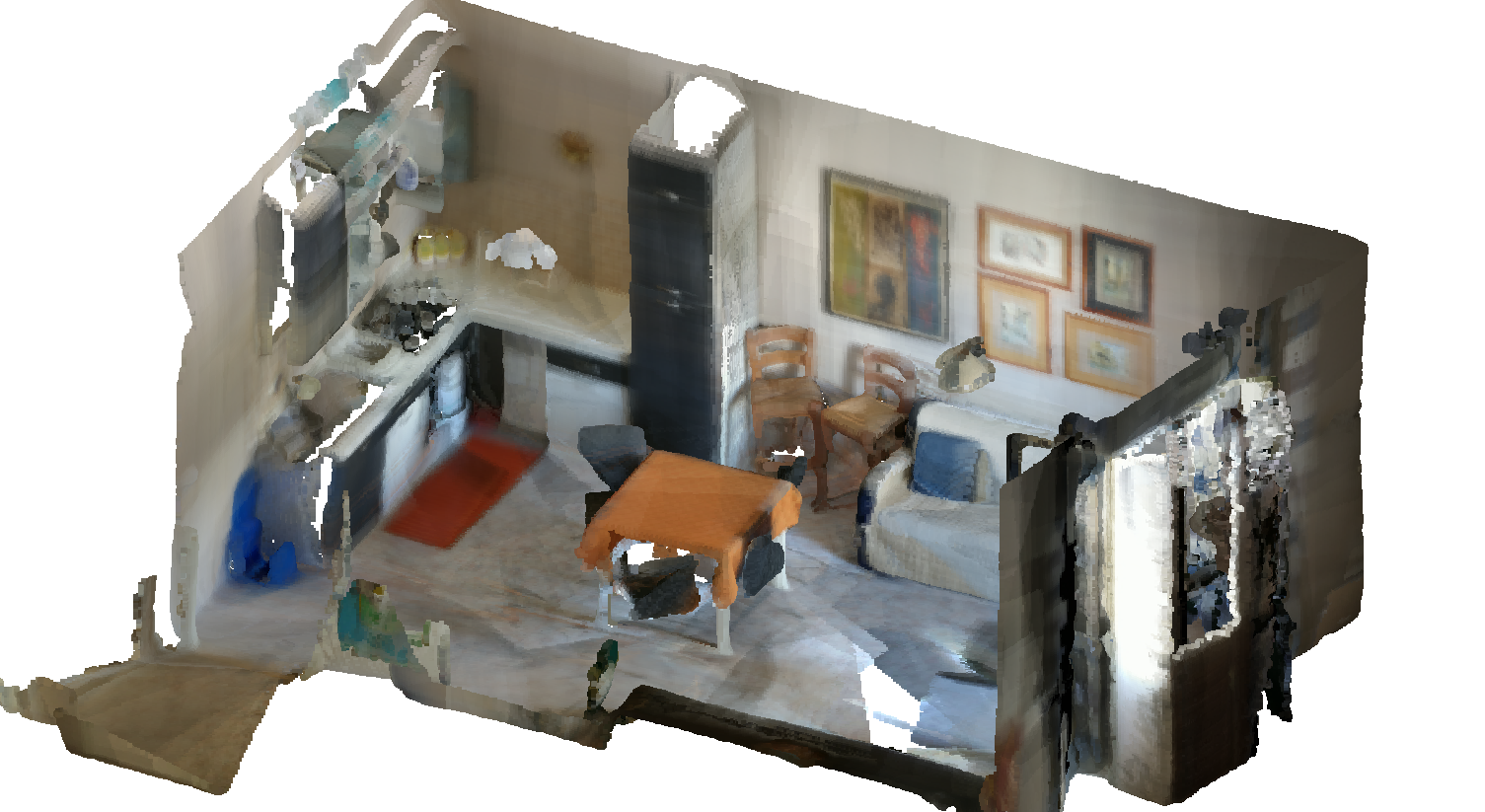}\vspace{-0.2cm}
    \includegraphics[width=0.6\columnwidth]{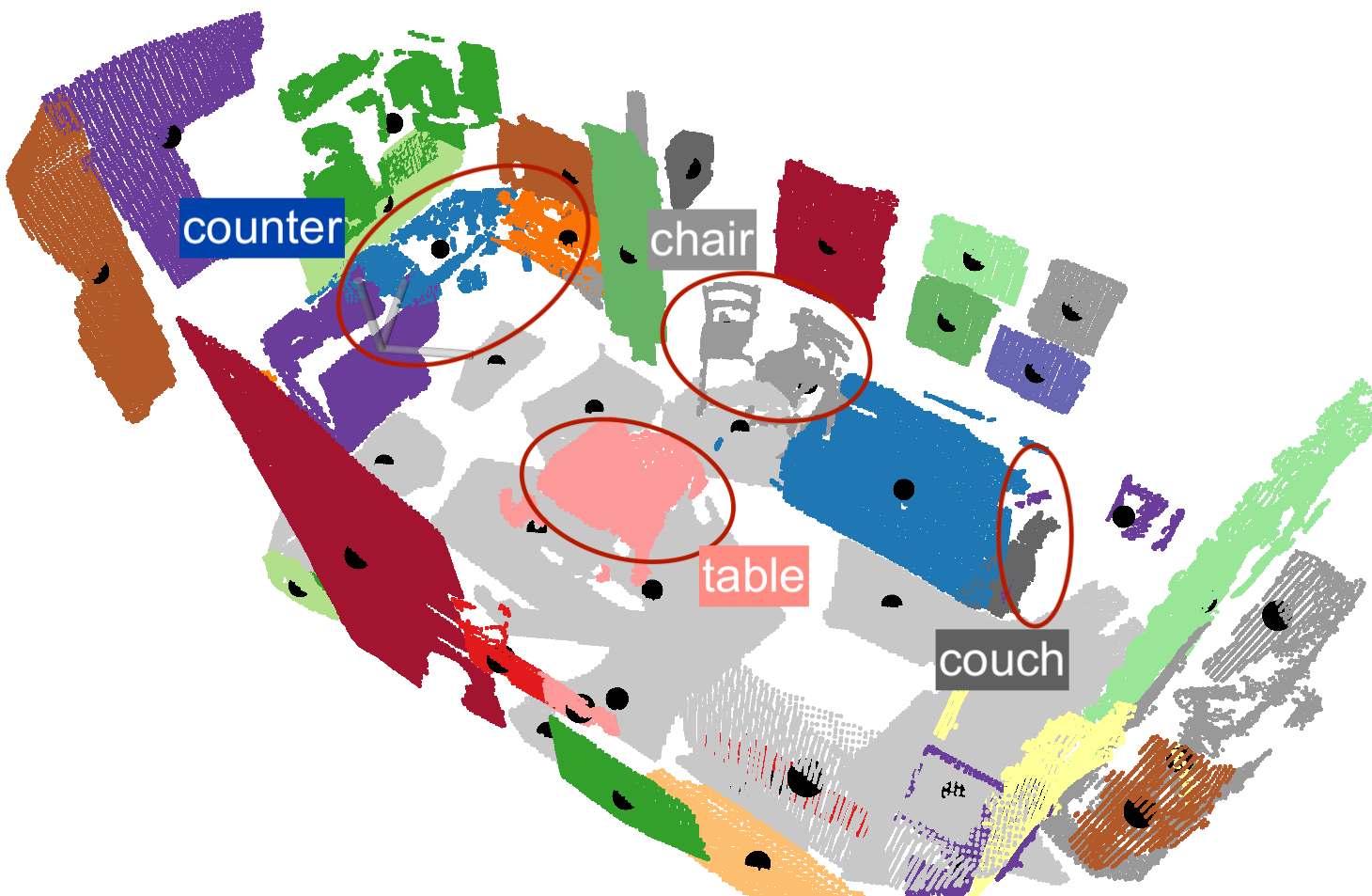}
    \includegraphics[width=0.7\columnwidth]{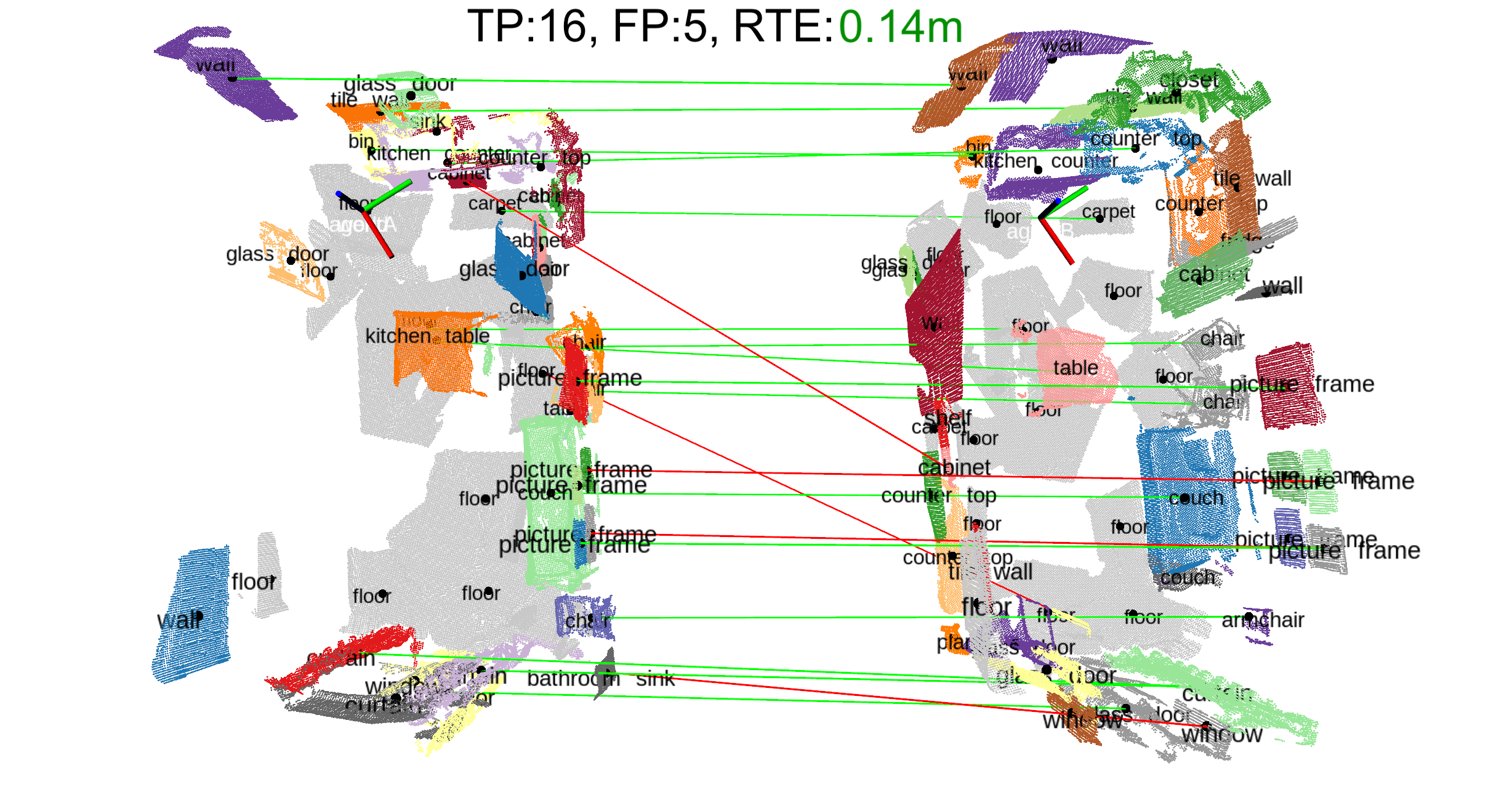}
    \includegraphics[width=0.6\columnwidth]{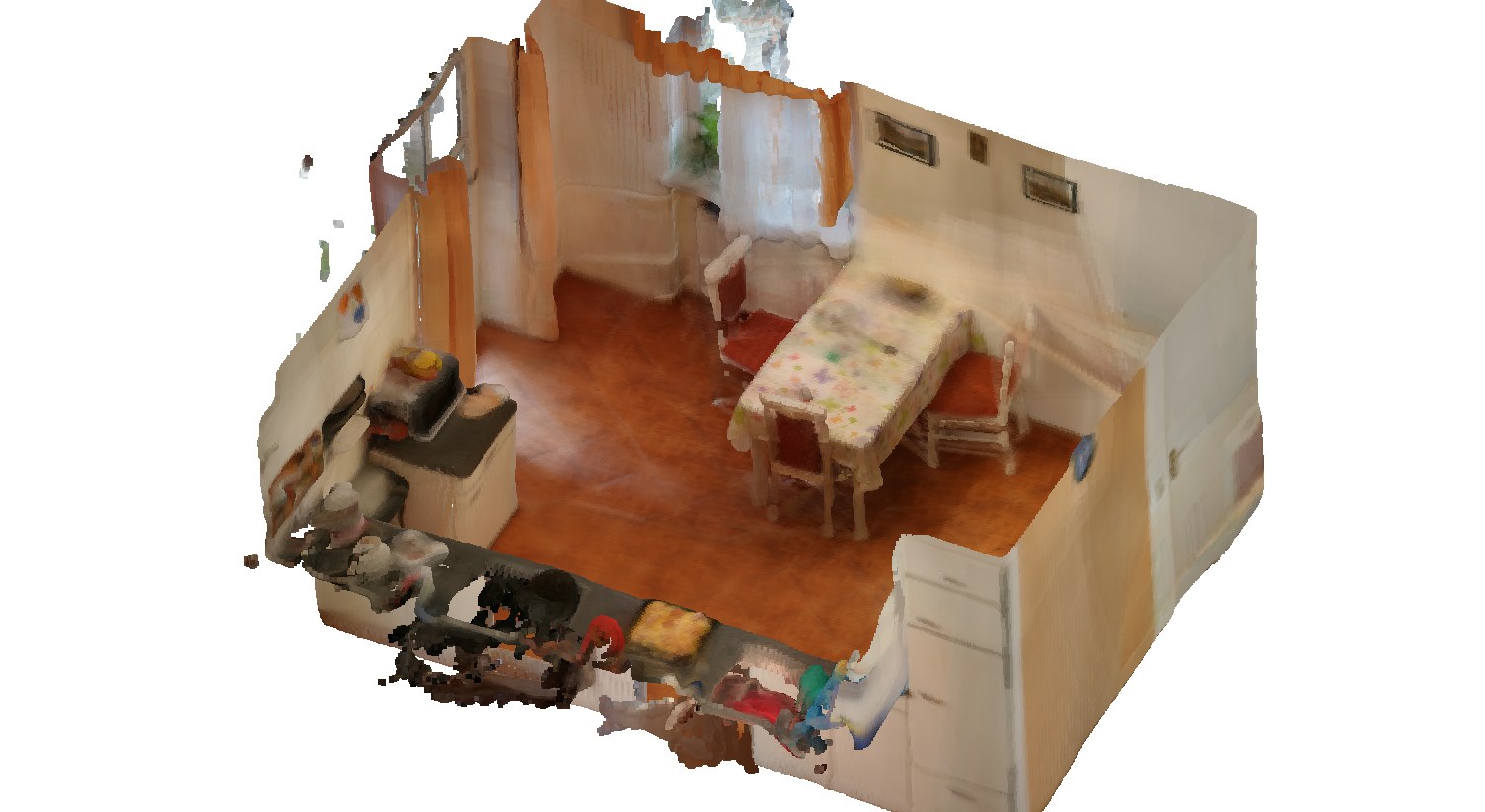}
    \includegraphics[width=0.6\columnwidth]{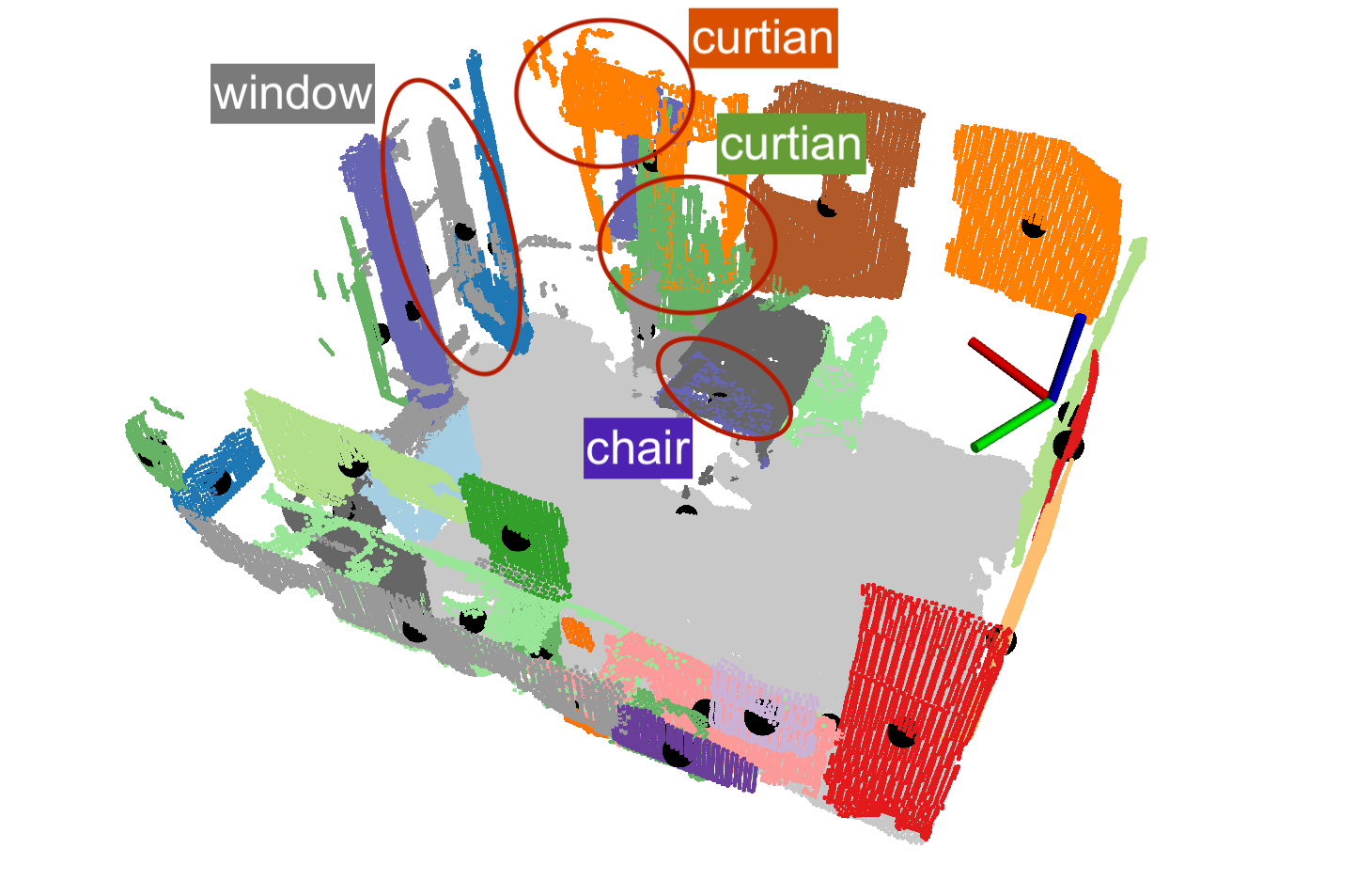}    
    \includegraphics[width=0.7\columnwidth]{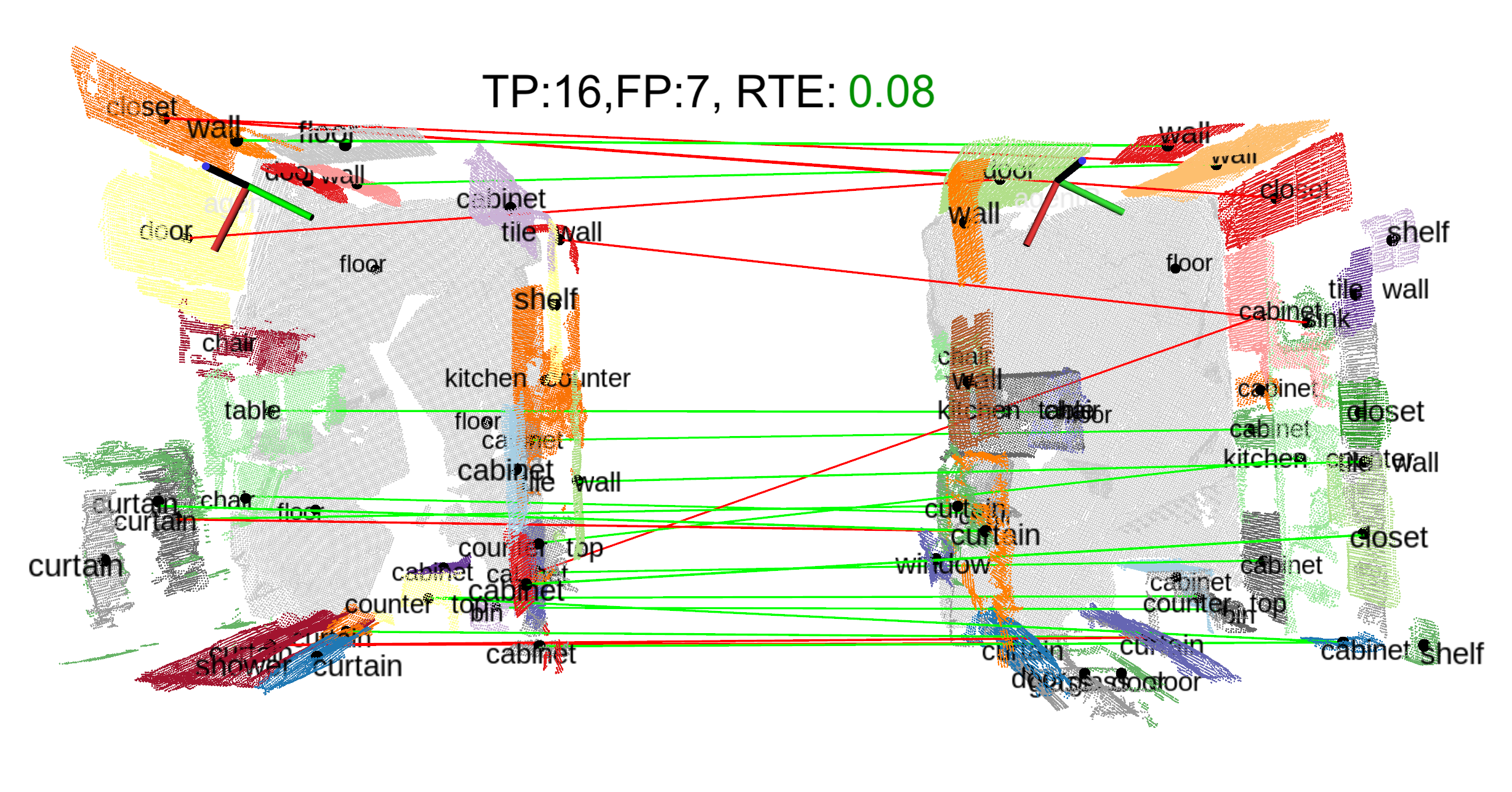}
    \makebox[0.6\columnwidth]{\small Colored point cloud}
    \makebox[0.6\columnwidth]{\small Semantic scene graph}
    \makebox[0.6\columnwidth]{\small Matched nodes}
    \caption{Two successful registration with semantic noise. In the semantic scene graph, we highlight a few of the nodes that are over-segmented, falsely labeled and partially reconstructed.}\label{fig:rio_successs}\vspace{-0.5cm}
\end{figure*}

\setlength{\subfigwidth}{0.66\columnwidth}
\begin{figure*}[h!]
   \vspace{+0.3cm}\includegraphics[width=0.88\subfigwidth]{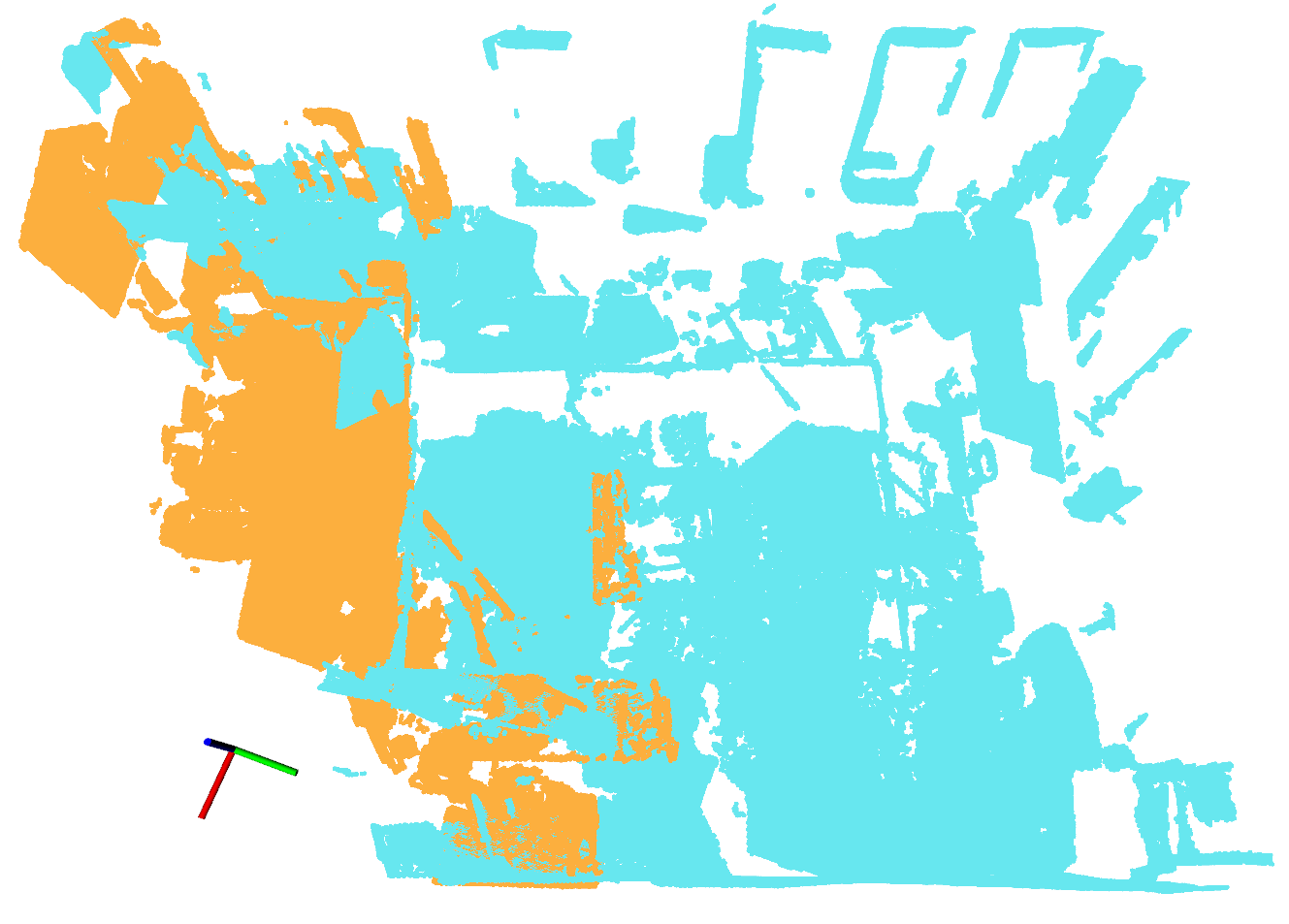}
  \includegraphics[width=\subfigwidth]{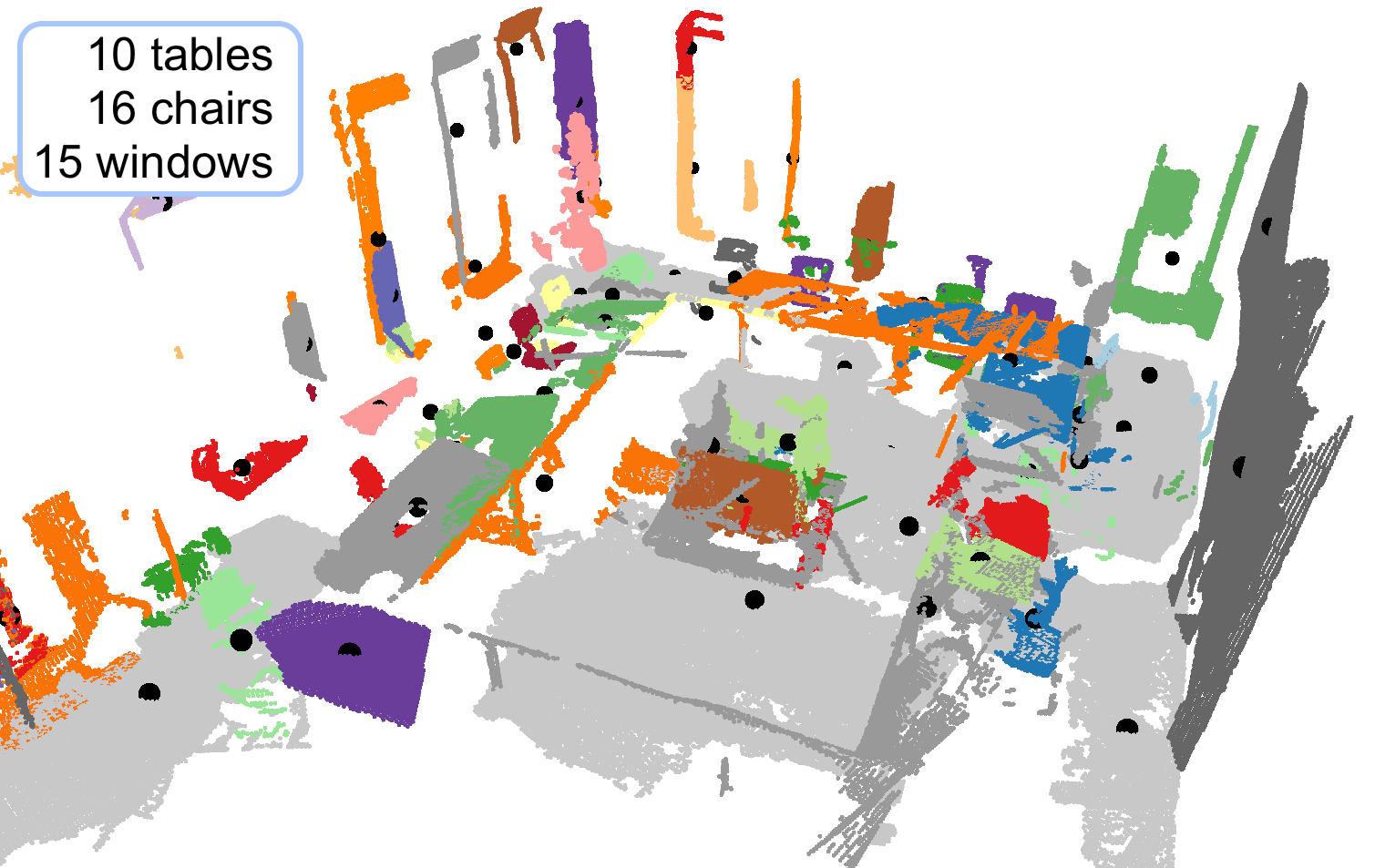}\vspace{-0.2cm}
   \includegraphics[width=1.3\subfigwidth]{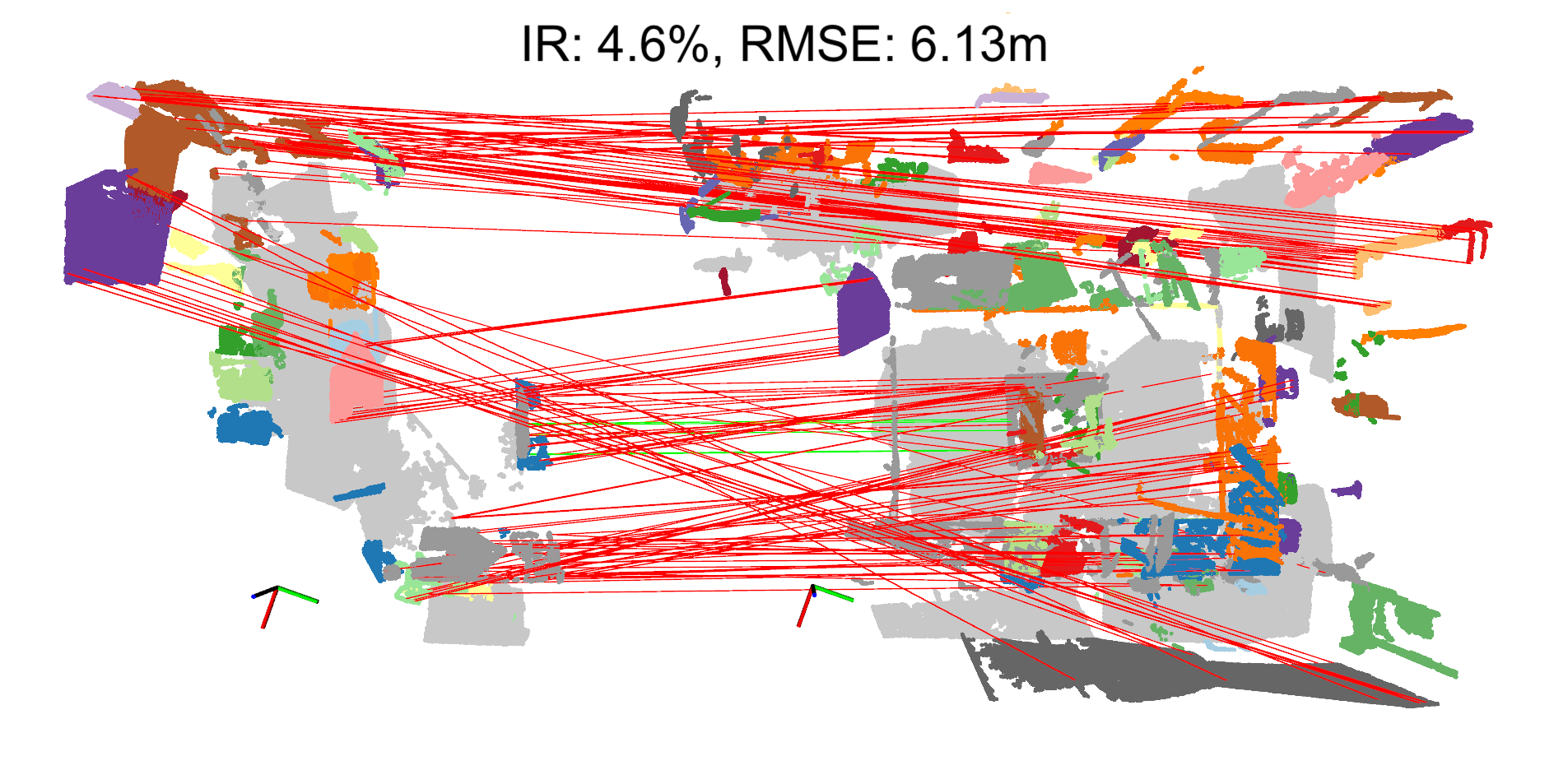}\vspace{+0.2cm}
   \makebox[1.9\columnwidth]{\small (a) Low overlapped scenes.}

   \includegraphics[width=0.8\subfigwidth]{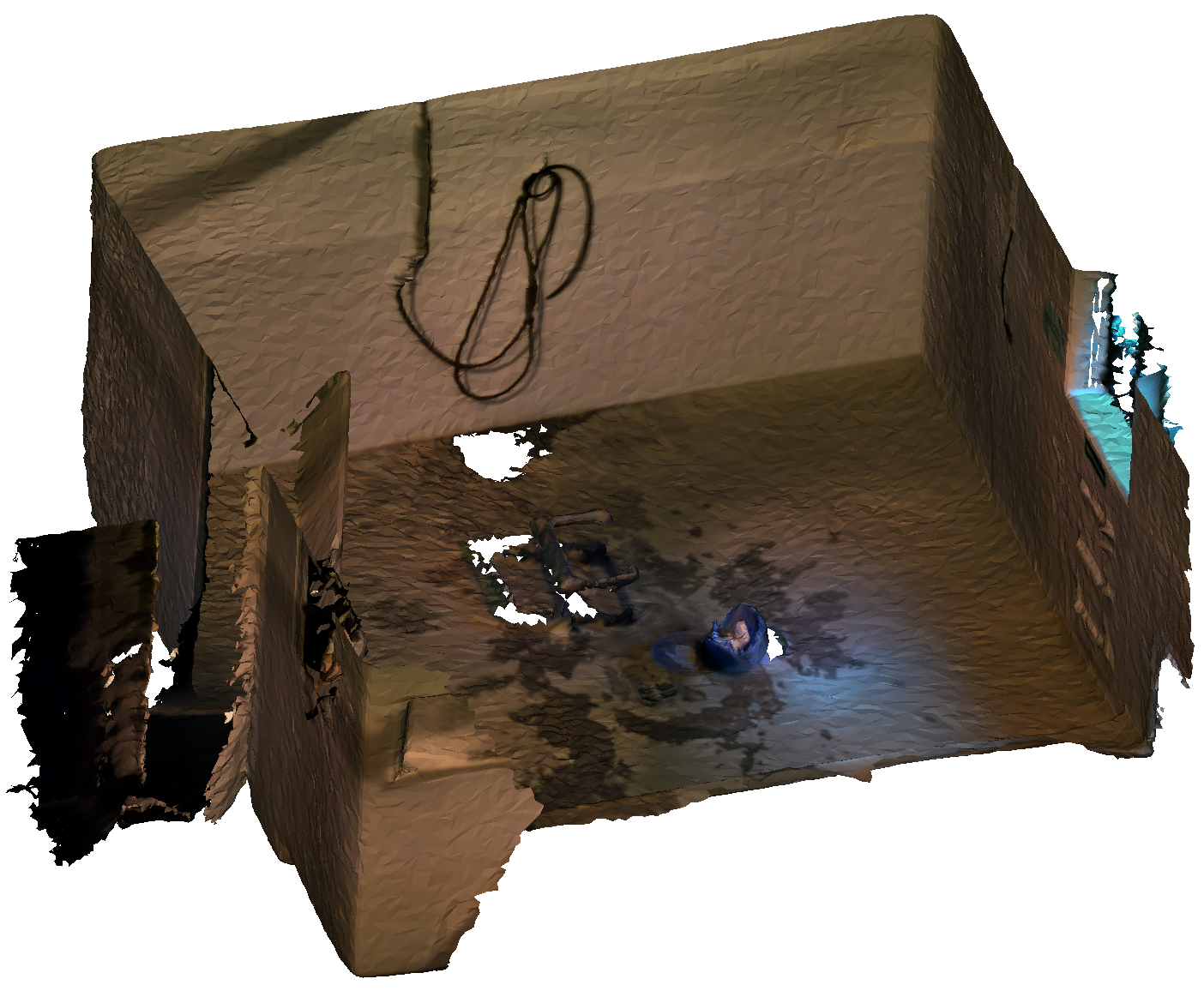}
  \includegraphics[width=1.1\subfigwidth]{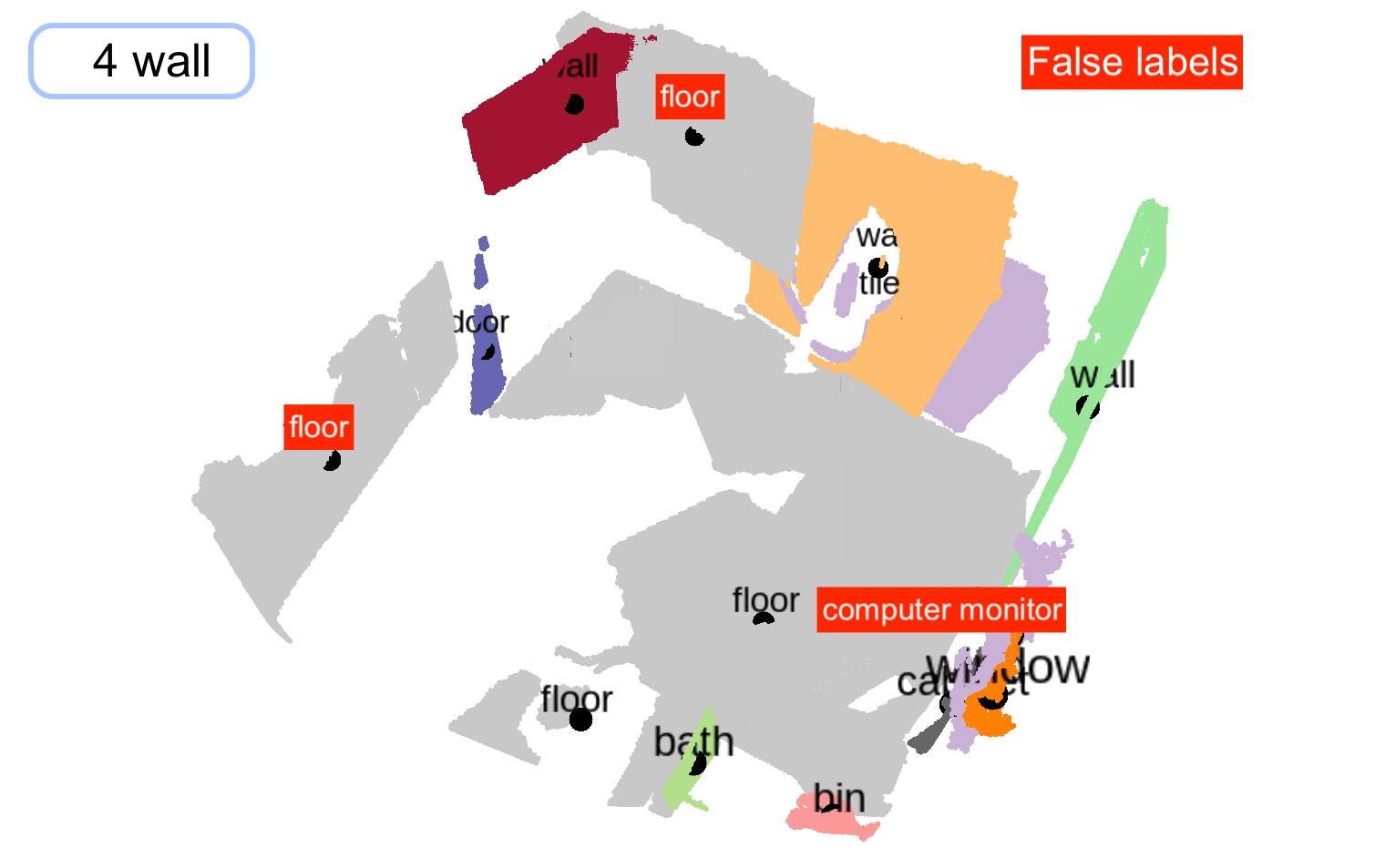}\vspace{-0.2cm}
   \includegraphics[width=1.2\subfigwidth]{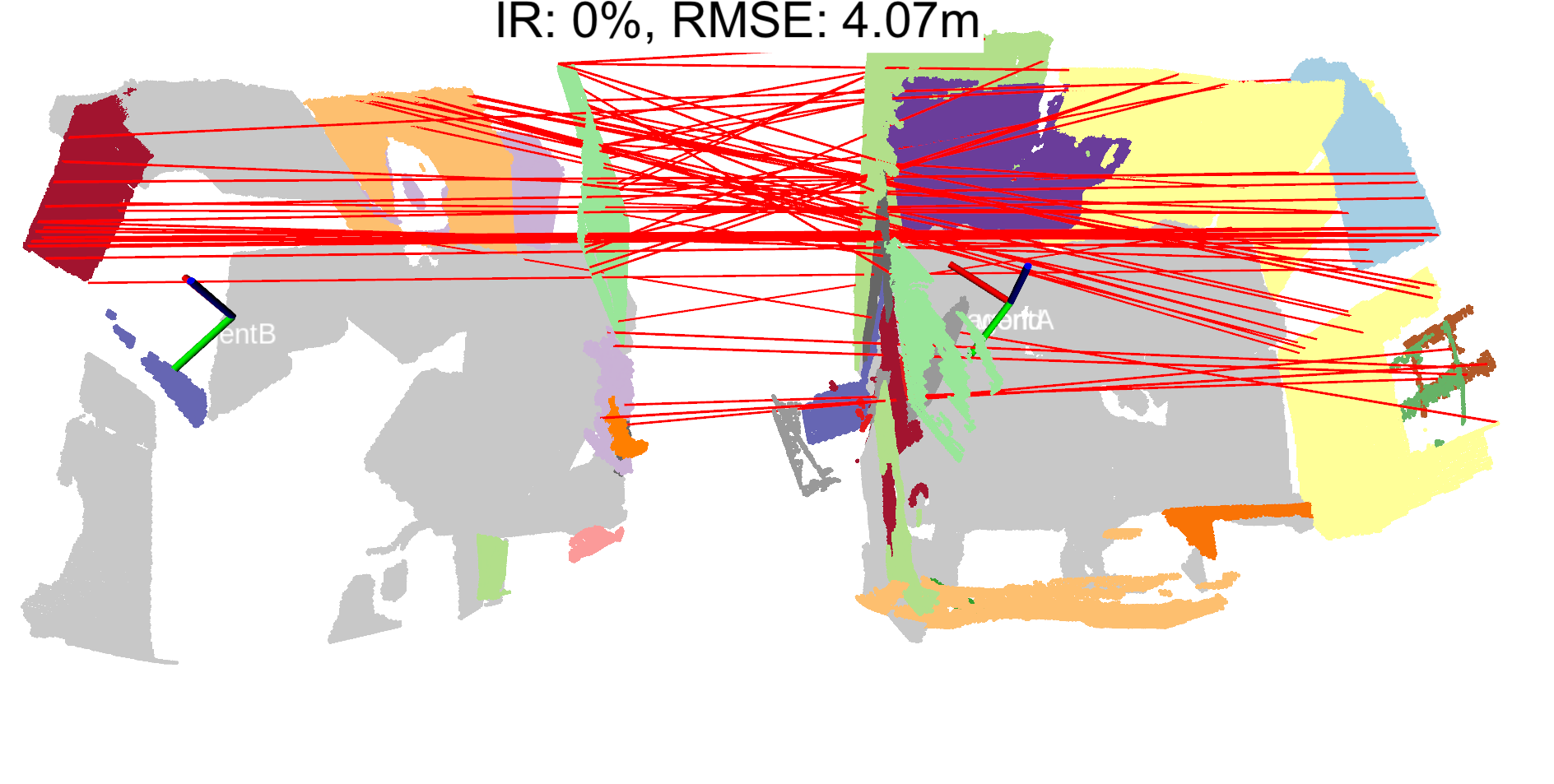}\vspace{+0.2cm}
   \makebox[1.9\columnwidth]{\small (b) Empty room.}

   \includegraphics[width=0.8\subfigwidth]{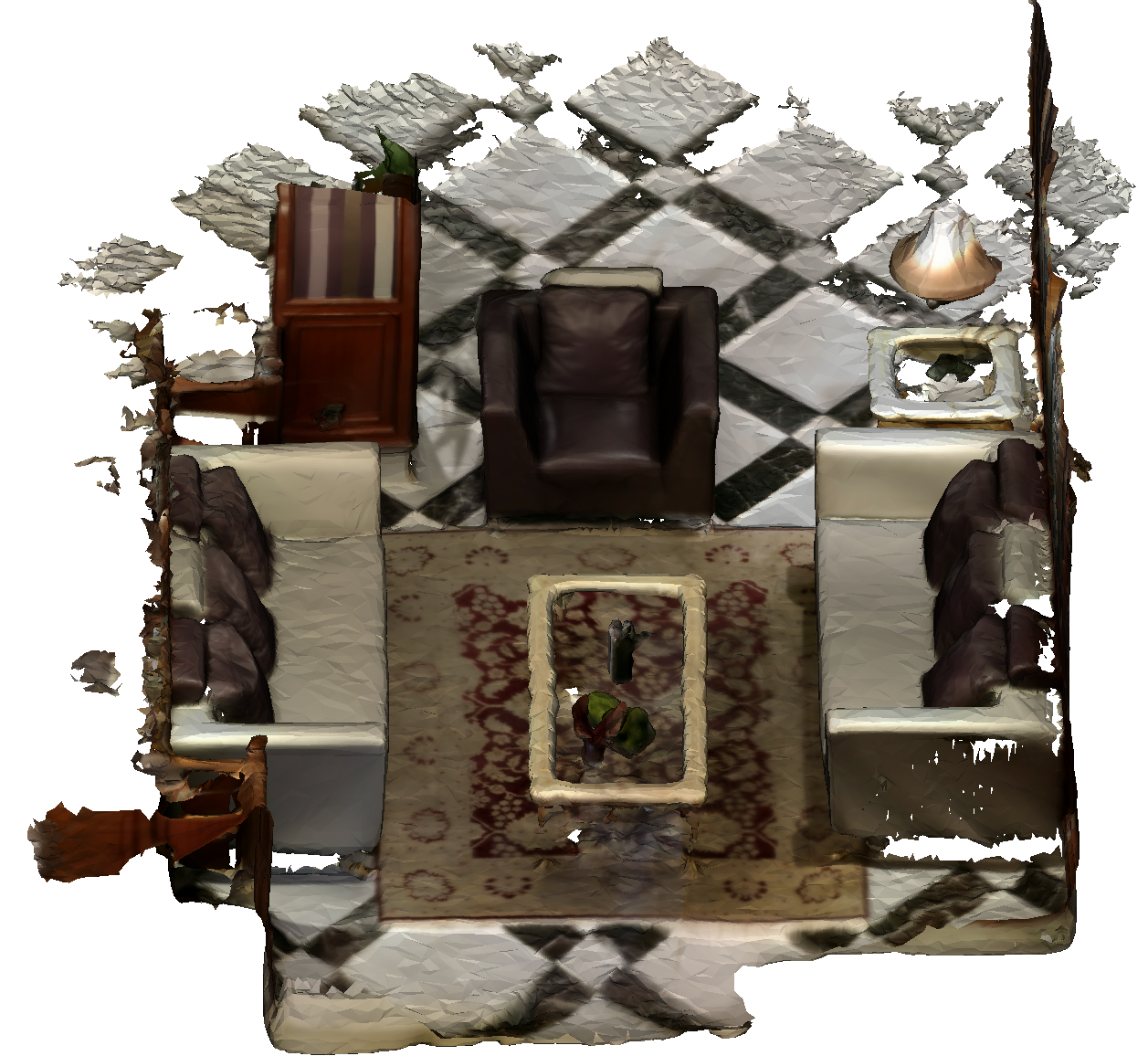}
  \includegraphics[width=1.1\subfigwidth]{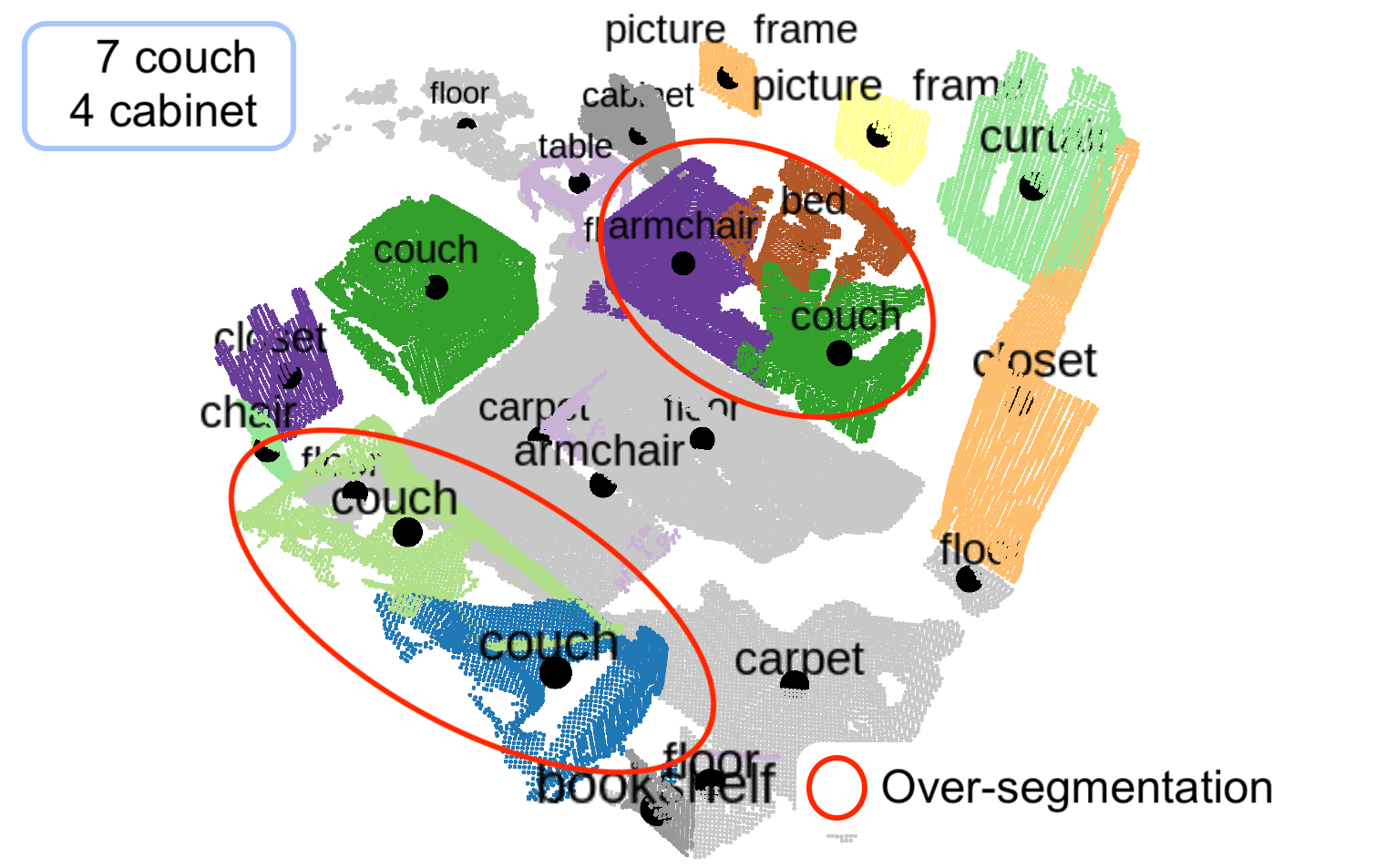}\vspace{-0.2cm}
   \includegraphics[width=1.2\subfigwidth]{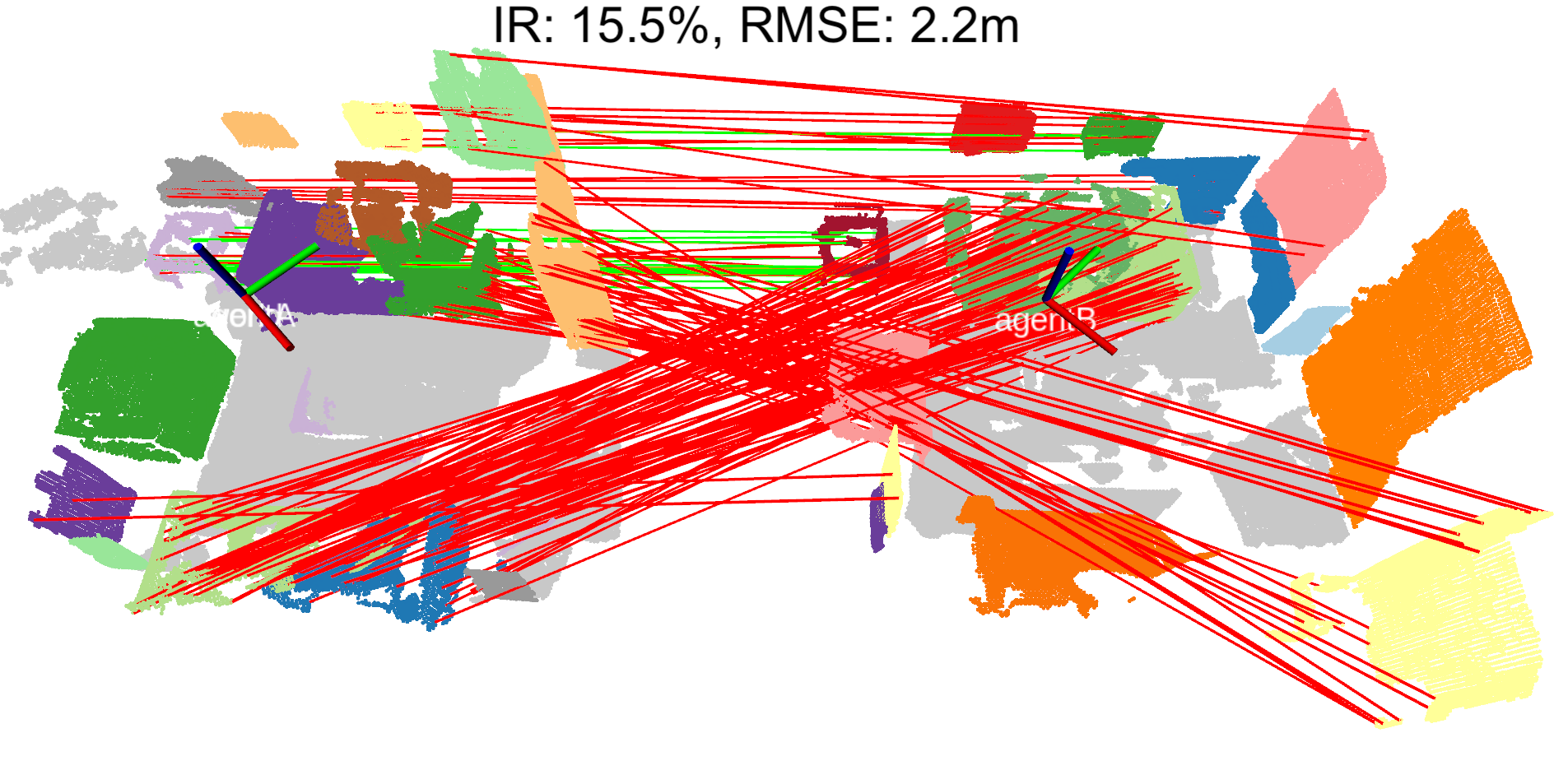}\vspace{+0.2cm}
   \makebox[1.9\columnwidth]{\small (c) Symmetric couch layout.}
    \caption{Failure registration. In the middle column, we highlight a few significant semantic noise. In the right column, we visualize the point correspondences.}\label{fig:failReg}
\end{figure*}
\subsubsection{Success cases}
We show two indoor scenes and highlight their semantic noise. As shown in Fig. \ref{fig:rio_successs}, SG-Reg registers them successfully.

\subsubsection{Failed cases}
We further analyze a few failure cases of SG-Reg to illustrate our upper limits. Some of the limitations can be further addressed in our future works. 
Fig. \ref{fig:failReg} shows three examples of failed registration. We analyze the reasons behind their failures.

\begin{itemize}
    \item \textit{Significant Semantic Noise}: Semantic noise is prevalent in all the scenes we reconstructed; however, the failed scenes exhibit more pronounced noise. As illustrated in the first column of Fig. \ref{fig:failReg}, these scenes contain numerous objects that are over-segmented and inaccurately labeled, which severely disrupts the spatial topological relationships between objects. Furthermore, many of these scenes feature objects belonging to identical semantic categories and positioned close to one another, leading to inaccurate node matches. This, in turn, results in a substantial number of outliers in the point correspondences, ultimately causing the registration failures.
    \item \textit{Low Overlapping Scenes}: The scene in Fig. \ref{fig:failReg}(a) exhibits very low overlap in their subvolumes, resulting in a significant number of outliers between the non-overlapping regions. Due to the low inlier ratio, SG-Reg fails to successfully register the scenes.
    \item \textit{Empty Room}: SG-Reg relies on diverse semantic objects to build descriptive node features. The scene in Fig. \ref{fig:failReg}(b) is an empty room devoid of furniture, containing only a few walls and a window with a noisy point cloud. Due to the lack of meaningful semantic objects, SG-Reg fails to identify any point inliers, resulting in registration failure.
    \item \textit{Symmetric Subvolumes}: The scene in Fig. \ref{fig:failReg}(c) exhibits a symmetric room layout, with a table in the center and two identical couches on either side. Although SG-Reg finds correspondences with a 15.5\% inlier ratio, many of these correspondences are adversarial, leading to a pseudo inlier ratio of $0\%$. Consequently, the registration fails.
\end{itemize}
To address the challenges shown in Fig. \ref{fig:failReg}, there are a few possible direction in the future. 1) Improving semantic scene graph quality by constructing an implicit semantic map. Currently, we rely on FM-Fusion\cite{liu2024fmfusion}, which is an explicit semantic mapping. The ambiguous semantic categories, occlusion, and segmentation noise affect the reconstruction quality. In implicit semantic mapping, we can design a neural network to refine the semantic instance map in the 3D space. It should be able to further improve the reconstruction quality, which benefits the downstream registration task. 2) Extracting the appearance feature from each semantic node and fusing them into the node features. 3) Visual image registration, such as HLoc, does not conflict with scene graph registration. We may incorporate the image keyframes into our semantic scene graph. It jointly considers the image matching and semantic node matching to avoid those long-tailed cases.


\section{Two-agent SLAM}
We supplement the implementation and additional results that have not been put in the main body due to page limits.

\subsubsection{Compute HLoc bandwidth}
We explain the HLoc configuration for calculating communication bandwidth in Table \ref{tab:comm}. In each keyframe, HLoc can extract fewer than $4096$ superpoints. Across all keyframe sequences, we have accumulated the total superpoints to be $460,000$. Each superpoint comprises a $256$-dimensional descriptor in \textit{float} format, a $2$-dimensional coordinate in \textit{int} format, and a scalar score in \textit{float}. The total data size is calculated in megabytes (MB) and is presented in Table \ref{tab:comm}.

\subsubsection{Discuss registration}
\begin{table}[ht]
    \centering
    \begin{tabular}{c c c c c c c}
        \toprule
         & & NR(\%) & NP(\%) & IR(\%) & RR(\%) &RMSE(m) \\
        \hline
        \multirow{3}{*}{Coarse} & Easy & $83.9$ & $88.4$ & - & $83.5$ & $0.08$ \\ 
        & Median & $68.0$  & $68.7$ & - & $51.2$ & $0.11$ \\
        & Difficult & $42.3$ & $26.2$ & - & $0.0$ & - \\
        \hline
        \multirow{3}{*}{Dense} & Easy & $86.1$ & $89.0$ & $6.4$ & $100.0$ & $0.05$ \\ 
        & Median & $77.5$ & $75.5$ & $5.8$ & $75.0$ & $0.08$ \\ 
        & Difficult & $62.1$ & $35.3$ & $1.4$ & $0.0$ & -\\
        \bottomrule
    \end{tabular}
    \caption{Evaluate registration between the agents.}\label{tab:online_registration}\vspace{-0.3cm}
\end{table}

We also report our results following the registration metrics. We calculate the IoU of the input scene graphs at each query frame. The query frames are grouped into three sets: Difficult set has $\text{IoU} \in [0.1,0.3)$; Median set has $\text{IoU} \in [0.3,0.7)$; Easy set has $\text{IoU}\in [0.7,1.0)$. The RMSE threshold is $0.2$m.
As shown in Table. \ref{tab:online_registration}, with coarse messages, our method has already registered $83.5\%$ of the easy scenes. Since we use a conservative registration strategy, our success rate is still low in the difficult set.
The result further implies that the best way to apply SG-Reg in SLAM is to run registration at frames with large enough IoU. The failed registration at low IoU can be rejected by setting a threshold of the matched nodes.


\end{document}